%% file: main.tex
\pdfoutput=1

\documentclass[11pt]{article}
\input{misc/commands}
\usepackage[final]{acl}

\usepackage{times}
\usepackage{latexsym}

\usepackage[T1]{fontenc}

\usepackage[utf8]{inputenc}

\usepackage{microtype}

\usepackage{inconsolata}

\usepackage{graphicx}
\usepackage[utf8]{inputenc} 
\usepackage[T1]{fontenc}    
\usepackage{hyperref}       
\usepackage{url}            
\usepackage{booktabs}       
\usepackage{amsfonts}       
\usepackage{nicefrac}       
\usepackage{microtype}      
\usepackage{xcolor}         
\usepackage{float}
\usepackage{stfloats}

\usepackage{ulem}
\usepackage{multirow}
\usepackage{diagbox}
\usepackage{booktabs}
\usepackage{xcolor}
\definecolor{mygreen}{HTML}{2CB600}
\usepackage{subfig}
\usepackage{xspace}
\usepackage{appendix}
\usepackage{graphicx}
\usepackage{epsfig}
\usepackage{amsmath}
\usepackage{pifont}
\usepackage{tabularx}
\usepackage{seqsplit}
\usepackage{nameref}
\usepackage{varioref}
\usepackage{hyperref}
\usepackage{cleveref}
\usepackage{todonotes}
\usepackage{comment}

\usepackage{amssymb}
\usepackage{pifont}
\usepackage{natbib}

\usepackage[section]{placeins}
\usepackage{wrapfig}
\usepackage{amssymb}
\usepackage[final]{acl}
\usepackage{pifont}

\title{ChartInsights: Evaluating Multimodal Large Language Models for Low-Level Chart Question Answering}

\author{%
~\textbf{Yifan Wu$^1$\thanks{Equal contribution}},%
~\textbf{Lutao Yan}$^3$\footnotemark[1],%
~\textbf{Leixian Shen}$^2$,%
~\textbf{Yunhai Wang}$^4$,%
~\textbf{Nan Tang}$^{1,2}$,%
~\textbf{Yuyu Luo}$^{1,2}$\thanks{Yuyu Luo is the corresponding author}
\\
$^{1}$The Hong Kong University of Science and Technology (Guangzhou)\\
$^{2}$The Hong Kong University of Science and Technology\\
$^{3}$South China University of Technology,~
$^{4}$Renmin University of China\\
 \texttt{ywu012@connect.hkust-gz.edu.cn}, \texttt{yuyuluo@hkust-gz.edu.cn}
\\
\texttt{\url{https://chartinsight.github.io/}}
}

\begin{document}
\maketitle
\input{secs/abstract}
\input{secs/introduction}
\input{secs/experiments}

\input{secs/conclusion}
\input{secs/limitations}

\bibliography{custom}

\clearpage

\appendix
\section*{Appendices}
\input{appendix/dataset}

\input{appendix/experiment}

\input{secs/methods}
\input{secs/related}
\input{appendix/prompt}

\end{document}

%% file: misc/commands.tex
\newcommand{\eat}[1]{}

\usepackage[many]{tcolorbox}  
\newtcolorbox{finding}{
	sharpish corners, 
	boxrule = 0pt,
	toprule = 4.5pt, 
	enhanced,
	fuzzy shadow = {0pt}{-2pt}{-.5pt}{0.5pt}{black!25} 
}

\usepackage{latexsym}
\usepackage{amsfonts}
\usepackage{amsmath}
\usepackage{amssymb}
\usepackage{colortbl}
\usepackage{epsfig}
\usepackage{xspace}
\usepackage{graphicx}
\usepackage{cite}
\usepackage{comment}
\usepackage{booktabs}
\usepackage{paralist,bbding,pifont}
\usepackage{enumitem}
\setlist{noitemsep,parsep=0pt,partopsep=0pt, leftmargin=10pt}

\usepackage{algorithm} 
\usepackage{algorithmic}  
\usepackage[algo2e]{algorithm2e}

\definecolor{green}{RGB}{0,128,0}

\definecolor{yellow}{RGB}{255,200,18}

\newcommand{\stab}{\vspace{1.2ex}\noindent}

\newcommand{\bi}{\begin{itemize}}
\newcommand{\ei}{\end{itemize}}

\newcommand{\be}{\begin{enumerate}}
\newcommand{\ee}{\end{enumerate}}
\newcommand{\beqn}{\begin{eqnarray*}}
\newcommand{\eeqn}{\end{eqnarray*}}

\newcommand{\etitle}[1]{\vspace{1mm}\noindent{\underline{\textit{#1}}}}

\newcommand{\ie}{\textit{i.e.,}\xspace}
\newcommand{\eg}{\textit{e.g.,}\xspace}

\usepackage{tabu}                      
\usepackage{booktabs}                  
\usepackage{lipsum}                    
\usepackage{mwe}                       

\usepackage{mathptmx}                  
\newcommand{\tdataset}{ChartInsights\xspace}
\newcommand{\dataset}{\textit{ ChartInsights}\xspace}
\newcommand{\ttextp}{Chain-of-Charts\xspace}
\newcommand{\textp}{\textit{Chain-of-Charts}\xspace}

\newcommand{\mllms}{MLLMs\xspace}
\newcommand{\gptfv}{GPT-4V\xspace}
\newcommand{\gptfo}{GPT-4o\xspace}

%% file: secs/abstract.tex
\begin{abstract}


Chart question answering (ChartQA) tasks play a critical role in interpreting and extracting insights from visualization charts. While recent advancements in multimodal large language models (MLLMs) like GPT-4o have shown promise in high-level ChartQA tasks, such as chart captioning, their effectiveness in low-level ChartQA tasks (\eg identifying correlations) remains underexplored.
In this paper, we address this gap by evaluating MLLMs on low-level ChartQA using a newly curated dataset, \dataset, which consists of 22,347 (chart, task, query, answer) covering 10 data analysis tasks across 7 chart types. 
We systematically evaluate 19 advanced MLLMs, including 12 open-source and 7 closed-source models. The average accuracy rate across these models is 39.8\%, with GPT-4o achieving the highest accuracy at 69.17\%.
To further explore the limitations of MLLMs in low-level ChartQA, we conduct experiments that alter visual elements of charts (\eg changing color schemes, adding image noise) to assess their impact on the task effectiveness. 
Furthermore, we propose a new textual prompt strategy, \textp, tailored for low-level ChartQA tasks,  which boosts performance by 14.41\%, achieving an accuracy of 83.58\%. 
Finally, incorporating a visual prompt strategy that directs attention to relevant visual elements further improves accuracy to 84.32\%. 


\end{abstract}

%% file: secs/introduction.tex
\section{Introduction}
\label{sec:introduction}


Visualization charts can effectively convey data insights, but the abundance of information they provide makes it challenging for users to efficiently and accurately extract desired information~\cite{DBLP:journals/pvldb/XieLLT24,DBLP:journals/tvcg/LiLFZLL24, DBLP:journals/pacmmod/LuoZ00CS23}. Automated chart question answering (ChartQA)~\cite{DBLP:conf/chi/ZengB23, ye2024generative} is crucial to help users pinpoint relevant information based on their intents~\cite{DBLP:conf/infovis/AmarES05,lowleveltasks}.

\begin{figure}[t!]
	\centering
	\includegraphics[width=\columnwidth]{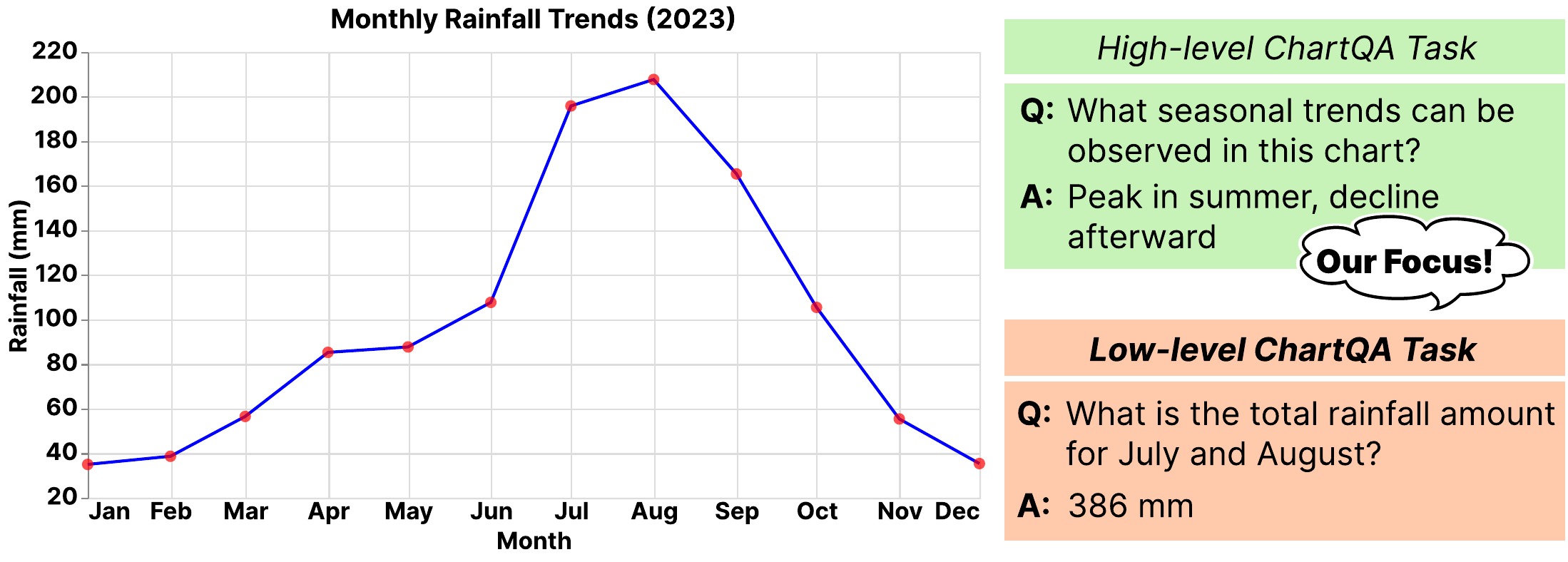}
	\caption{Examples of Two Types of ChartQA Tasks}
	\label{fig:example}
 \vspace{-1.em}
\end{figure}

\paragraph{High-Level and Low-Level ChartQA Tasks.}
ChartQA tasks can generally be categorized into two types: {\bf high-level} tasks and {\bf low-level} tasks. High-level tasks focus on questions that require understanding the overall context or summary of the chart, involving broader, goal-oriented inquiries that seek to understand overarching trends or patterns. In contrast, low-level tasks focus on specific, detail-oriented inquiries that seek precise data points or comparisons within the chart, involving straightforward, factual information retrieval~\cite{lowleveltasks, DBLP:conf/infovis/AmarES05}.
For example, as shown in Figure~\ref{fig:example}, given the same line chart showing rainfall in different months of the same year, a high-level task might ask about the cyclical or seasonal trends in the chart, while a low-level task would be more focused on the data itself in the chart, such as asking how much total rainfall there was in July and August.

Traditionally, ChartQA has been a challenging problem due to the limited capabilities in natural language understanding and the high complexity of chart reasoning~\cite{DBLP:journals/pvldb/LiLCLT24, DBLP:conf/cidr/0001YF0LH24, DBLP:journals/corr/abs-2408-05109}. Fortunately, recent advancements in multimodal large language models ({MLLMs}) have made it possible for users to interact with systems using natural language to extract specific information from data across various modalities. This progress has illuminated new possibilities for ChartQA on different levels of tasks.

\begin{table*}[t!]
	\centering
	\caption{Comparison with Existing Datasets}
	\label{Datasets_comparison}
	\vspace{-0.5em}
	\resizebox{\textwidth}{!}{%
		\begin{tabular}{cccccccccccc@{}}
			\toprule
			Task Levels
			& Datasets
			& \#-Task Types
			& \#-Chart Types
			& \#-Charts & \#-Queries
			&  \#-Queries/\#-Charts
			& 
			Metadata?
			\\ \midrule
			\multicolumn{1}{c}{\multirow{2}{*}{High-level ChartQA}} 
			& \multicolumn{1}{c}{DVQA~\cite{dvqa}} & 3  & 1  & 300K  & 3.5M  & 11.7    & 
			\Checkmark  \\
			& \multicolumn{1}{c}{ChartQA~\cite{chartqa}}  & 4  & 3  & 4.8K  & 9.6K & 2  & 
			\Checkmark         
			\\ \midrule
			\multicolumn{1}{c}{\multirow{3}{*}{Mix}} 
			& \multicolumn{1}{c}{FigureQA~\cite{figureqa}} & 6   & 5  & 120K   & 1.5M  & 12.9   & 
			\XSolidBrush      
			\\  
			& \multicolumn{1}{c}{ChartLlama~\cite{chartllama}}  & 7  & 10  & 11K & 160K & 14.5  & 
			\XSolidBrush      
			\\ 
			& \multicolumn{1}{c}{ChartBench~\cite{chartbench}} & 4 & 11   & 2.1K  & 16.8K & 8  & 
			\XSolidBrush                          
			\\ \midrule
			
			\multicolumn{1}{c}{Low-level ChartQA}  & \multicolumn{1}{c}{\bf \tdataset (ours)} & \textbf{10} & 7 & 2K & 22K  & 11.2   & 
			\Checkmark  
			\\ \bottomrule
		\end{tabular}
	}
\end{table*}

\paragraph{Prior Art: High-Level ChartQA with \mllms.}
Recent studies have explored the capabilities of \mllms in performing high-level ChartQA~\cite{chartreader, enhanced_chart_understanding, LVLMS_understanding_charts, chartbench, chartllama, chartx, unichart}. The findings reveal that state-of-the-art \mllms like \gptfo have demonstrated promising results in addressing high-level tasks, and have outlined future research directions.

\paragraph{\mbox{Our Focus: Low-Level ChartQA with \mllms}.}
Existing evaluations of ChartQA primarily focus on high-level tasks, such as chart captioning, while overlooking low-level ChartQA tasks (\eg characterizing distributions) that humans frequently encounter in daily life. Specifically, studies from the visualization and visual analysis community have well-defined 10 widely used low-level ChartQA tasks~\cite{DBLP:conf/infovis/AmarES05, lowleveltasks}.
Thus, our study seeks to systematically evaluate the effectiveness of MLLMs in addressing these 10 low-level ChartQA tasks.

\paragraph{Contributions.} The key contributions are:

\stab(1) \textbf{ChartInsights Dataset.}
 We curate ChartInsights, \textit{the first} dataset for evaluating low-level data analysis tasks on charts. \tdataset includes diverse chart variants, textual and visual prompts, and comprehensive metadata, enabling the investigation of MLLMs' performance across various low-level ChartQA scenarios.

\vspace{-.5em}
\stab(2)    \textbf{Comprehensive Evaluations.} Our study establishes benchmarks by evaluating 19 MLLMs on 10 low-level ChartQA tasks, providing valuable insights into the current capabilities of MLLMs in processing and analyzing chart information.

\vspace{-.5em}
\stab(3)   \textbf{New Experimental Findings.} 
We summarize 12 experimental findings, highlighting the importance of visual prompts, chart elements, and image quality in performing low-level ChartQA tasks.

\vspace{-.5em}
\stab(4)   \textbf{Chain-of-Charts.}
We introduce the \textit{Chain-of-Charts} strategy, a new textual prompt designed to enhance MLLMs' reasoning capabilities in ChartQA tasks by leveraging a series of interconnected question-answer pairs to guide the model.

%% file: secs/experiments.tex
\section{\tdataset Dataset}
\label{sub:dataoverview}

\begin{figure}[t!]
	\centering
	\includegraphics[width=\columnwidth]{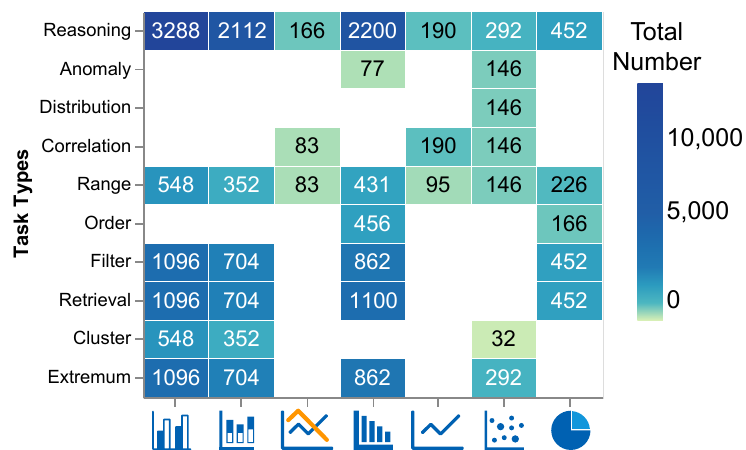}
	\vspace{-1em}
	\caption{Low-level Tasks vs. Chart Types}
	\label{fig:dataset_stat}
        \vspace{-1em}
\end{figure}

Since no dataset exists for low-level ChartQA tasks, we construct a large-scale dataset, ChartInsights, to systematically evaluate the performance of MLLMs in these tasks. The construction details of ChartInsights are provided in Appendix~\ref{subsec:construct}.

\paragraph{Overview of \tdataset.}
\tdataset contains 22,347 (chart, task, query, answer) samples across 7 chart types for 10 low-level data analysis tasks on charts~\cite{DBLP:conf/infovis/AmarES05}.
Figure~\ref{fig:dataset_stat} shows the distribution of 10 low-level tasks and 7 chart types. 
\textit{Please refer to Appendix~\ref{app:dataset} for more details.}


\paragraph{Comparison with Existing Datasets.}
As shown in Table~\ref{Datasets_comparison}, \tdataset differs from existing ChartQA datasets by emphasizing low-level data analysis tasks. 
It covers 10 distinct tasks across 7 chart types, including 2000 charts with an average of 11.2 queries each. 
Additionally, all relevant metadata are available, making ChartInsights a valuable resource for future research in ChartQA.
\textit{\mbox{Refer to Appendix~\ref{sub:datagoal} for further discussions.}}

\section{Experiments}
\label{sec:expt}

\subsection{Experimental Design}

\begin{figure*}[t!]
	\centering
	\includegraphics[width=\textwidth]{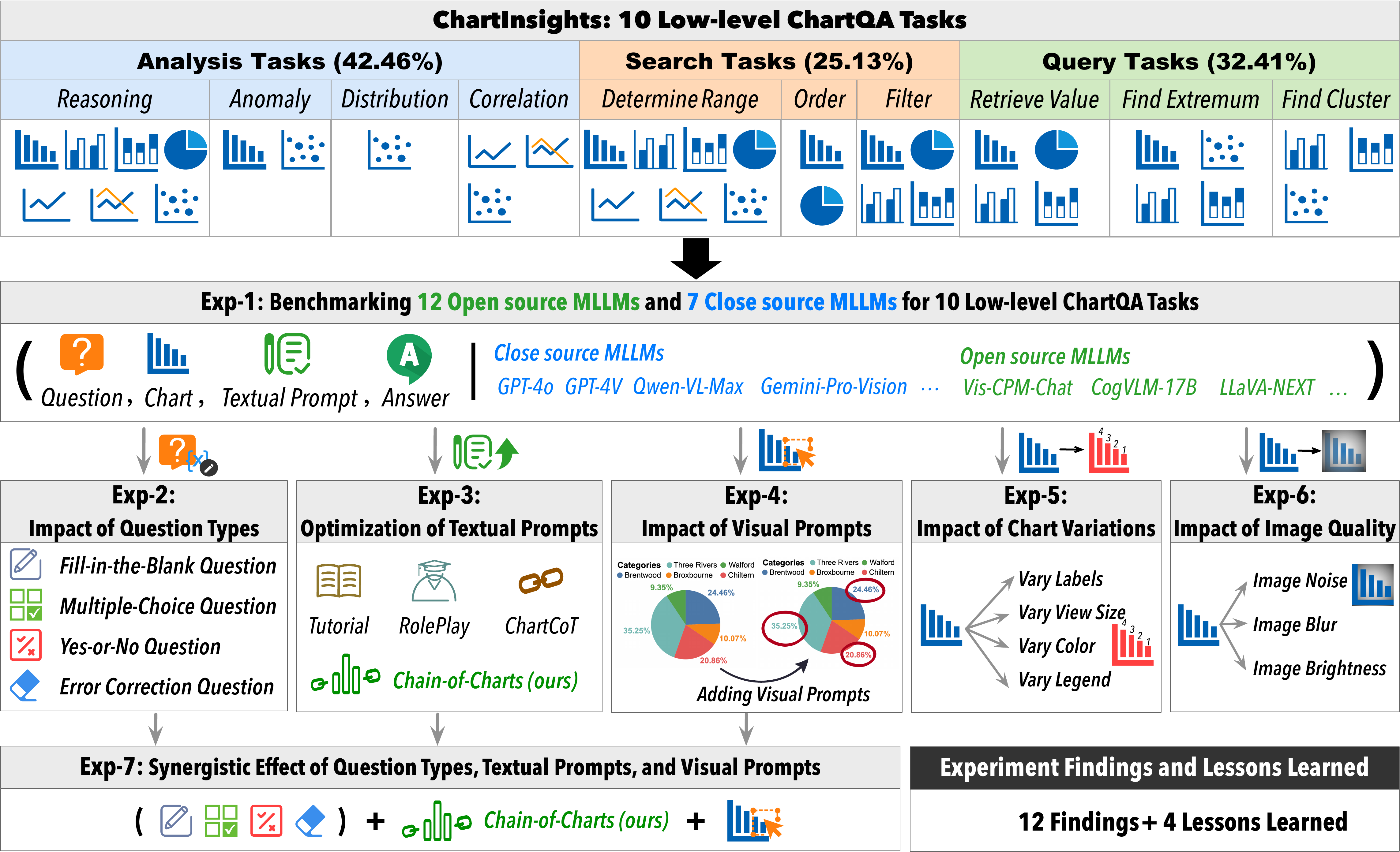}
	\caption{An Overview of Experimental Settings}
	\label{fig:teaser}
        \vspace{-.5em}
\end{figure*}

As shown in Figure~\ref{fig:teaser}, we utilize our ChartInsights to systematically evaluate the effectiveness of \mllms in 10 low-level ChartQA from different angles. The experiments are designed as follows:

\stab \textbf{Exp-1} \textit{Benchmarking MLLMs}: We start by benchmarking the performance of widely used MLLMs across 10 low-level ChartQA tasks involving 7 different types of charts. This experiment establishes a baseline for understanding the capabilities and limitations of MLLMs in low-level ChartQA tasks.

\vspace{-.5em}
\stab \textbf{Exp-2} \textit{Impact of Question Types}: 
We analyze how different question types influence \mllms interactions, helping to identify which types elicit the most accurate and informative responses.

\vspace{-.5em}
\stab \textbf{Exp-3} \textit{Textual Prompt Strategies}: We investigate the effect of various textual prompt strategies, such as Chain-of-Thoughts, on \mllms performance. 

\vspace{-.5em}
\paragraph{Rethink.}
Prior evaluations on high-level ChartQA tasks~\cite{chartreader, enhanced_chart_understanding, LVLMS_understanding_charts, chartbench, chartllama} primarily focused on optimizing \textit{textual} prompts. Given that ChartQA involves both ``reading'' and ``understanding'' chart images, we conduct an in-depth exploration of the impact of \textit{visual modification} on charts and prompts.

\vspace{-.5em}
\stab \textbf{Exp-4} \textit{Impact of Visual Prompts}: 
We conduct an in-depth exploration of the impact of visual prompts on \mllms performance to understand how guiding the \mllms' attention to specific visual elements can enhance its analytical capabilities.

\stab \textbf{Exp-5} \textit{Impact of Chart Variations}: 
We vary chart elements to analyze how changes in color schemes, view sizes, and legends affect the performance.

\stab \textbf{Exp-6} \textit{Impact of Image Quality}: 
We evaluate the effect of image quality by introducing various levels of image perturbations, such as noise and resolution changes, to understand the robustness of MLLMs in handling low-quality charts.

\stab \textbf{Exp-7} \textit{Synergistic Effects of Different Strategies}: 
Finally, we explore the synergistic effects of combining different question types, textual prompts, and visual prompts to enhance the overall performance of MLLMs in low-level ChartQA tasks.

Next, we discuss the main results and findings in Section~\ref{sec:expr_results} and lessons learned in Appendix~\ref{app:lesson learned}.

 
\begin{table*}[t!]
	\centering
1	\caption{Performance of 19 MLLMs across 10 Low-level ChartQA Tasks.}
	\label{tab:MLLMs_task_results}
    \resizebox{\linewidth}{!}{
	\begin{tabular}{l|c@{\hspace{6pt}}c@{\hspace{6pt}}c@{\hspace{6pt}}c|
			c@{\hspace{6pt}}c@{\hspace{6pt}}c|
			c@{\hspace{6pt}}c@{\hspace{6pt}}c|
			c}
		\toprule
		\multirow{2}{*}{Models} & \multicolumn{4}{c|}{Analysis} & \multicolumn{3}{c|}{Search} & \multicolumn{3}{c|}{Query} & \multirow{2}{*}{Overall ($\%$)} \\
		\cmidrule(lr){2-5} \cmidrule(lr){6-8} \cmidrule(lr){9-11}
		& \begin{tabular}[c]{@{}c@{}}Reasoning\end{tabular} & \begin{tabular}[c]{@{}c@{}}Anomaly\end{tabular} & \begin{tabular}[c]{@{}c@{}}Distribution\end{tabular} & \begin{tabular}[c]{@{}c@{}}Correlation\end{tabular} & \begin{tabular}[c]{@{}c@{}}Range\end{tabular} & \begin{tabular}[c]{@{}c@{}}Order\end{tabular} & \begin{tabular}[c]{@{}c@{}}Filter\end{tabular} & \begin{tabular}[c]{@{}c@{}}Retrieval\end{tabular} & \begin{tabular}[c]{@{}c@{}}Extremum\end{tabular} & \begin{tabular}[c]{@{}c@{}}Cluster\end{tabular} & \\
		\midrule
		\multicolumn{12}{l}{\hfill {Open Source MLLMs}} \\
		\midrule
            VisCPM-Chat-v1.1 ~\cite{viscpm} & 28.4 &\textbf{46.1} & 33.3 & 51.9 & 23.0 & 6.4 & 25.1 & 15.8 & 32.0 & 29.6 & 26.2 \\
            BLIP2 ~\cite{blip2}& 24.8 & 23.4 & 25.0 & 15.1 & 25.3 & 20.2 & 39.8 & 27.8 & 30.3 & 30.1 & 28.3 \\
            CogVLM-17B ~\cite{cogvlm}& 20.3 & 23.1 & 43.6 & 29.6 & \textbf{37.7} & 10.8 & 9.1 & 37.9 & 56.6 & 26.7 & 29.4 \\
            OmniLMM-12B ~\cite{Omnilmm}& 24.7 & 19.9 & 27.0 & 34.9 & 35.7 & 28.3 & 30.0 & 33.0 & 39.9 & 33.1 & 31.1 \\
            LLaVA1.5 ~\cite{llava1.5} & 32.4 & 6.3 & 30.9 & 23.1 & 21.7 & 32.7 & 35.6 & 32.6 & 35.8 & 43.5 & 32.2 \\
            ChartAssistant ~\cite{chartassisstant}& 24.6 & 27.7 & 35.8 & 28.1 & 30.5 & 22.5 & 14.7 & 39.4 & 63.0 & 26.4 & 32.4 \\
            MiniCPM-v2 ~\cite{minicpm} & 19.5 & 55.1 & 33.3 & 56.5 & 24.9 & 16.7 & 36.3 & 37.9 & 52.4 & 32.0 & 33.0 \\
            mPLUG-Owl2 ~\cite{mplugowl2}& 31.0 & 27.0 & 29.4 & 35.3 & 28.4 & 22.5 & 40.3 & 30.9 & 41.1 & 27.3 & 33.3 \\
            Qwen-VL-Chat ~\cite{qwenvl}& 27.8 & 36.3 & \textbf{45.1} & \textbf{55.8} & 33.8 & 20.0 & 28.7 & 31.3 & 50.2 & 27.1 & 33.4 \\
            ViP-LLaVA ~\cite{vipllava}& 28.8 & 6.6 & 34.8 & 30.3 & 21.9 & \textbf{35.8} & \textbf{40.4} & 42.2 & 38.3 & 33.8 & 33.8 \\
            LLaVA-NEXT ~\cite{llavanext} & \textbf{30.6} & 7.4 & 26.5 & 38.0 & 29.5 & 33.3 & 23.4 & \textbf{53.5} & 59.8 & \textbf{52.3} & 38.5 \\
            Sphinx-v2 ~\cite{sphinx} & 30.0 & 28.9 & 37.8 & 36.1 & 25.8 & 23.5 & 36.7 & 49.7 & \textbf{66.3} & 45.3 & \textbf{40.2} \\

		\midrule
		\multicolumn{12}{l}{\hfill {Closed Source MLLMs}} \\
		\midrule
            Qwen-VL-Plus~\cite{qwenvl} & 30.8 & 27.3 & 47.1 & 47.1 & 43.0 & 34.6 & 20.7 & 58.7 & 65.5 & 62.5 & 42.6 \\
            Gemini-Pro-Vision ~\cite{gemini}& 25.6 & 30.1 & 45.6 & 58.7 & 75.3 & 32.9 & 30.1 & 60.4 & 80.9 & 55.3 & 48.4 \\
            ChatGLM-4V ~\cite{GLM}& 34.1 & 28.9 & 39.2 & 42.3 & 55.5 & 18.9 & 43.4 & 58.1 & 69.3 & 71.4 & 48.4 \\
         Claude3-Haiku ~\cite{claude3}& 33.0 & 9.0 & 42.7 & 46.2 & 60.4 & 26.2 & 40.0 & 62.3 & 75.1 & 66.8 & 49.5 \\
            Qwen-VL-Max \cite{qwenvl} & 28.8 & 25.8 & 62.3 & 63.0 & 66.1 & 40.2 & 38.9 & 67.0 & 79.6 & 66.8 & 51.7 \\
            GPT-4V ~\cite{gpt4v}& 35.2 & 19.5 & 53.4 & 59.6 & 70.0 & 41.9 & 44.3 & 67.6 & 88.7 & 72.9 & 56.1 \\
            GPT-4o ~\cite{openai2024gpt4}& \textbf{55.9} & \textbf{34.0} & \textbf{70.1} & \textbf{68.5} & \textbf{80.6} & \textbf{68.9} & \textbf{49.9} & \textbf{82.6} & \textbf{93.9} & \textbf{74.3} & \textbf{69.2} \\

		\bottomrule
	\end{tabular}
}
\end{table*}

\begin{table*}[t!]
	\centering
	\caption{Performance of 19 MLLMs across 7 Chart Types and 4 Question Types. (FB: Fill-in-the-Blank Question; MC: Multiple-Choice Question; YN: Yes-or-No Question; EC: Error Correction Question)}
 	\label{tab:MLLMs_chart_results}
 	\vspace{-.5em}
    \resizebox{\linewidth}{!}{
	\begin{tabular}{l|ccccccc|cccc}
		\toprule
		\multirow{2}{*}{Models} & \multicolumn{7}{c|}{Chart Types} & \multicolumn{4}{c}{Question Types}  \\
		\cmidrule(lr){2-8} \cmidrule(lr){9-12}
		& Grouped Bar & Stacked Bar & Grouped Line & Basic Bar & Basic Line & Scatter Plot & Pie & FB & MC & YN & EC \\
		\midrule
		\multicolumn{12}{l}{\hfill {Open Source MLLMs}} \\
		\midrule
            VisCPM-Chat-v1.1 ~\cite{viscpm} & 24.8 & 21.5 & 24.4 & 30.3 & 25.3 & \textbf{34.9} & 20.4 & 7.4 & 39.8 & 49.3 & 8.3 \\
            BLIP2 ~\cite{blip2} & 32.2 & 31.1 & 24.7 & 34.2 & 15.0 & 18.2 & 8.7 & 3.1 & 46.9 & 57.7 & 5.5 \\
            CogVLM-17B ~\cite{cogvlm} & 26.2 & 24.2 & \textbf{32.1} & 40.1 & 35.3 & 30.2 & 19.6 & 27.6 & 49.7 & 31.3 & 9.2 \\
            OmniLMM-12B ~\cite{Omnilmm} & 28.8 & 25.9 & 26.5 & 37.8 & 43.4 & 27.5 & 34.4 & 10.1 & 46.3 & 57.0 & 10.8 \\
            LLaVA1.5 ~\cite{llava1.5} & 31.3 & 30.5 & 22.6 & 35.7 & 40.6 & 26.8 & 36.3 & 9.6 & 41.4 & 67.9 & 9.9 \\
            ChartAssistant ~\cite{chartassisstant} & 32.9 & 27.8 & 24.1 & 41.2 & 35.6 & 28.2 & 23.3 & \textbf{30.2} & 46.3 & 40.9 & 12.2 \\
            MiniCPM-v2 ~\cite{minicpm} & 33.4 & 28.7 & 18.8 & 40.4 & 36.3 & 29.9 & 28.0 & 21.0 & 42.0 & 60.8 & 8.3 \\
            mPLUG-Owl2 ~\cite{mplugowl2} & 32.4 & 31.8 & 29.2 & 37.0 & 38.4 & 29.2 & 34.3 & 13.2 & 49.9 & 54.6 & 15.5 \\
            Qwen-VL-Chat ~\cite{qwenvl} & 31.2 & 26.3 & 25.9 & \textbf{39.3} & 42.8 & 38.5 & 34.7 & 22.5 & 52.3 & 48.6 & 10.4 \\
            ViP-LLaVA ~\cite{vipllava} & 32.2 & 31.2 & 16.1 & 39.8 & 34.4 & 28.1 & 39.1 & 7.9 & 40.1 & \textbf{73.4} & 13.7 \\
            LLaVA-NEXT ~\cite{llavanext} & 37.1 & 33.7 & 18.8 & 50.4 & \textbf{48.4} & 27.4 & 37.1 & 23.1 & 42.2 & 63.2 & \textbf{25.6} \\
            Sphinx-v2 ~\cite{sphinx} & \textbf{39.4} & \textbf{35.7} & 26.5 & \textbf{51.2} & 42.5 & 31.5 & 36.7 & 28.2 & \textbf{56.1} & 60.0 & 16.7 \\

		\midrule
		\multicolumn{12}{l}{\hfill {Closed Source MLLMs}} \\
		\midrule
            Qwen-VL-Plus~\cite{qwenvl} & 36.5 & 35.8 & 26.2 & 57.0 & 32.8 & 44.2 & 42.5 & 32.9 & 59.2 & 54.0 & 24.2 \\
            Gemini-Pro-Vision ~\cite{gemini} & 45.0 & 43.3 & 35.1 & 57.7 & 43.8 & 47.3 & 51.2 & 40.9 & 55.4 & 43.9 & 53.3 \\
            ChatGLM-4V ~\cite{GLM} & 46.0 & 43.9 & 36.9 & 58.7 & 56.3 & 39.7 & 49.9 & 32.1 & 53.7 & 72.2 & 35.8 \\
            Claude3-Haiku ~\cite{claude3} & 50.7 & 48.1 & 32.1 & 55.6 & 46.9 & 40.4 & 48.3 & 39.4 & 51.2 & 61.8 & 45.9 \\
            Qwen-VL-Max \cite{qwenvl} & 47.1 & 46.3 & 38.4 & 66.4 & 43.1 & 47.7 & 48.1 & 44.2 & 75.3 & 48.1 & 39.0 \\
            GPT-4V ~\cite{gpt4v} & 52.0 & 48.2 & 47.3 & 67.2 & 49.4 & 62.7 & 53.6 & 44.1 & 64.4 & 66.4 & 49.7 \\
            GPT-4o ~\cite{openai2024gpt4} & \textbf{62.7} & \textbf{59.8} & \textbf{53.9} & \textbf{84.1} & \textbf{56.9} & \textbf{74.6} & \textbf{70.9} & \textbf{61.0} & \textbf{77.8} & \textbf{74.8} & \textbf{63.1} \\

		\bottomrule
	\end{tabular}
    }
\end{table*}


\subsection{Experimental Results and Findings}
\label{sec:expr_results}

\subsection*{\mbox{\ding{117} Exp-1: Evaluation and Benchmarking}}
\label{sec:compare}
\paragraph{Experimental Settings.}
We evaluate 19 widely used models (see Table~\ref{tab:MLLMs_task_results}) from both academia and industry, including 12 open-source and 7 closed-source MLLMs. 
Inspired by existing evaluation strategies~\cite{mathvista,mmeBenchmark}, we randomly selected 20\% of the ChartInsights dataset as our test set to reduce testing costs while ensuring the reliability of experimental results.

The test set comprises 400 charts, spanning 10 low-level analysis tasks and encompassing 4,388 (chart, task, query, and answer) samples. This sizable validation dataset allows for a comprehensive evaluation of MLLMs' capabilities in low-level ChartQA tasks. Then, we analyze the answers of \mllms, compare them with Ground Truth, and calculate the accuracy.
For each model, we evaluate through four question types in textual prompts, \ie Fill-in-the-Blank, Multiple-Choice, Yes-or-No, and Error Correction questions.


\paragraph{Overall Results.}
Table~\ref{tab:MLLMs_task_results} presents the performance of the 19 models across 10 low-level tasks, while Table~\ref{tab:MLLMs_chart_results} showcases their performance across 7 chart types and 4 question types. 
The findings indicate that closed-source models outperform open-source models by a significant margin among the 19 models evaluated. Notably, GPT-4o demonstrates exceptional performance across all tasks.

\vspace{.5em}
 \noindent\fbox{%
 	\parbox{0.97\columnwidth}{%
 		\textbf{Finding-1}: \textit{Closed-Source models exhibit far superior generalization performance in low-level analysis tasks compared to open-source models. }
 	}%
 }
 \vspace{.25em}

In addition, our analysis of open-source models reveals that the ability to comprehend low-level ChartQA tasks is not directly proportional to the number of model parameters. For example, MiniCPM (with 2B parameters) outperforms larger models like CogVLM (with 17B parameters). This suggests that factors beyond scale significantly influence a model's capacity.


\vspace{.25em}
\noindent\fbox{%
\parbox{0.97\columnwidth}{%
    \textbf{Finding-2}: \textit{
The ability of open-source models to understand low-level charts is not directly proportional to their number of model parameters.
}
}%
}
\vspace{.25em}

Moreover, Table~\ref{tab:MLLMs_chart_results} shows that although some open-source models perform poorly overall, they have high accuracy on Yes-or-No tasks, with ViP-LLaVA's accuracy even being close to that of GPT-4o. We analyze the recall rates of ViP-LLaVA and GPT-4o on Yes-or-No tasks based on the confusion matrix and found that ViP-LLaVA's recall rate for the ``No'' label is only 25\%, which is quite different from the final accuracy of 73.4\%; while GPT-4o's recall rate for the ``No'' label is 72\%, which is relatively close to the final accuracy of 74.8\%. We believe that open-source models like ViP-LLaVA have a certain tendency towards the ``No'' label, and it is because there are more ``No'' labels in our data that these open-source models have a high accuracy on Yes-or-No questions.

\vspace{.5em}
\noindent\fbox{%
\parbox{0.97\columnwidth}{%
    \textbf{Finding-3}: \textit{We find open-source models like ViP-LLaVA show higher accuracy on Yes-or-No questions because of a possible bias towards ``No'' labels.
}
}%
}
\vspace{.25em}
 

\paragraph{GPT-4o as a representative research object.}
Since GPT-4o significantly outperforms other MLLMs, we utilize it as a representative MLLM to systematically evaluate the effectiveness of low-level ChartQA tasks under different scenarios.
Specifically, we will conduct a series of in-depth evaluations (\ie \textbf{Exp-3} to \textbf{Exp-7}) using GPT-4o. 

\begin{figure*}[t!]
	\centering
	\includegraphics[width=\textwidth]{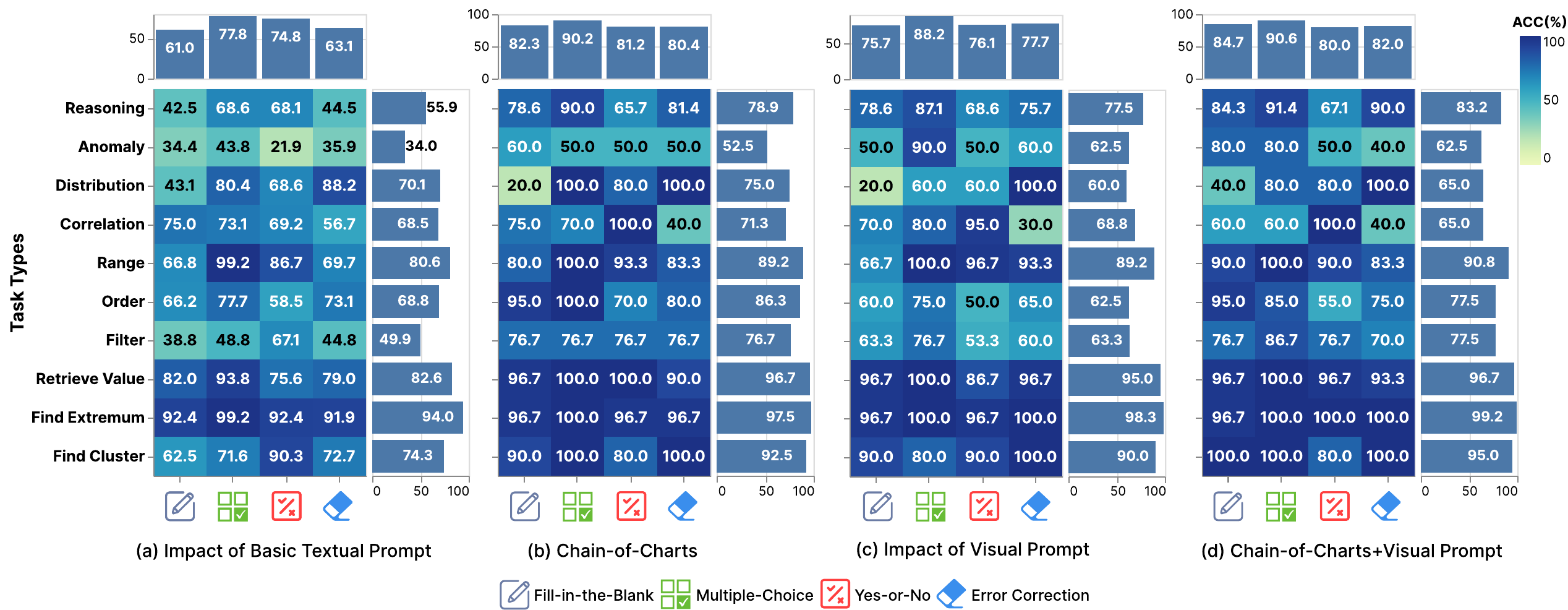}
	\caption{The Effectiveness of GPT-4o across 10 Low-level Tasks and 4 Question Types}
	\label{fig:overall}
        \vspace{-.5em}
        
\end{figure*}

\paragraph{Overall Results of GTP-4o.}
The overall result of  GPT-4o is shown in Figure~\ref{fig:overall}.
These heatmaps visualize the performance of GPT-4o on various low-level ChartQA tasks under different prompt conditions. The progression from subfigures (a) to (d) clearly indicates the incremental benefits of incorporating visual prompts, optimization strategies, and their combination, culminating in the most effective approach for improving GPT-4o's performance in low-level ChartQA tasks.

\begin{figure}[t!]
	\centering	
	\includegraphics[width=\columnwidth]{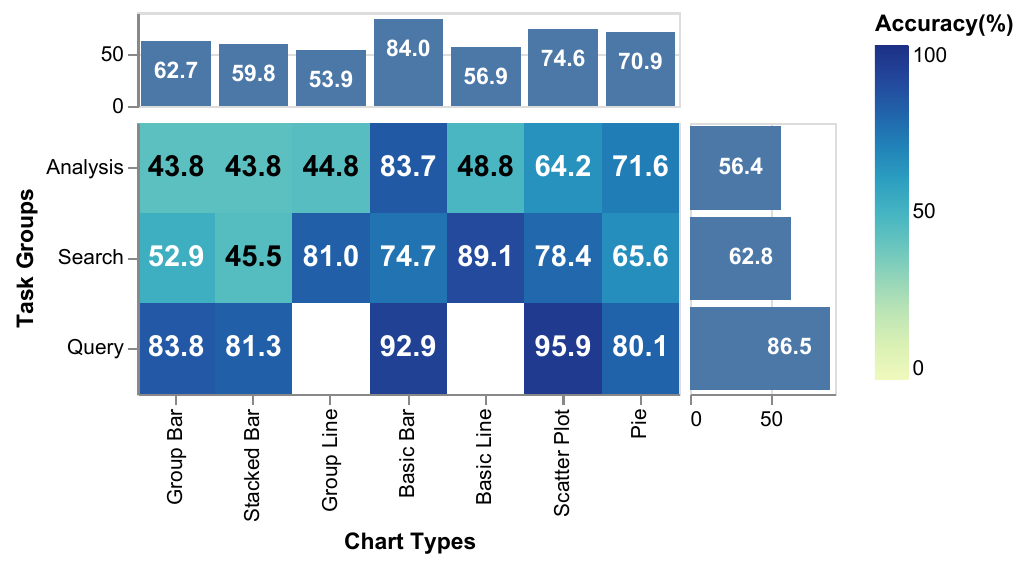}
	\caption{Effectiveness of GPT-4o}
	\label{fig:heatmap_evalres}
        \vspace{-1em}
\end{figure}

As depicted in the bar chart at the top of Figure~\ref{fig:heatmap_evalres}, GPT-4o's accuracy reaches near 90\% for basic bar charts, but hovers around 66\% for similar tasks involving other chart types. This performance gap suggests that, despite the dataset containing a significant number of reasoning tasks that are generally straightforward for humans, the effectiveness of GPT-4o in ChartQA tasks has not yet reached that of the average human.

 \vspace{.5em}
 \noindent\fbox{%
	\parbox{0.97\columnwidth}{%
		\textbf{Finding-4}: \textit{
			The performance of GPT-4o declines as task complexity increases, mirroring human performance, but it has not yet matched the analytical capabilities of average humans.
		}
	}%
}
 \vspace{.25em}


\subsubsection*{\ding{117} Exp-2: Impact of Question Types}
\label{sub:expbaseline}
The experimental settings are the same as in Exp-1.

\paragraph{The Effectiveness of Question Types.}
Figure~\ref{fig:overall} (a) provides an overview of GPT-4o's performance across 10 low-level tasks using four different textual prompts. Notably, GPT-4o achieves the highest overall accuracy of 77.8\% in ``Multiple-Choice'' questions. It also demonstrates a strong performance in ``Yes or No'' questions, with an accuracy of 74.8\%. Comparatively, GPT-4o performs relatively better in ``Multiple-Choice'' and ``Yes-or-No'' questions compared to ``Fill-in-the-Blank'' and ``Error Correction'' questions. The former prompt types inherently provide GPT-4o with answer options to select or evaluate, while the latter two require direct answer generation from GPT-4o.

 \vspace{.5em}
\noindent\fbox{%
	\parbox{0.97\columnwidth}{%
		\textbf{Finding-5}:
		\textit{Structured textual prompts and candidate answers significantly enhance GPT-4o's ability to reason out correct responses.}
	}%
}
 \vspace{.25em}

\subsubsection*{\ding{117} Exp-3: Optimization of Textual Prompts}


This experiment will examine the influence of commonly used textual prompt optimization strategies on MLLMs, including RolePlay~\cite{roleplay}, Tutorial, and ChartCoT~\cite{chartbench}, which is based on the Chain-of-Thought (CoT) prompt strategy~\cite{chainofthought}.
The Chain-of-Thought (CoT) prompt strategy has proven effective in various scenarios~\cite{chainofthought}. 
CoT aims to guide the model by mimicking the step-by-step reasoning process humans use to solve problems.
Recently, Xu et al.~\cite{chartbench} implemented the CoT strategy for ChartQA tasks, namely ChartCoT~\cite{chartbench}. ChartCoT poses a series of questions to guide the model in understanding the chart's details before formulating an answer. However, ChartCoT faces challenges in ensuring the accuracy of GPT-4o's responses to guiding questions, especially with complex charts.

\paragraph{\ttextp Prompts.} 
Therefore, we introduce a novel prompting strategy, termed \textp, which builds based on the Chain-of-Thoughts approach~\cite{chainofthought}, as shown in Appendix~\ref{app:chain_of_charts_prompt}. The core of \textp lies in orchestrating a sequence of questions and their corresponding answers (($q_1$, $a_1$), ($q_2$, $a_2$),...($q_m$, $a_m$)) to progressively guide the model towards a deeper understanding of the chart's details, thereby enhancing its ability to formulate accurate responses.


\paragraph{Experimental Settings.}
Shanahan et al.~\cite{roleplay} suggested that giving large models specific roles could enhance their performance on particular tasks. Inspired by this, we assigned GPT-4o the role of a visualization expert to see if it would enhance its performance. 
Our observations showed that more detailed prompts resulted in more precise and accurate responses from GPT-4o. As a result, we created a detailed ChartQA tutorial called the Tutorial prompt. Examples of these two prompts can be found in Appendix~\ref{app:roleplayprompt} and Appendix~\ref{app:tutorialprompt} respectively.



\paragraph{Overall Results.} 
Figure~\ref{fig:overall}(b) reports the performance of GPT-4o with \ttextp.
Compared with Figure~\ref{fig:overall}(a), we can observe a significant enhancement in GPT-4o's capabilities across 10 tasks and four question types.
Table~\ref{tab:text_and_visual_p_effect}(a) shows the overall accuracy of GPT-4o on 10 low-level tasks under 5 Textual Prompt strategies.
Overall, Chain-of-Charts leads in average accuracy across all tasks with 83.5\%, outperforming ChartCoT's accuracy of 76.1\% by 6.9\%.
Specifically, Chain-of-Charts achieves the highest accuracy on five tasks including Reasoning, Determine Range, Order, Filter, Retrieve Value, Find Extremumm and Find Cluster, with accuracy of 78.9\%, 89.2\%, 86.3\%, 76.7\%, 96.7\%, 97.5\% and 92.5\%.
These tasks demand precise reasoning from GPT-4o, based on accurate identification of element coordinates and values. The Chain-of-Charts prompt framework effectively provides GPT-4o with the correct value and coordinate references, significantly aiding in the accurate positioning of different elements.

 \vspace{.25em}
\noindent\fbox{%
	\parbox{0.97\columnwidth}{%
		\textbf{Finding-6}:
		\textit{\ttextp supplies GPT-4o with accurate chart reference information,  enhancing the model's comprehension and detailed reasoning of chart structures and elements.}
	}%
}
 \vspace{.25em}

\subsubsection*{\ding{117} Exp-4: Impact of Visual Prompts}



\paragraph{Experimental Settings.}
 In this experiment, we design three types of visual prompts based on graphical overlay strategies~\cite{overlays}, namely, handwriting, regular shape, and special design, as shown in Figure~\ref{fig:visualprompts}. The design of visual prompts are detailed in Appendix~\ref{app:visual_prompts}. Specifically, we generate 255 visual prompts for 35 charts as the testing samples. These 255 visual prompts are associated with 1,020 test samples and cover 10 low-level tasks. Please refer to {\bf Step-5} in the Appendix~\ref{subsec:construct} for details.

\paragraph{Overall Results.} 
Figure~\ref{fig:overall}(c) demonstrates the strong performance of GPT-4o with visual prompts across 10 tasks and 4 question types.
Additionally, Figure~\ref{fig:gpt_4oradar}(a)-(c) showcases the performance of visual prompts under different textual prompts, low-level ChartQA tasks, and chart types.
In general, visual prompts enhance GPT-4o's performance. Particularly, Figure ~\ref{fig:gpt_4oradar}(b) reveals significant improvements in \textit{Reasoning} and \textit{Anomaly Detection} tasks, indicating the model's ability to accurately analyze and reason with relevant data.

 \vspace{.5em}
\noindent\fbox{%
	\parbox{0.97\columnwidth}{%
		\textbf{Finding-7}:
		\textit{Visual prompts greatly improve GPT-4o's performance, showing the value of visual information for comprehension and reasoning.}
	}%
}
 \vspace{.25em}

However, GPT-4o does not exhibit significant benefits from visual prompts in \textit{Correlation} and \textit{Order} tasks. These tasks often challenge GPT-4o to discern complex relationships among more than three distinct elements. In such scenarios, visual prompts may lose their specificity and lead to confusion due to the introduction of multiple new visual elements, especially in tasks like Order, where the added visual information can be misleading.

 \vspace{.5em}
\noindent\fbox{%
	\parbox{0.97\columnwidth}{%
		\textbf{Finding-8}:\textit{
			Different tasks require tailored visual prompts for effective chart comprehension. Using the same style of visual prompts across tasks can have a negative impact on certain tasks.
	}}%
}
\vspace{.25em}

\subsubsection*{\ding{117} Exp-5: Impact of Chart Variations}

\paragraph{Experimental Settings.}
A chart can exhibit visual differences by varying its constituent elements (e.g., data labels), as illustrated in Appendix~\ref{subsec:construct} and Figure~\ref{fig:vary_chart_grammar}(a). Intuitively, these visual variations may influence GPT-4o's performance on low-level ChartQA tasks.
To investigate this hypothesis, we explore how varying chart elements affect the performance of GPT-4o.
To this end, we develop 356 visual variants of 35 base charts as test samples. These variants are associated with 17,972 textual prompts, spanning 10 low-level tasks.

\paragraph{Overall Results.} 
\begin{figure*}[t!]
	\centering
	\includegraphics[width=\textwidth]{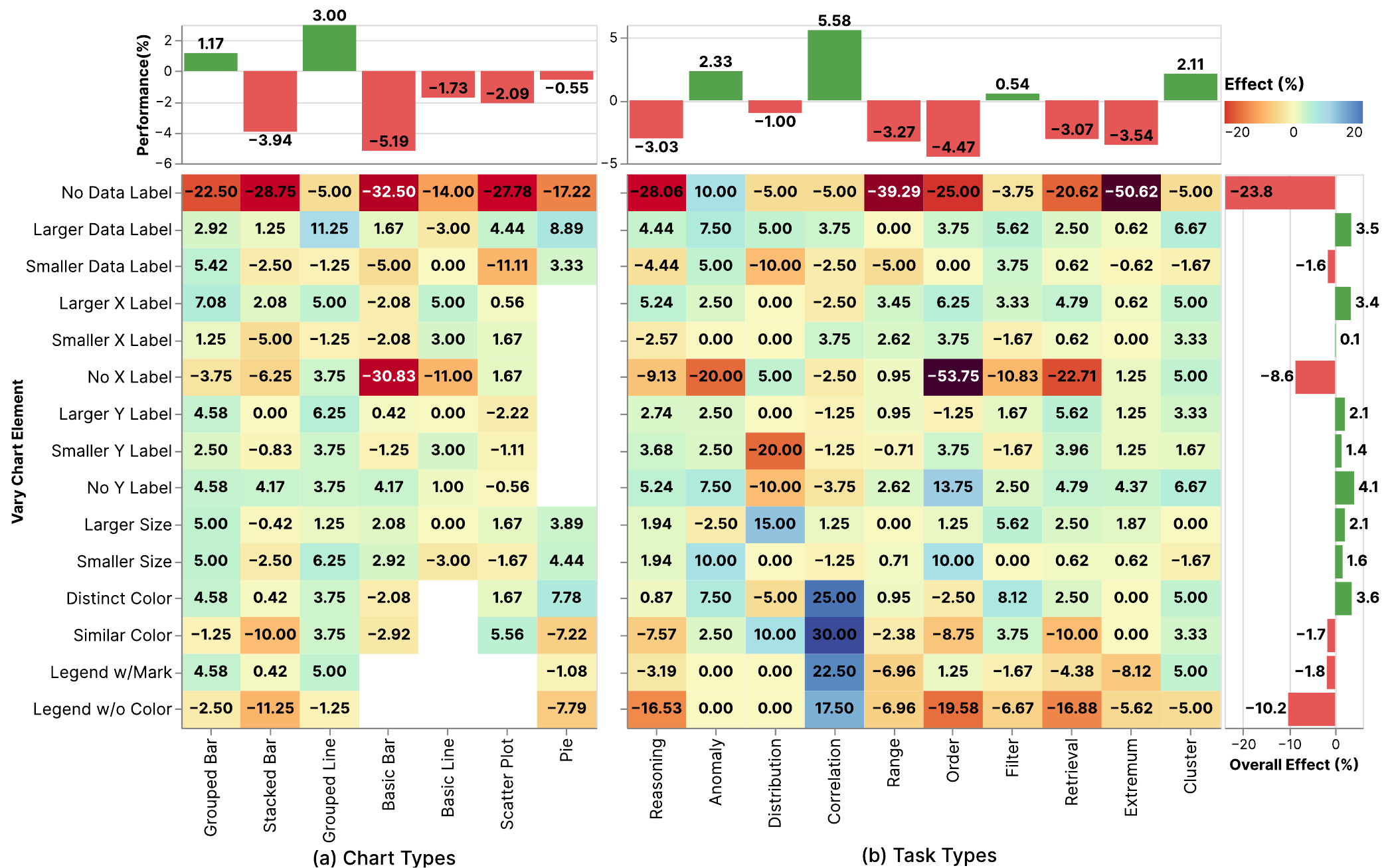}
	\caption{The Impact of Varying  Chart Elements on GPT-4o's Performance.}
	\label{fig:vary_element}
        \vspace{-.5em}
\end{figure*}

Overall, Figure~\ref{fig:vary_element}(a) indicates that most chart variants have a minor negative impact on GPT-4o's performance. 
However, the absence of data labels significantly impairs its performance across seven chart types (Figure~\ref{fig:vary_element}(a)). This is understandable, as data labels aid GPT-4o in comprehending the underlying insights conveyed by the charts. Interestingly, GPT-4o shows a performance boost of 17.5\% in anomaly detection and 5.5\% in filtering tasks when data labels are not present (Figure~\ref{fig:vary_element}(b)). This implies that data labels may sometimes hinder GPT-4o's ability to identify anomalies and filter values effectively.

Figure~\ref{fig:vary_element}(b) reveals that certain chart variants, such as larger x/y/data labels, positively impact GPT-4o's performance in tasks like anomaly detection, filtering, and clustering. These tasks inherently involve comparisons between elements. We hypothesize that alterations in chart elements can shift GPT-4o's focus towards visual comparisons rather than numerical ones.

The bar chart on the right in Figure~\ref{fig:vary_element} illustrates the varied impacts of 15 chart variants on GPT-4o's performance. We posit that data labels play a crucial role in GPT-4o's low-level data analysis capabilities, as removing or reducing them tends to diminish its effectiveness. Furthermore, adding marks to the legend or eliminating the legend's color significantly affects GPT-4o's performance on specific tasks by introducing visual clutter or removing essential visual cues, respectively. For instance, changes in color and legend can greatly assist GPT-4o in solving Correlation tasks.

 \vspace{.5em}
 \noindent\fbox{%
	\parbox{0.97\columnwidth}{%
		\textbf{Finding-9}:
		\textit{While most chart variants have a minimal impact on GPT-4o's performance, the absence of data labels significantly affects its accuracy. Additionally, larger labels and the removal of data labels can actually enhance GPT-4o's performance in anomaly detection and filtering tasks, as it shifts its focus to visual comparisons.
  }
	}%
}
\vspace{.25em}

\subsubsection*{\ding{117} Exp-6: Impact of Image Quality}


\paragraph{Experimental Settings.}
In this experiment, we evaluate the robustness of GPT-4o in low-level ChartQA tasks by introducing six types of noise, as shown in Figure~\ref{fig:vary_chart_grammar} (b). We use 245 visual variants of 35 charts as testing samples, along with 8,456 textual prompts covering 10 tasks.

\paragraph{Overall Results.}
Figure~\ref{fig:vary_quality} reports the experimental results. Generally, six methods of degrading image quality tend to negatively impact GPT-4o across a broad range of tasks and chart types. Among these, Median Blur stands out as the most detrimental, causing an average performance decline of 16.8\%.  We consider that median blurring makes numerical labels unreadable, resulting in a significant decrease in the performance of tasks directly related to numerical values.
Interestingly, both increasing and decreasing the brightness show positive effects on the majority of tasks, with an average improvement of 0.6\% and 1.3\% respectively.
In addition, Figure~\ref{fig:vary_quality}(b) reveals a unique finding where the distribution and cluster tasks did not experience any negative impact under six types of noise. In fact, the performance of the distribution task improved by an average of 9.17\%, while the cluster task improved by an average of 5.28\%.

 \vspace{.5em}
 \noindent\fbox{%
	\parbox{0.97\columnwidth}{%
		\textbf{Finding-10}:
		\textit{
  GPT-4o's accuracy varies under the influence of different types of noise on various charts. Interestingly, there are instances where the accuracy improves, particularly in visually semantic tasks. This suggests that GPT-4o can rely more on visual information when the textual information is compromised.
  }
	}%
}

\begin{table*}[t!]
	\centering
	\caption{GPT-4o: The Synergistic Effect of Visual and Textual Prompts (Overall Accuracy (\%))}
	\label{tab:text_and_visual_p_effect}
	\vspace{-.5em}
	\resizebox{\linewidth}{!}{
		\begin{tabular}{l|c@{\hspace{6pt}}c@{\hspace{6pt}}c@{\hspace{6pt}}c|
				c@{\hspace{6pt}}c@{\hspace{6pt}}c|
				c@{\hspace{6pt}}c@{\hspace{6pt}}c|
				c}
			\toprule
			\multirow{2}{*}{Prompts} & \multicolumn{4}{c|}{Analysis} & \multicolumn{3}{c|}{Search}& \multicolumn{3}{c|}{Query} & \multirow{2}{*}{Overall(\%)} \\
			\cmidrule(lr){2-5} \cmidrule(lr){6-8} \cmidrule(lr){9-11}
			& \begin{tabular}[c]{@{}c@{}}Reasoning\end{tabular} & \begin{tabular}[c]{@{}c@{}}Anomaly\end{tabular} & \begin{tabular}[c]{@{}c@{}}Distribution\end{tabular} & \begin{tabular}[c]{@{}c@{}}Correlation\end{tabular} & \begin{tabular}[c]{@{}c@{}}Range\end{tabular} & \begin{tabular}[c]{@{}c@{}}Order\end{tabular} & \begin{tabular}[c]{@{}c@{}}Filter\end{tabular} & \begin{tabular}[c]{@{}c@{}}Retrieval\end{tabular} & \begin{tabular}[c]{@{}c@{}}Extremum\end{tabular} & \begin{tabular}[c]{@{}c@{}}Cluster\end{tabular} & \\
			
			\midrule
			\multicolumn{12}{l}{\hfill {(a) The Effectiveness of Textual Prompts}} \\
			\midrule
			Basic Textual Prompts & 55.9 & 34.0 & 70.1 & 68.5 & 80.6 & 68.9 & 49.9 & 82.6 & 93.9 & 74.3 & 69.2\\
			ChartCoT & 75.7 & 72.5 & 65.0 & 63.7 & 80.0 & 62.5 & 60.0 & 89.2 & 92.5 & 87.5 &  76.1 \\
			Role-Play & 61.8 & 72.5 & 60.0 & 71.2 & 85.8 & 71.2 & 64.2 & 87.5 & 96.7 & 80.0 & 74.6\\
			Tutorial & 75.0 & \textbf{77.5} & \textbf{80.0} & \textbf{81.2} & 86.7 & 37.5 & 75.8 & 86.7 & 94.2 & 90.0  & 78.4\\
			
			\textbf{Chain-of-Charts (ours)} & \textbf{78.9} & 52.5 & 75.0 & 71.3 & \textbf{89.2} & \textbf{86.3} & \textbf{76.7} & \textbf{96.7} & \textbf{97.5} & \textbf{92.5} & \textbf{83.5}\\
			\midrule
			\multicolumn{12}{l}{\hfill {(b) Synergistic Effect of Visual and Textual Prompts}} \\
			\midrule
			Basic Textual Prompts & 77.5& 62.5 & 60.0 & 68.8 & 89.2 & 62.5 & 63.3 & 95.0 & 98.3 & 90.0 & 79.4 \\
			ChartCoT & 80.7 & 65.0 & 55.0 & 71.2 & 85.0 & 56.2 & 60.8 & 91.7 & 92.5 & 87.5 &78.0\\
			Role-Play & 77.9 & 70.0 & 55.0 & 75.0 & 90.8 & 55.0 & 61.7 & 94.2 & 99.2 & 85.0 &79.4\\
			Tutorial & 80.0 & \textbf{80.0} & \textbf{75.0} & \textbf{85.0} & \textbf{92.5} & 32.5 & 75.8 & 90.8 & 99.2 & 92.5 & 81.6\\
			\textbf{Chain-of-Charts (ours)} & \textbf{83.2} & 62.5 & 65.0 & 65.0 & 90.8 & \textbf{77.5} & \textbf{77.5} & \textbf{96.7} & \textbf{99.2} & \textbf{95.0} & \textbf{84.3}\\
			
			\bottomrule
		\end{tabular}
	}
\end{table*}

\subsubsection*{\ding{117} Exp-7:  Synergistic Effect of Question Types, Textual Prompts, and Visual Prompts}

\paragraph{Experimental Settings.}
As previously mentioned, using Visual Prompts alone cannot robustly improve the accuracy of GPT-4o. By comparing the bar charts on the right side of Figure~\ref{fig:overall}(a) and (c), we can see that the accuracy of GPT-4o using Visual Prompts actually decreases in Distribution and Order tasks. In contrast, by comparing the bar charts on the right side of Figure~\ref{fig:overall}(a) and (c), we find that under the influence of Chain-of-Charts, GPT-4o has made significant improvements in all 10 tasks. 
Therefore, we ask: \textit{Can the combination of visual and text prompts enhance performance in low-level ChartQA tasks with GPT-4o?}
We use the same samples as in Exp-4 for this experiment.

\paragraph{Overall Results.}
Figure~\ref{fig:overall}(d) and shows GPT-4o's accuracy following the integration of Chain-of-Charts and Visual Prompt, demonstrating a clear enhancement over the outcomes depicted in Figures~\ref{fig:overall}(a), (b), and (c), which demonstrates the combined strategy's effectiveness.

Table~\ref{tab:text_and_visual_p_effect} reports the performance improvements in GPT-4o after integrating various textual prompts with visual prompts.  Furthermore, this combination attained the highest accuracy in six tasks.

\vspace{.5em}
\noindent\fbox{%
	\parbox{0.97\columnwidth}{%
		\textbf{Finding-11}:\textit{
Combining visual prompts with the Chain-of-Charts strategy significantly improves the performance, suggesting that integrating multiple types of prompts can leverage their respective strengths.
	}}%
}

\paragraph{Discussions about Textual and Visual Prompts.}
As depicted in Figure~\ref{fig:gpt_4oradar}, the integration of Chain-of-Charts and visual prompts enables GPT-4o to outperform other settings. However, the improvement over using Chain-of-Charts alone is slight. We discuss the possible reasons behind: 

First, after carefully analyzing the experimental results, we discover that GPT-4o exhibits a certain degree of hallucination in chart understanding. For example, even if the calculation process is accurate, GPT-4o may provide answers that do not match any of the multiple-choice options, leading to incorrect results. This indicates that the model's accuracy is significantly affected by hallucination. Moreover, we also observe that GPT-4o might output numerical information unrelated to the chart even when explicitly recognizing values, further evidencing the hallucination phenomenon in chart reading.



\vspace{.25em}
\noindent\fbox{%
	\parbox{0.97\columnwidth}{%
		\textbf{Finding-12}:\textit{
    Adding a Visual Prompt improves performance, but its impact is limited when applied to the Chain-of-Charts strategy.
	}}%
}


%% file: secs/conclusion.tex
\section{Related Work}
\label{related_work_in_paper}

\vspace{-.5em}
\paragraph{Low-Level ChartQA Tasks.}
Low-level data analysis tasks in chart involve activities such as data retrieval and correlation determination. These tasks were defined by~\citet{DBLP:conf/infovis/AmarES05} and later evaluated by~\citet{lowleveltasks} in a crowdsourced experiment. In this paper, we use these ten tasks as a framework to assess the effectiveness of \mllms in low-level ChartQA.



\paragraph{Evaluating \mllms in ChartQA Tasks.}
Recent studies~\cite{chartreader, enhanced_chart_understanding, LVLMS_understanding_charts, chartbench, chartllama, chartx, unichart} have leveraged MLLMs to perform high-level ChartQA tasks, such as chart captioning.
For example, \citet{LVLMS_understanding_charts} evaluated the capabilities of representative \mllms, such as \gptfv and Gemini~\cite{gemini}, on chart captioning tasks, finding challenges in accurately reflecting factual chart information.
Diverging from this focus on high-level tasks, our research uniquely targets low-level ChartQA tasks~\cite{lowleveltasks}. 

\paragraph{ChartQA Datasets.} 
In the last decade, several ChartQA datasets have been presented, as shown in Table~\ref{Datasets_comparison}.
For example, ChartBench~\cite{chartbench} includes 2.1K charts for four types of ChartQA tasks.
However, a gap remains - no existing dataset comprehensively evaluates the 10 critical low-level ChartQA tasks~\cite{DBLP:conf/infovis/AmarES05}.
In addition, to conduct customized evaluations, we need access to chart metadata (\eg underlying data), not just images.
Therefore, we curate a large-scale dataset, \tdataset, consisting of 22,347 quartets - each with a chart, a query, and its answer.

\textit{We also include more detailed discussion about the related work in Appendix~\ref{sec:related}.}

\section{Conclusion}
In this paper, we curate a large-scale dataset, ChartInsights, specifically designed for low-level ChartQA tasks. To evaluate the capabilities of MLLMs on these tasks, we conduct a series of experiments using 19 widely used MLLMs from multiple perspectives. Specifically, we investigate the impact of chart variants and visual prompts on performance, demonstrating the importance of chart quality and visual attention.
We also propose a new textual prompt strategy, named Chain-of-Charts, to harness the capabilities of MLLMs for low-level ChartQA. By incorporating visual prompts, we achieve the best average accuracy of 84.32\% using GPT-4o.
Future work can explore incorporating data prompts and multi-agent frameworks to further enhance the effectiveness of MLLMs in diverse ChartQA tasks.

%% file: secs/limitations.tex
\section*{Limitations}

\paragraph{Limited Chart Types.}
Our experiments set benchmarks for the performance of GPT-4o across seven widely used chart types, providing valuable insights into the model's capabilities in low-level ChartQA tasks. However, this focus inherently excludes a range of more complex chart types, such as heatmaps, radar charts, and others, which present unique analytical challenges and opportunities for data representation.
Therefore, including a more diverse chart type, especially those with complex structure and interpretation such as heat maps and radar charts, will provide a more comprehensive perspective on ChartQA for MLLM. This extension is critical for assessing the adaptability and effectiveness of MLLM in a wider range of graph interpretation tasks.

\paragraph{Limited Visual Prompts Design Space.} Our exploration of visual prompts in facilitating ChartQA tasks with GPT-4o has shown their potential to enhance model performance. However, our investigation into the design space of visual prompts has been preliminary, lacking a comprehensive and systematic exploration of the full spectrum of possibilities. This limitation narrows the scope of our findings and potentially overlooks more effective strategies that could further improve the accuracy and efficiency of MLLMs in interpreting and analyzing charts.


 \paragraph{Lacking of Considering the Data Prompts.} Our approach primarily relied on chart images, neglecting the underlying data that generated these charts. This omission could hinder the model’s ability to perform more complex analysis and reasoning based on the actual data points. Future work could explore integrating the underlying data as part of the prompt, potentially through multimodal inputs, to provide a richer context for the model’s analyses.

\paragraph{Without Fine-tuning MLLMs.} We only use the ``off-the-shelf'' GPT-4o to conduct evaluation, without considering other MLLMs because GPT-4o is known as one of the best models in the visual question-answering task. In addition, we don't perform task-specific fine-tuning because we want to benchmark GPT-4o in low-level tasks and investigate the impact of textual and visual prompts, which is orthogonal to fine-tuning the MLLMs.
Future work can fine-tune MLLMs using our dataset to investigate their effectiveness.



\section*{Acknowledgements}
This paper is supported by NSF of China (62402409, 62132017), Guangdong Basic and Applied Basic Research Foundation (2023A1515110545),  CCF-Huawei Populus Grove Fund (CCF-HuaweiDB202403),
Shandong Provincial Natural Science Foundation under Grant (ZQ2022JQ32), Fundamental Research Funds for the Central Universities, and the Research Funds of Renmin University of China (24XNKJO3).

%% file: appendix/dataset.tex
\section{\tdataset Construction}
\label{sec:overview}

\begin{figure*}[b!]
	\centering
	\includegraphics[width=\textwidth]{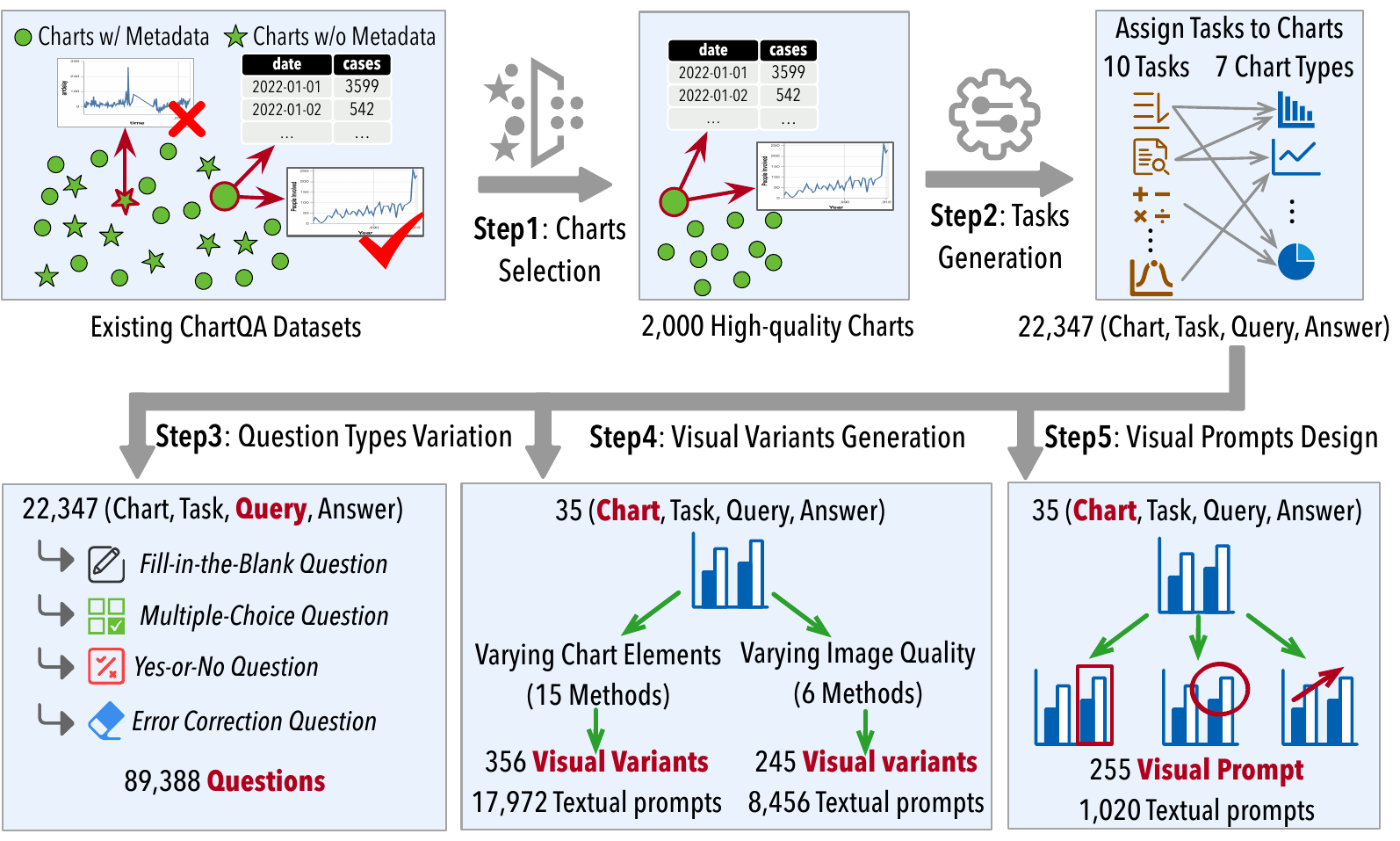}
	\caption{The Pipeline for \tdataset Construction}
	\label{fig:workflow}
\end{figure*}

In this section, we will first discuss the design goals for curating datasets for low-level tasks (Section~\ref{sub:datagoal}).
We will then provide details of constructing \tdataset (Section~\ref{subsec:construct}).

\subsection{Design Goals for \tdataset}
\label{sub:datagoal}

\paragraph{G1: Supporting Low-level Data Analysis Tasks.} 
Our first goal is to facilitate the support of 10 low-level data analysis tasks~\cite{DBLP:conf/infovis/AmarES05, lowleveltasks}.  This focus addresses a critical gap in existing ChartQA datasets, which often overlook the granularity required to fully understand and interact with the data presented in charts.


\paragraph{G2: Evaluating Visual and Textual Variants on Charts.}
We highlight the critical role of visual variants (\textit{e.g.} color, size, shape) in data visualizations, which are key to conveying and interpreting information effectively. Despite their importance, these variants are often neglected in existing ChartQA datasets and evaluations. Our goal is to address this issue by incorporating a diverse array of visual variants, including varying chart elements, image quality, and visual prompts. In addition, we also want to investigate the impact of different textual prompts on the low-level analysis task.

\paragraph{G3: Making Metadata Available.}
The third goal tackles the prevalent issue of inaccessible data and metadata in current ChartQA datasets. By offering comprehensive access to each chart's metadata, such as the source data, chart type, and visual element specifics (like color schemes and labels), our dataset enhances analytical depth into chart design's impact on ChartQA performance.

\subsection{\mbox{Construction Pipeline for \tdataset}}
\label{subsec:construct}

To fulfill our three design goals, our construction process begins with the collection of charts with metadata from existing datasets. {After collecting and reviewing a large number of datasets, we decided to extract charts from nvBench and ChartQA. The reason is that most charts in these two datasets contain numerical information of elements, which can meet the requirements for 10 low-level ChartQA tasks. We extracted approximately 900 charts from ChartQA~\cite{chartqa} and about 1100 charts from nvBench~\cite{nvbench}}. Next, we meticulously assign specific low-level data analysis tasks to appropriate chart types. Lastly, we develop diverse textual prompt strategies, along with visual variants and prompts, tailored to each chart. Note that we save all metadata during the construction process, which can make the users customize their dataset based on \tdataset easily.

As shown in Figure~\ref{fig:workflow}, the construction of our \tdataset consists of five steps: {Candidate Charts Selection}, 
{Low-Level Tasks Generation}, 
{Textual Prompts Design}, 
{Visual Variants Generation}, and 
{Visual Prompts Design}.

\paragraph{Step 1: Candidate Charts Selection.}
In order to more comprehensively evaluate the ability of MLLMs on low-level data analysis tasks, and to conduct more detailed and extended experiments, the datasets (tabular data) and visualization charts we collected need to meet the following three requirements: 
First, these datasets should contain the original metadata of the chart such as the underlying data for rendering, allowing us to create customized reasoning tasks based on the metadata data. Second, the charts in these datasets should contain data labels, because the lack of data labels will greatly limit the types of low-level tasks. {Taking Figure~\ref{fig: Order_example} as an example, the chart shows data for different days of the week, and the data labels are as follows: the data label corresponding to Fri is 92020 and the data label corresponding to Mon is 75806.}
Third, these datasets should contain both simple and complex charts so that the difficulty of the charts is reasonable. 

Then, we get a total of 2K high-quality charts as well as their metadata as our initial dataset. The initial dataset contains a total of 7 types of charts, namely stacked bar charts, grouped bar charts, basic bar charts, line charts, grouped line charts, scatterplots, and pie charts.

\begin{figure*}[t!]
	\centering
\includegraphics[width=\textwidth]{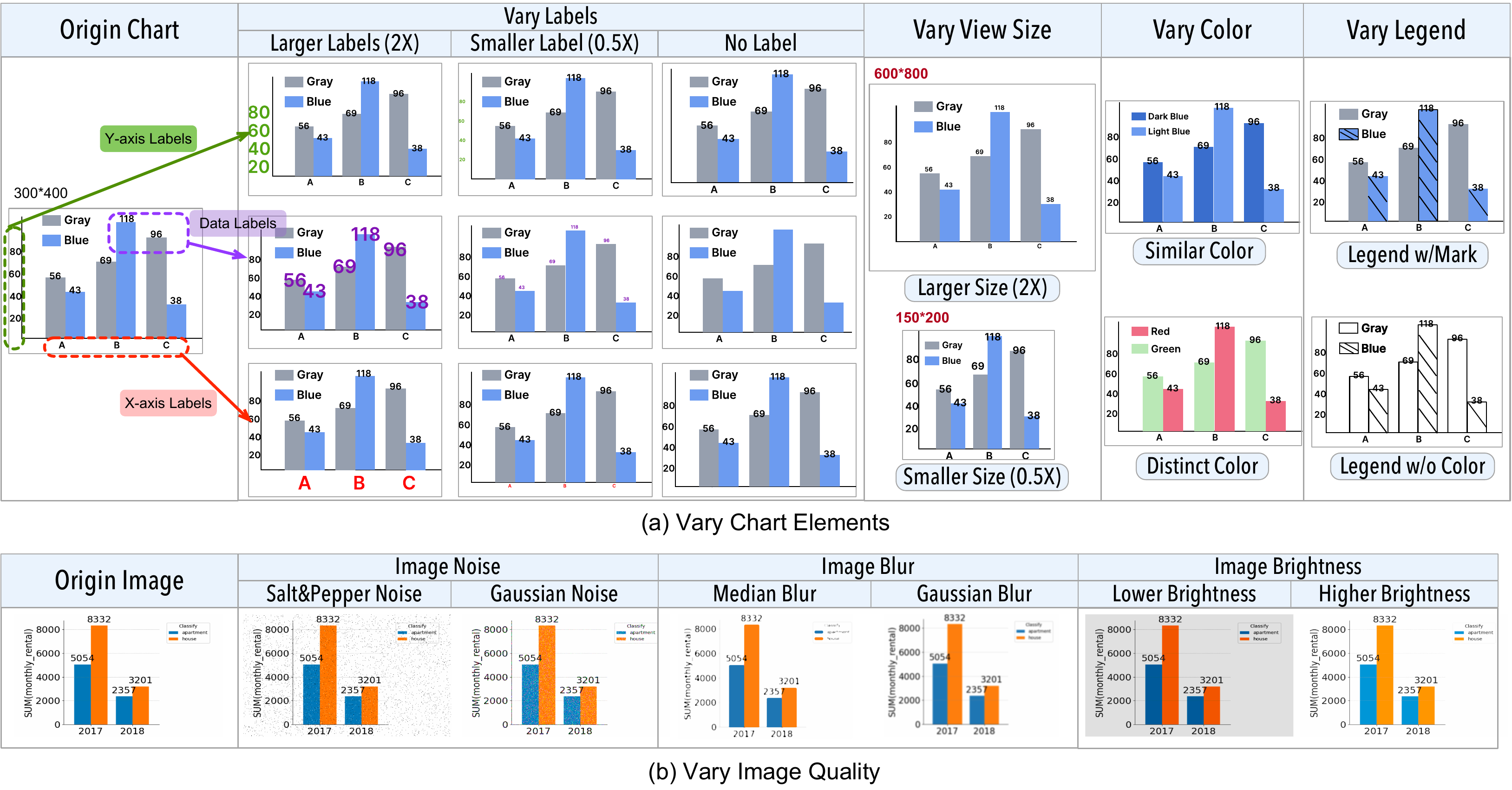}
	\caption{Vary Visual Elements on Charts. (a) We vary chart labels, view size, color, and legend in a total of 15 ways. (b) We alter the image quality by adding noise, applying blur, and adjusting brightness.}
\label{fig:vary_chart_grammar}
\end{figure*}

\begin{figure*}[t!]
	\centering
\includegraphics[width=\textwidth]{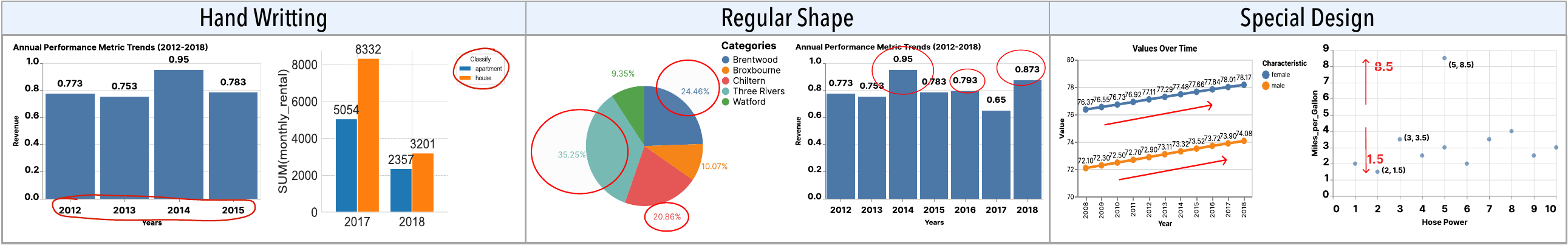}
	\caption{Three Types of Visual Prompts.}
	\label{fig:visualprompts}
\end{figure*}

\paragraph{Step 2: Low-level Tasks Generation.}
Next, we design a set of low-level tasks for the collected charts. We follow the approach of previous works on designing low-level tasks for charts~\cite{visualizationanalysis, DBLP:conf/infovis/AmarES05, lowleveltasks}, resulting in 10 low-level tasks in this paper, as shown in top of Figure~\ref{fig:teaser}.  
We group the 10 low-level tasks into three categories, namely Analysis, Search, and Query, based on their purpose and required reasoning abilities~\cite{visualizationanalysis}.

Next, we should decide which tasks are applicable to which types of charts. We will follow the recommendations on the task-based effectiveness of humans to assign the tasks to each chart type~\cite{lowleveltasks}.
Finally, we have 22,347 (chart, task, question, answer).

\paragraph{Step 3: Question Type Variation.}
In order to better explore the impact of different types of questions influence the interaction with MLLMs. We have designed 4 question types, namely Fill-in-the-Blank, Multiple-choice, Yes-or-No, and Error Correction questions. 
1) For Fill-in-the-Blank prompt, we maintain the asking method of the initial question and set the answer format for Fill-in-the-Blank prompt; 
2) For Multiple-choice prompt, we still maintain the asking method of the initial question, but at this time we will provide a list of choices for \mllms, which usually contains one correct answer and two wrong answers, and tells \mllms to choose the answer from the options; 
3) For Yes-or-No prompt, we first change the initial question to a true or false question and tell \mllms whether it needs to be answered correctly or Wrong; and
4) For Error Correction prompt, we put the wrong answer into the original question with a certain probability and change it into a statement.

We vary the 22,347 quartets (chart, task, query, answer) by the four question types mentioned above, resulting in 89,388 quartets (chart, task, question, answer).

\paragraph{Step 4: Visual Variants Generation.}
Visual variants (\textit{e.g.} color, size, shape) of a chart play a key role in delivering insights, but these variants are often overlooked in existing ChartQA datasets and evaluations, and thus we aim to bridge this gap. To this end, we vary the chart elements and add image noise to vary the chart quality.

\etitle{Step 4.1: Varying the Chart Elements.} As shown in Figure~\ref{fig:vary_chart_grammar}(a), we change the visual elements of these charts from four aspects, namely labels, chart scale, element color, and legend.
To achieve this, we sample 5 charts from each category of charts as seeds, resulting in 35 charts.
For varying labels, we enlarge, reduce, and remove the x-axis, y-axis, and data labels, respectively.
For varying view sizes, we enlarge and reduce the chart, respectively.
For varying element color, we change the elements in the chart to the same color or a higher contrast color;
For varying legend, we first add marks to different types of categories, and then delete the colors.
Finally, we generate 356 visual variants for 35 charts. These 356 visual variants (charts) are associated with 17,972 textual prompts and cover 10 low-level tasks.

\etitle{Step 4.2: Varying the Image Quality.} 
We add image noise, apply image blur, and adjust the brightness to vary the chart image quality, as shown in Figure~\ref{fig:vary_chart_grammar}(b).
To achieve this, we sample 5 charts from each category of charts as seeds, resulting in 35 charts.
For adding image noise, we choose Gaussian noise and salt and pepper noise;
For applying image blue, we use median blur and Gaussian blur;
For adjusting image brightness, we choose to make the brightness of the chart higher and lower.
Finally, we generate 245 visual variants for 35 charts. These 245 visual variants (charts) are associated with 8,456 textual prompts and cover 10 low-level tasks.

\paragraph{Step 5: Visual Prompts Design.}
Kong et al.~\cite{overlays} presented five types of graphical overlays to enhance users' capabilities in performing data analysis tasks such as extraction and comparison of numerical values. Intuitively, we want to verify whether overlays would have a positive impact on \mllms's performance.
Therefore, we design three types of visual prompts (\textit{i.e.} graphical overlays) for the charts.

We consider three types of visual prompts, as shown in Figure~\ref{fig:visualprompts}.
The first is to directly circle the content in the chart that is highly relevant to the question in handwriting, such as circling the values of the two elements mentioned in the reasoning question.
The second method is regular shapes, which uses regular shapes (such as circles or rectangles) to label elements in the diagram. This makes it easier to use the size of a shape to imply the sequential relationship of elements. For example, use three circles of different sizes to correspond to the three values in the ordering task.
The third way is special design. We design effective visual prompts tailored for different low-level tasks. For example, we use arrows to represent the monotonicity of the trend, for the correlation task. 
To generate the visual prompts, we first sample 35 charts from seven chart types, then apply various visual prompt strategies to them, resulting in 255 charts with different visual prompts. These 255 charts are associated with 1020 questions for 10 low-level tasks.

\section{Ten Low-Level ChartQA Tasks}
\label{app:dataset}

\begin{figure}[t!]
	\centering
\includegraphics[width=\columnwidth]{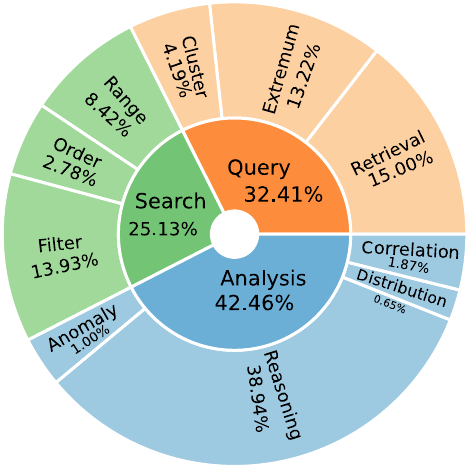}
	\caption{Low-level Tasks Distribution}
	\label{fig:dataset_stat1}
\end{figure}

Our \tdataset include 10 low-level data analysis tasks on charts, as shown in Figure~\ref{fig:dataset_stat1}.
These tasks are well-defined by the visualization and visual analysis community~\cite{DBLP:conf/infovis/AmarES05, lowleveltasks}.

\begin{figure}[htbp]

	\includegraphics[width=\columnwidth]{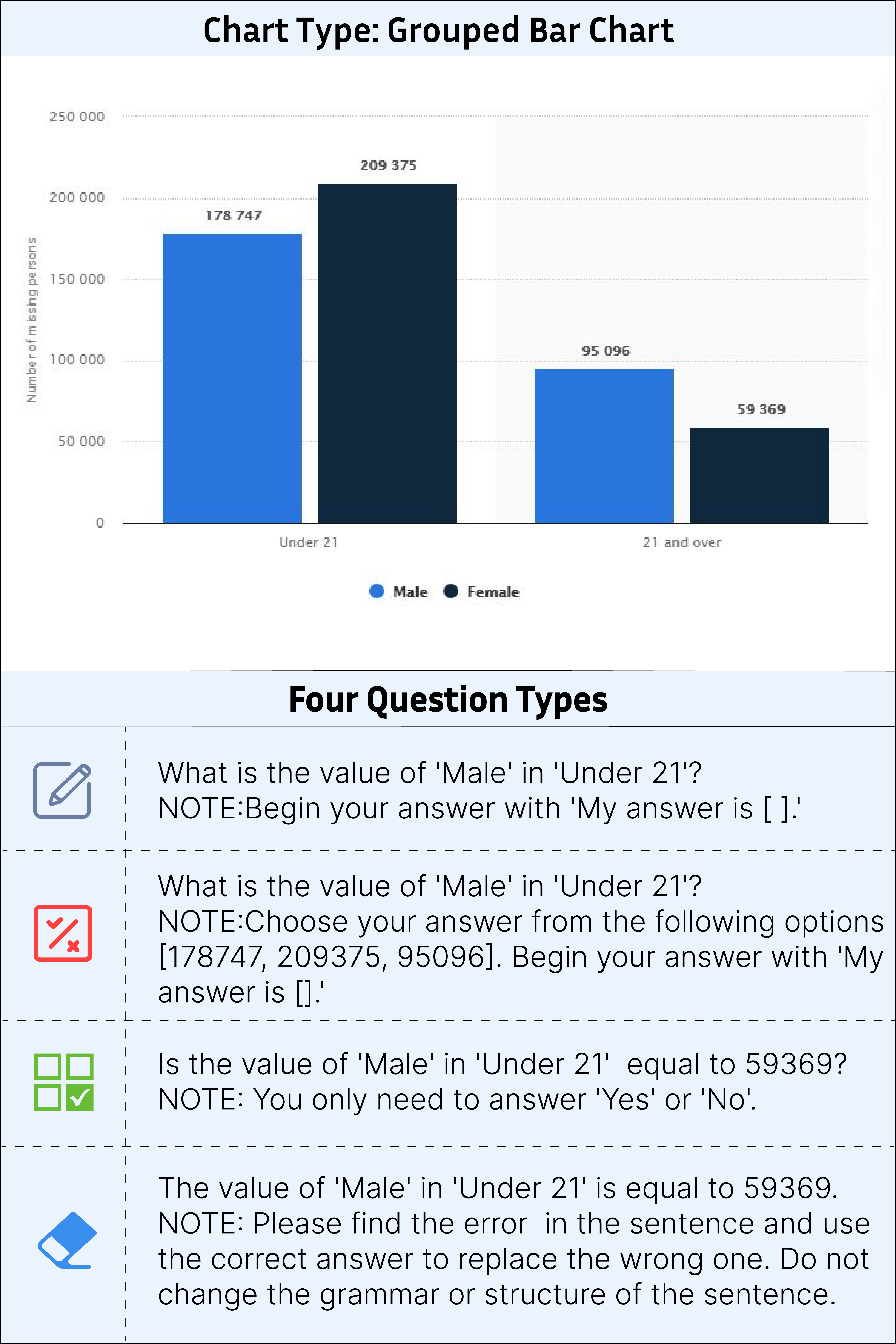}
	\vspace{-2em}
	\caption{An Example of Data Retrieval Tasks}
	\label{fig:data_retrieval_example}
\end{figure}

\paragraph{T1: Data Retrieval.} Users will be asked to locate the value of an element based on some information, or they may be asked to answer structural questions such as how many elements there are in the chart. Figure~\ref{fig:data_retrieval_example} shows an example. 

\begin{figure}[htbp!]
	
	\includegraphics[width=\columnwidth]{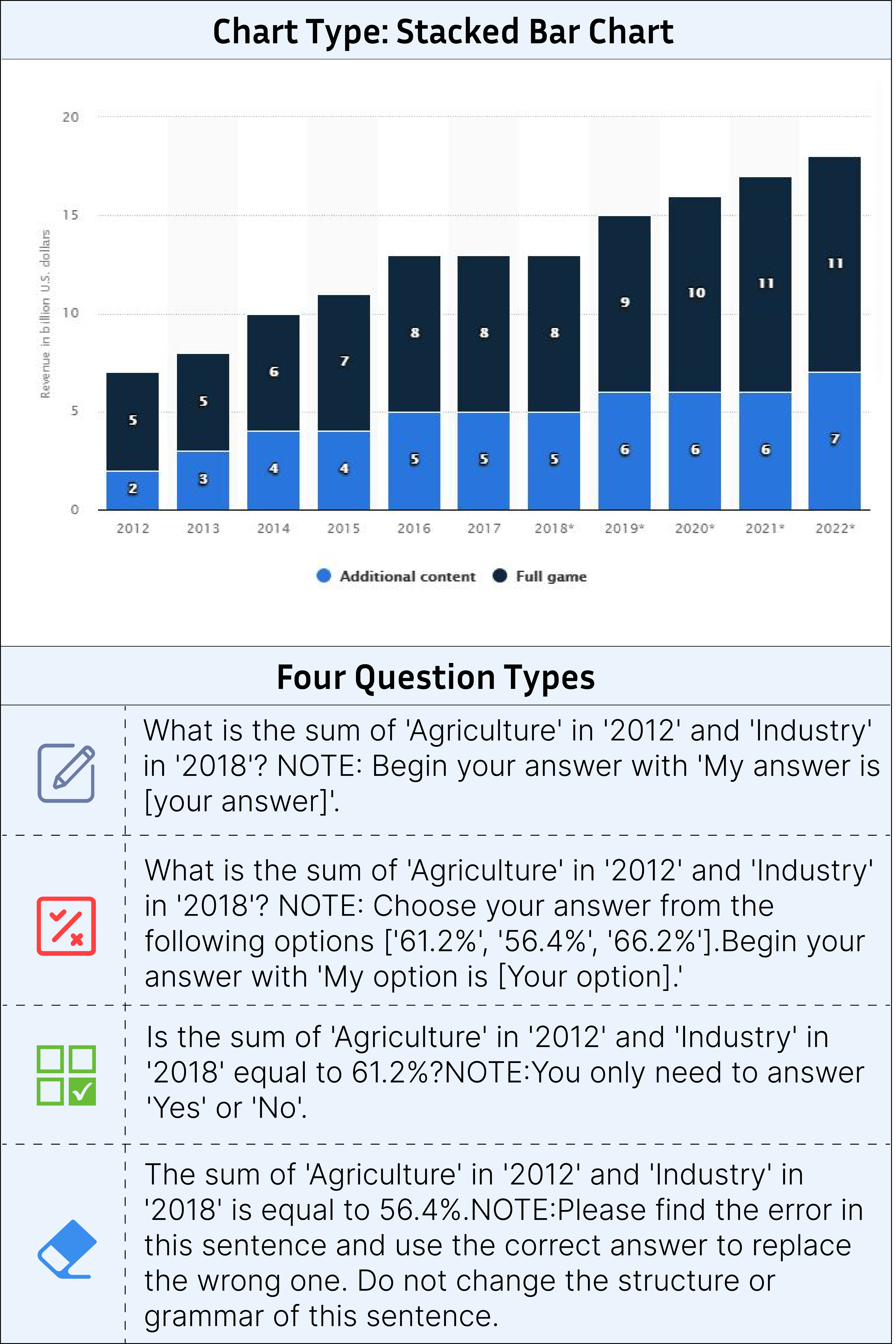}
	\vspace{-2em}
	\caption{An Example of Reasoning Tasks}
	\label{fig:Reasoning_example}
\end{figure}

\paragraph{T2: Reasoning.} Users need to calculate the aggregate value of multiple elements based on the data in the chart and the requirements of the question.  Figure~\ref{fig:Reasoning_example} shows an example.

\begin{figure}[htbp!]

	\includegraphics[width=\columnwidth]{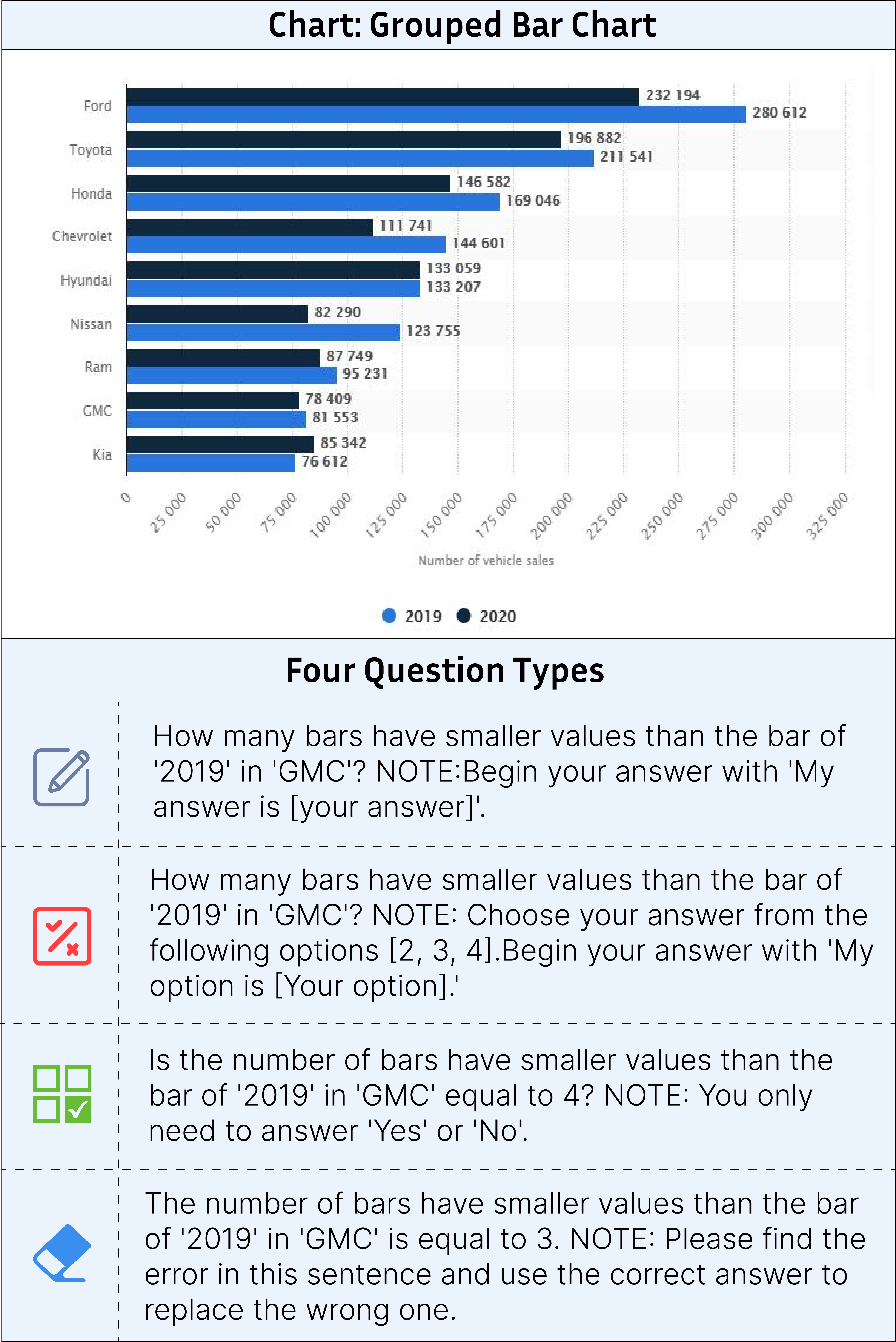}
	\vspace{-2em}
	\caption{An Example of Filter Tasks}
	\label{fig:Filter_example}
\end{figure}

\paragraph{T3: Filter.} This task will randomly select a value of an element in the current chart as a benchmark, and users need to filter the remaining elements according to this benchmark and the requirements of the question. Figure~\ref{fig:Filter_example} shows an example.

\begin{figure}[htbp]

	\includegraphics[width=\columnwidth]{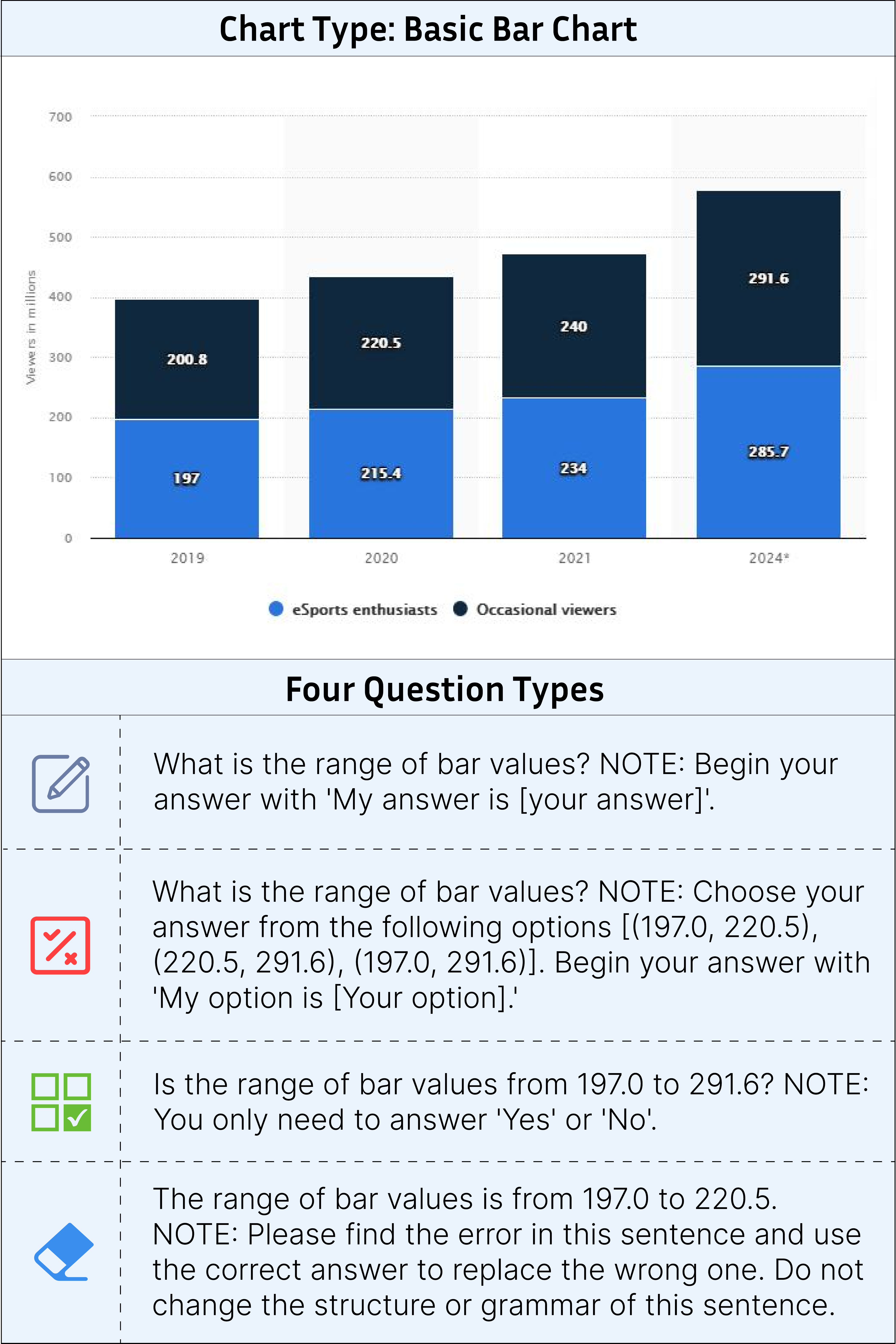}
	\vspace{-2em}
	\caption{An Example of Determine Range Tasks}
	\label{fig:Determine_Range_example}
\end{figure}

\paragraph{T4: Determine Range.} This task requires users to determine the value range of the chart based on the values of the elements in the chart. Figure~\ref{fig:Determine_Range_example} shows an example.

\begin{figure}[htbp]

	\includegraphics[width=\columnwidth]{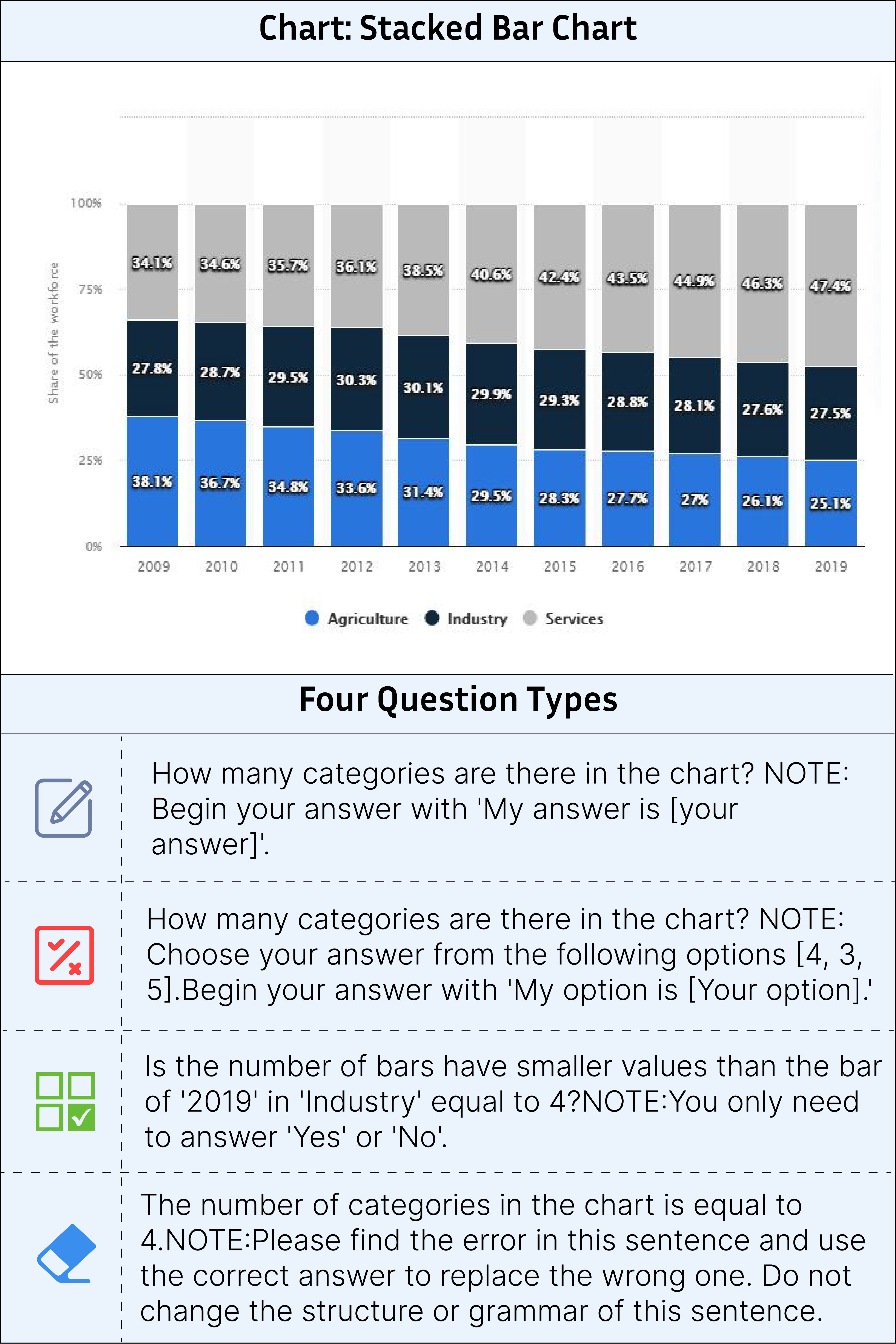}
	\vspace{-2em}
	\caption{An Example of Cluster Tasks}
	\label{fig:Cluster_example}
\end{figure}

\paragraph{T5: Cluster.} As shown in Figure~\ref{fig:Cluster_example}, this task requires users to return the number of categories of elements in the chart. 

\begin{figure}[htbp]
	\includegraphics[width=\columnwidth]{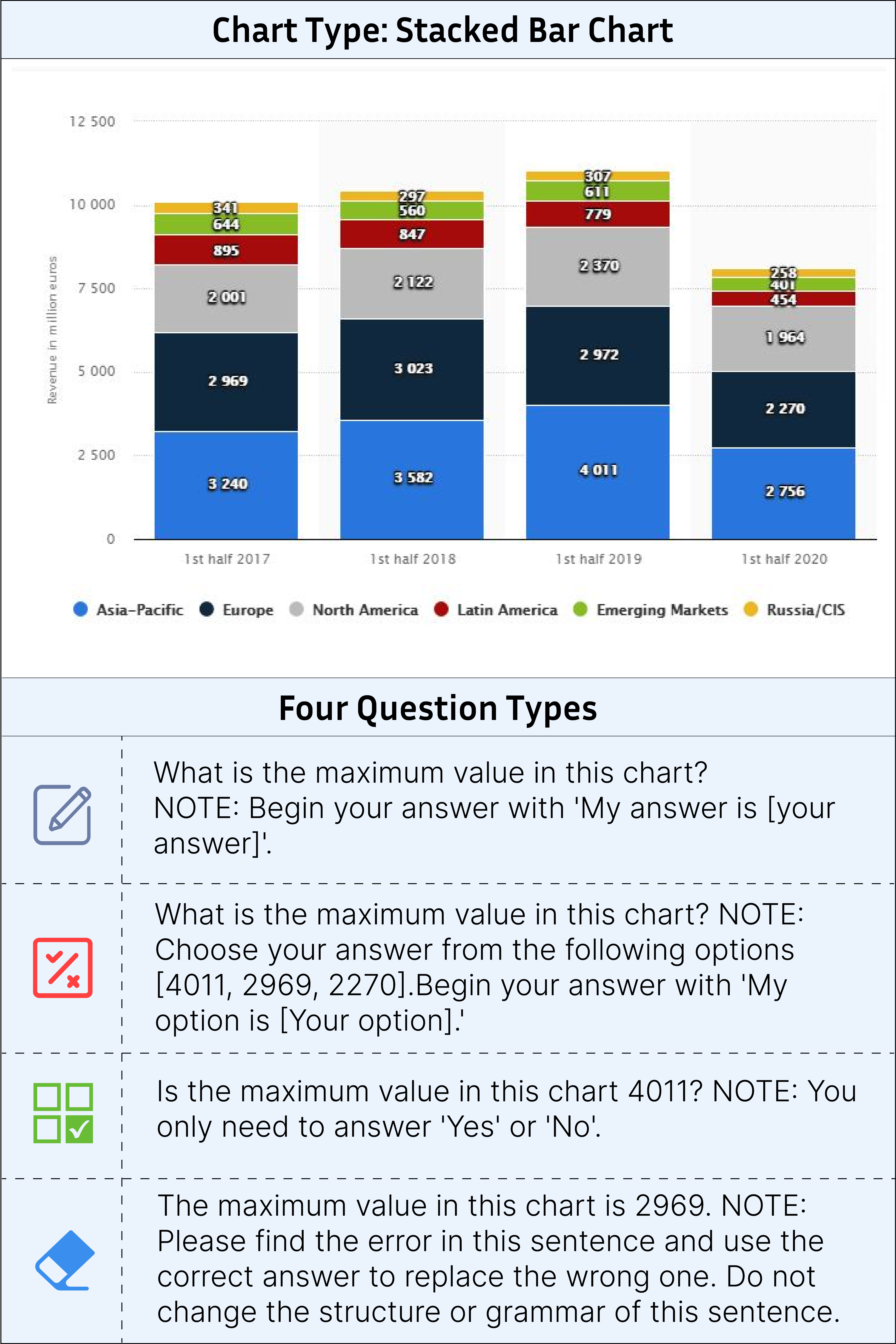}
	\vspace{-2em}
	\caption{An Example of Find Extreme Tasks}
	\label{fig:Find_extreme_example}
\end{figure}

\paragraph{T6: Find Extreme.} This question requires users to find the maximum and minimum values in the chart and return them, as depicted in Figure~\ref{fig:Find_extreme_example}.

\begin{figure}[htbp]
	\centering
	\includegraphics[width=\columnwidth]{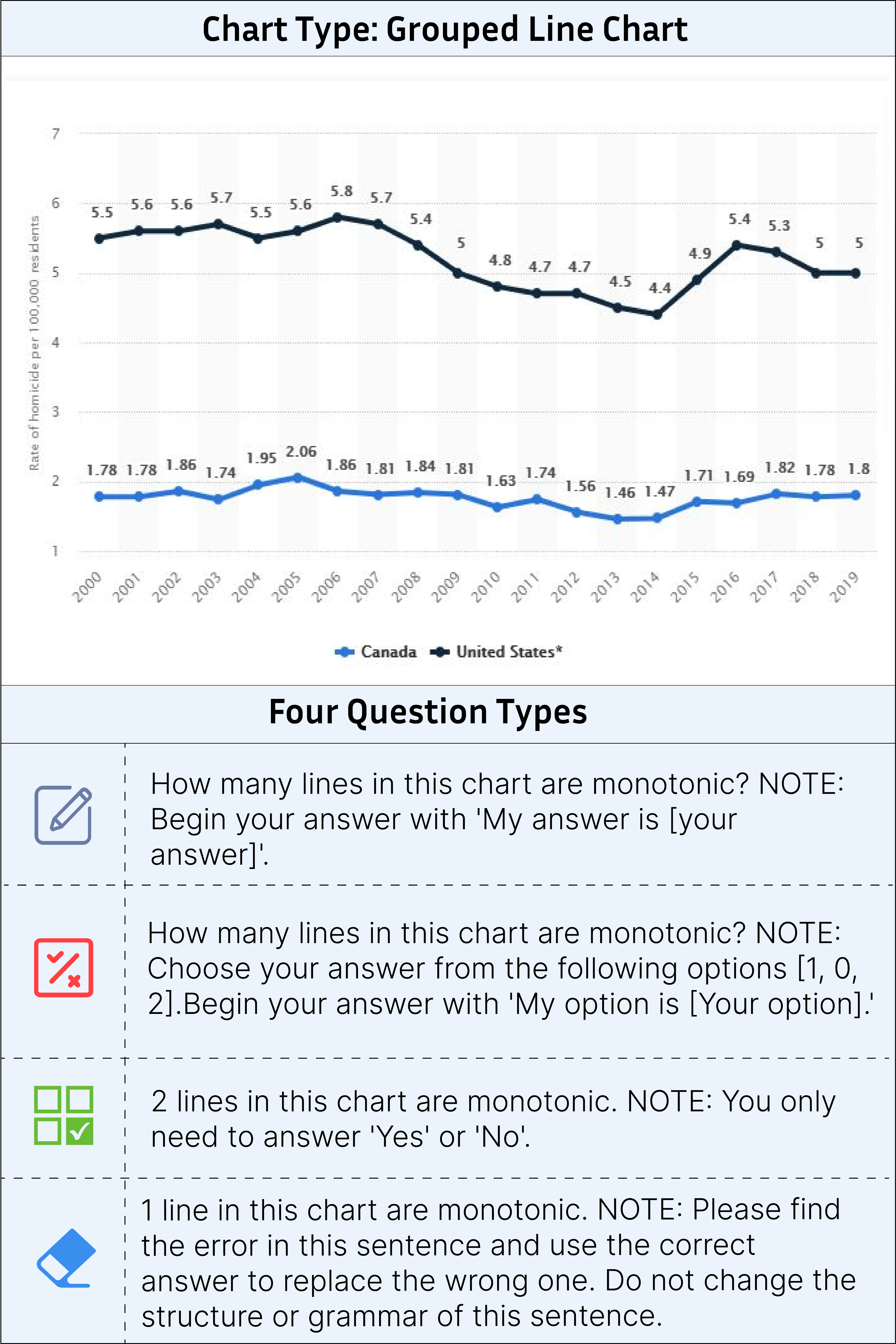}
	\vspace{-2em}
	\caption{An Example of Correlation Tasks}
	\label{fig:Correlation_example}
\end{figure}

\paragraph{T7: Correlation.} Users need to determine the relationship between the changes in the elements in the chart and the axes. Some questions can be answered visually, but many require a reasoning process. Figure~\ref{fig:Correlation_example} shows an example.

\begin{figure}[htbp]
	
	\includegraphics[width=\columnwidth]{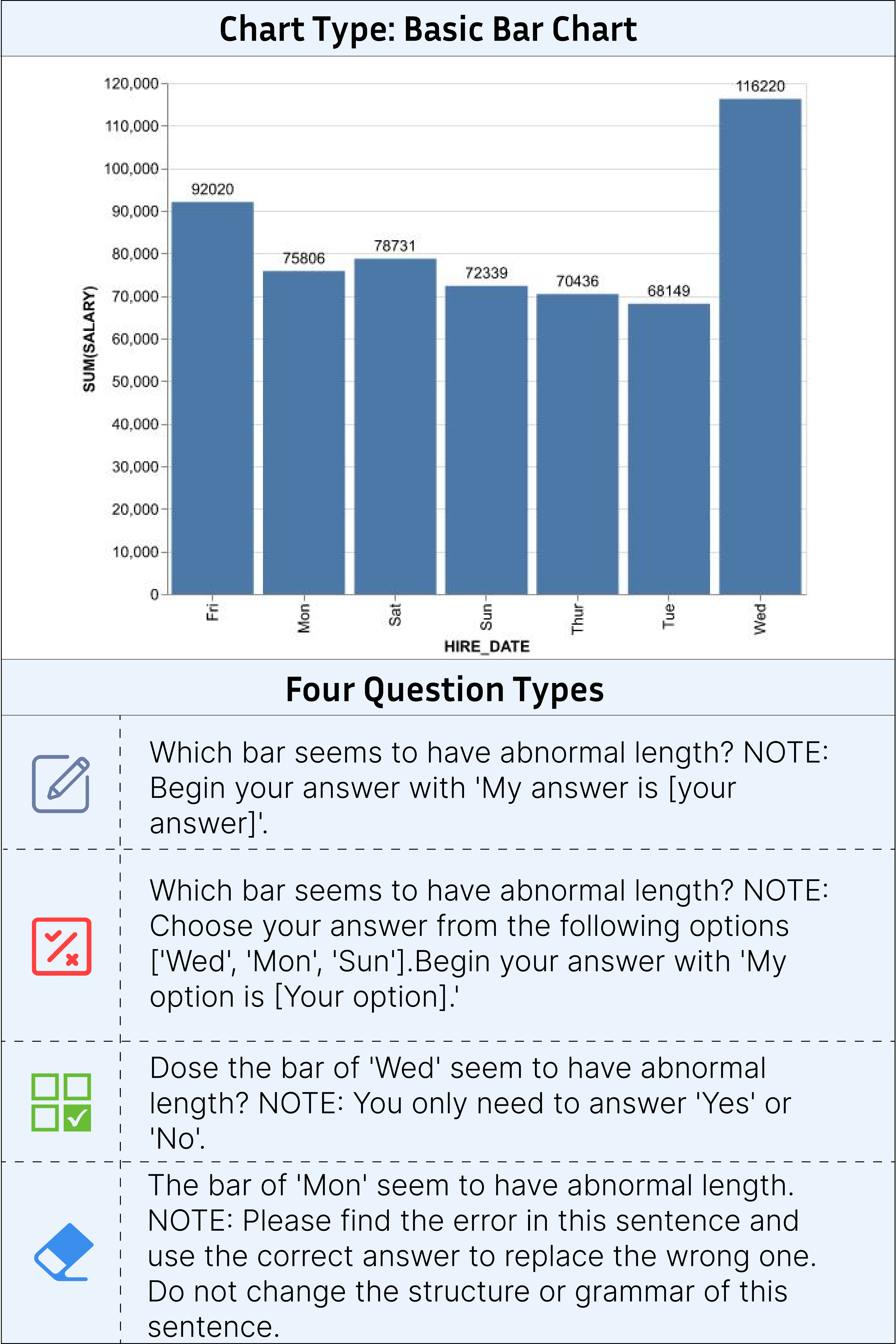}
	\vspace{-2em}
	\caption{An Example of Find Anomaly Tasks}
	\label{fig: Find_Anomaly_example}
\end{figure}

\paragraph{T8: Find Anomaly.} This task requires users to identify values that appear to be anomalies based on their different values, either visually or through calculation. Figure~\ref{fig: Find_Anomaly_example} shows an example.

\begin{figure}[htbp]
	\includegraphics[width=\columnwidth]{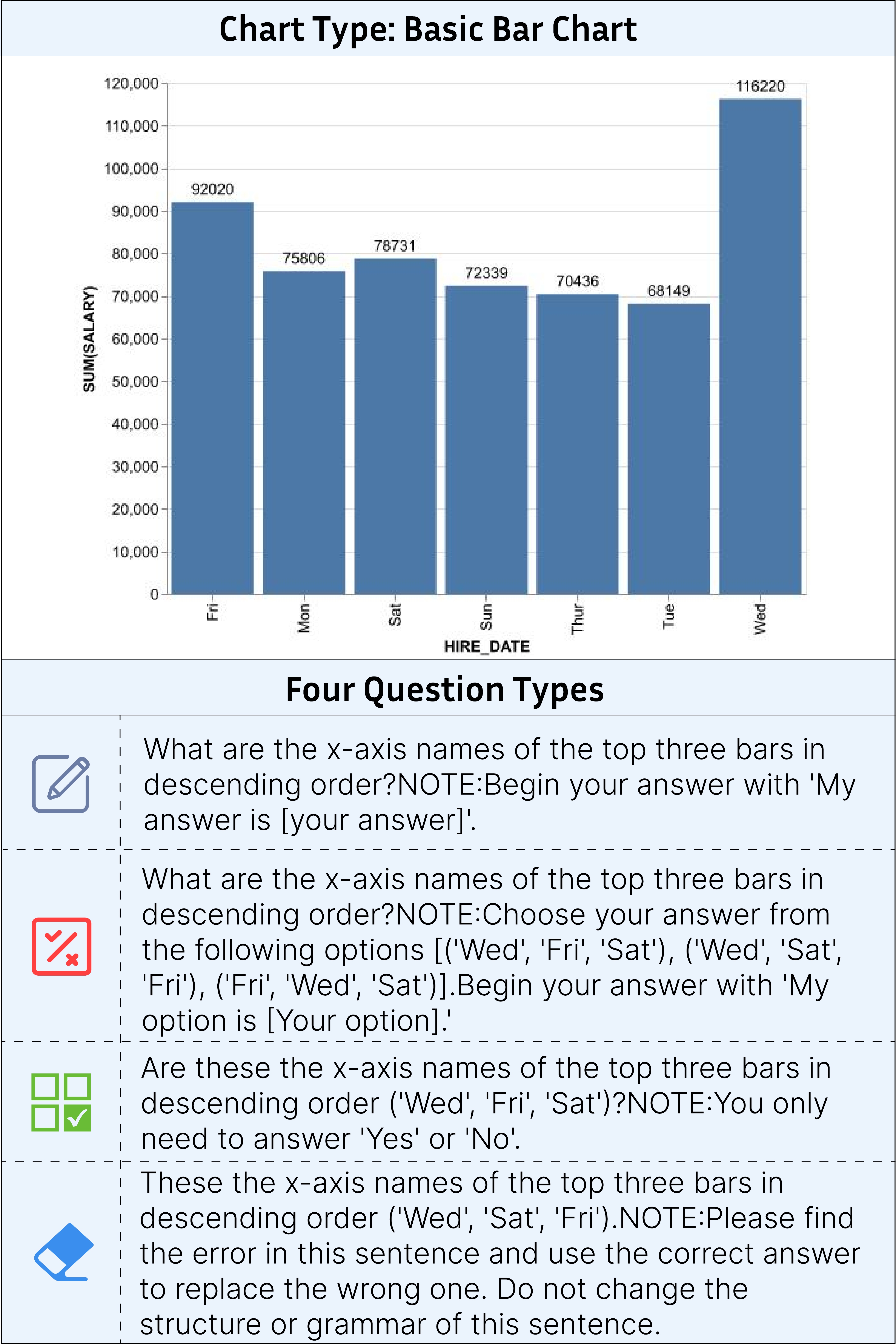}
	\vspace{-2em}
	\caption{An Example of Order Tasks}
	\label{fig: Order_example}
\end{figure}

\paragraph{T9: Order.} This task involves sorting the elements in the chart. Users will be asked to sort the elements in ascending or descending order of value and output the names of the top three elements for each type of sorting. Figure~\ref{fig: Order_example} shows an example.
\begin{figure}[htbp]
	\includegraphics[width=\columnwidth]{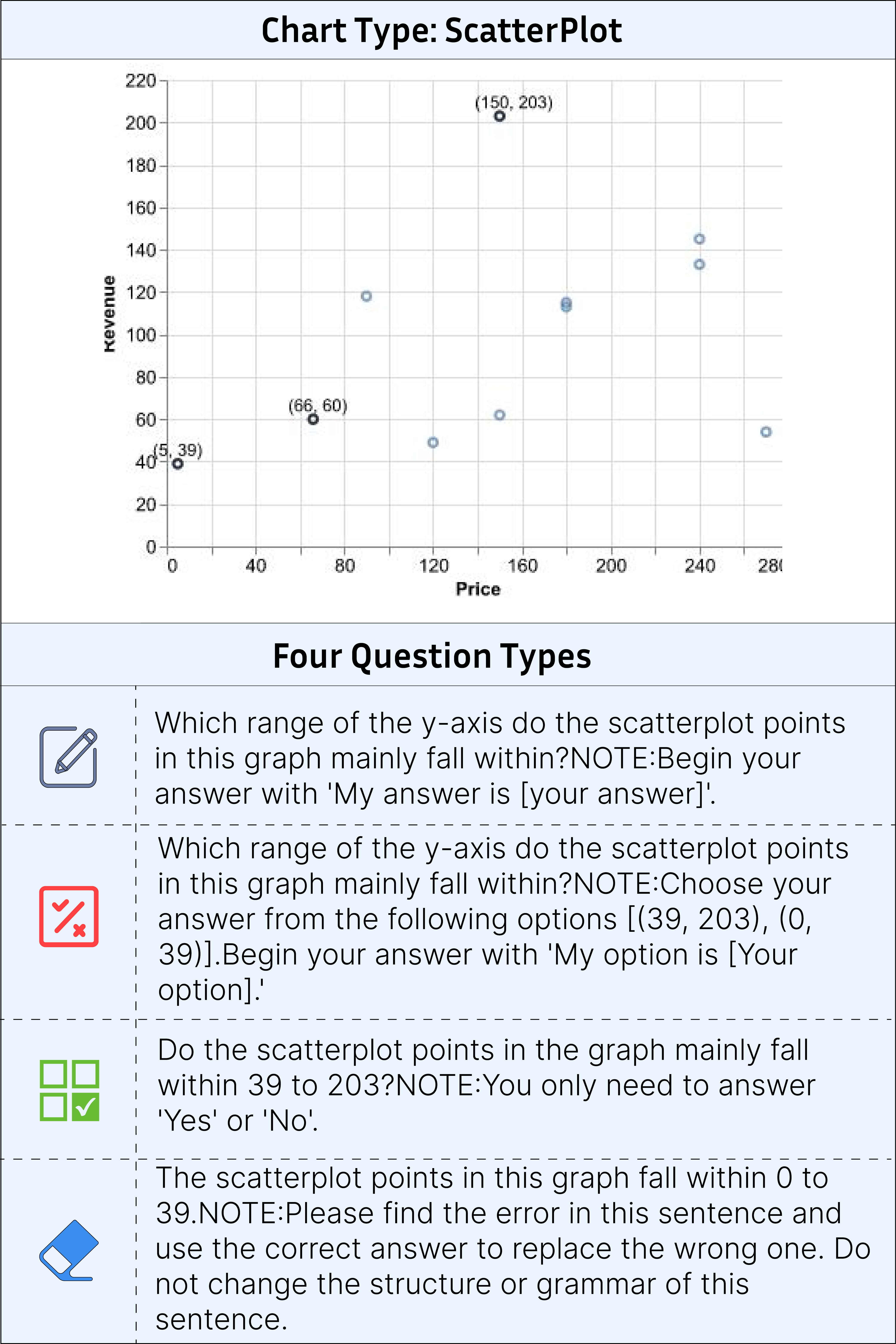}
	\vspace{-2em}
	\caption{An Example of Distribution Tasks}
	\label{fig: Distribution_example}
\end{figure}

\paragraph{T10: Distribution.} This task is mainly for scatter plots, where users need to determine the distribution range of the dots. Figure~\ref{fig: Distribution_example} shows an example.


%% file: appendix/experiment.tex
\section{More Experimental Details on GPT-4o}
\label{app:experiment}

The results include more details about evaluation on GPT-4o.

\paragraph{More Discussions on Exp-1.} Table~\ref{tab:overall_result_seven_charts_gpt-4o} shows the overall accuracy of GPT-4o for different chart types in 10 low-level tasks. Overall, GPT-4o has the highest performance on basic bar charts, reaching an average accuracy of 84.05\%. The main reason is that the chart structure of the basic bar chart is relatively simple. Similarly, GPT-4o achieves better results on charts with simple structures such as scatter plots and pie charts. For charts with complex structures, such as stacked bar charts, grouped bar charts, and grouped line charts, the average accuracy of GPT-4o is close to 50\%.

\begin{table*}[htbp]
    \centering
    \caption{The Average Accuracy of Different Chart Types  vs.Ten Low-level Tasks (``--'' means ``N/A'') on GPT-4o}
    \label{tab:overall_result_seven_charts_gpt-4o}
    \fontsize{8pt}{10pt}\selectfont
    \resizebox{\linewidth}{!}{
        \begin{tabular}{l|cccc|ccc|ccc|c}
            \toprule
            \multirow{2}{*}{Chart Types} & \multicolumn{4}{c|}{Analysis}& \multicolumn{3}{c|}{Search} & \multicolumn{3}{c|}{Query}  & \multirow{2}{*}{Overall (\%)} \\
            
            \cmidrule(lr){2-5} \cmidrule(lr){6-8} \cmidrule(lr){9-11}
            & \begin{tabular}[c]{@{}c@{}}Reasoning\end{tabular} & \begin{tabular}[c]{@{}c@{}}Anomaly\end{tabular} & \begin{tabular}[c]{@{}c@{}}Distribution\end{tabular} & \begin{tabular}[c]{@{}c@{}}Correlation\end{tabular} & \begin{tabular}[c]{@{}c@{}}Range\end{tabular} & \begin{tabular}[c]{@{}c@{}}Order\end{tabular} & \begin{tabular}[c]{@{}c@{}}Filter\end{tabular} & \begin{tabular}[c]{@{}c@{}}Retrieval\end{tabular} & \begin{tabular}[c]{@{}c@{}}Extremum\end{tabular} & \begin{tabular}[c]{@{}c@{}}Cluster\end{tabular} & \\
            \midrule
            Grouped Bar & 43.75 & -- & -- & -- & 83.68 & -- & 37.5 & 80.79 & 96.97 & 63.42 & 62.72 \\
            Stacked Bar & 43.75 & -- & -- & -- & 58.46 & -- & 38.97 & 79.41 & 81.07 & 85.66 & 59.83 \\
            Grouped Line & 31.55 & -- & -- & 71.43 & 80.95 & -- & -- & -- & -- & -- & 53.87 \\
            Basic Bar & \textbf{82.98} & \textbf{96.15} & -- & -- & \textbf{95.17} & \textbf{72.34} & \textbf{65.77} & \textbf{87.38} & \textbf{99.43} & -- & \textbf{84.05} \\
            Basic Line & 60.94 & -- & -- & 36.72 & 89.06 & -- & -- & -- & -- & -- & 56.87 \\
            Scatter Plot & 72.79 & 18.14 & \textbf{70.10} & \textbf{87.25} & 78.43 & -- & -- & -- & 96.08 & \textbf{94.23} & 74.58 \\
            Pie & 71.59 & -- & -- & -- & 78.41 & 59.72 & 61.65 & 80.11 & -- & -- & 70.86 \\
            \bottomrule
        \end{tabular}
    }
\end{table*}

\begin{table*}[htbp]
    \centering
    \caption{The Effectiveness of Question Types  vs. Ten Low-level Tasks on GPT-4o}
    \label{tab:overall_result_four_questions_gpt-4o}
    \resizebox{\textwidth}{!}{%
        \begin{tabular}{l|cccc|ccc|ccc|c}
            \toprule
            \multirow{2}{*}{Question Types} & \multicolumn{4}{c|}{Analysis}& \multicolumn{3}{c|}{Search} & \multicolumn{3}{c|}{Query}  & \multirow{2}{*}{Overall (\%)} \\
            
            \cmidrule(lr){2-5} \cmidrule(lr){6-8} \cmidrule(lr){9-11}
            & \begin{tabular}[c]{@{}c@{}}Reasoning\end{tabular} & \begin{tabular}[c]{@{}c@{}}Anomaly\end{tabular} & \begin{tabular}[c]{@{}c@{}}Distribution\end{tabular} & \begin{tabular}[c]{@{}c@{}}Correlation\end{tabular} & \begin{tabular}[c]{@{}c@{}}Range\end{tabular} & \begin{tabular}[c]{@{}c@{}}Order\end{tabular} & \begin{tabular}[c]{@{}c@{}}Filter\end{tabular} & \begin{tabular}[c]{@{}c@{}}Retrieval\end{tabular} & \begin{tabular}[c]{@{}c@{}}Extremum\end{tabular} & \begin{tabular}[c]{@{}c@{}}Cluster\end{tabular} & \\
            \midrule
            Fill-in-the-Blank & 42.45 & 34.38 & 43.14 & \textbf{75.00} & 66.84 & 66.15 & 38.81 & 82.05 & 92.38 & 62.50 & 61.03 \\
            Multiple-Choice & \textbf{68.56} & \textbf{43.75} & \textbf{80.39} & 73.08 & \textbf{99.22} & \textbf{77.69} & 48.81 & \textbf{93.75} & \textbf{99.17} & 71.59 & \textbf{77.78} \\
            Yes-or-No & 68.12 & 21.88 & 68.63 & 69.23 & 86.68 & 58.46 & \textbf{67.12} & 75.64 & 92.38 & \textbf{90.34} & 74.79 \\
            Error Correction & 44.49 & 35.94 & 88.24 & 56.73 & 69.71 & 73.08 & 44.75 & 79.01 & 91.89 & 72.73 & 63.08 \\
            \midrule
            Overall Accuracy (\%) & 55.91 & 33.99 & 70.10 & 68.51 & 80.61 & 68.84 & 49.87 & 82.61 & \textbf{93.96} & 74.29 & 69.17 \\
            \bottomrule
        \end{tabular}
    }
\end{table*}

\paragraph{More Discussions on Exp-2.} Table~\ref{tab:overall_result_four_questions_gpt-4o} presents the overall performance of GPT-4o across 10 low-level tasks with four Question Types. Specifically, GPT-4o exhibits the highest overall accuracy with the Multiple-Choice prompt, achieving 74.79\%. In addition, it also performs well with the Yes-or-No prompt, with 77.78\% accuracy.

\paragraph{More Discussions on Exp-5.}
Heatmaps (Figure~\ref{fig:vary_element}) demonstrate the percentage change in performance across different chart types and task types when chart elements are varied. Bar charts summarize the overall effect, indicating that data labels significantly influence performance.
It consists of two main parts:

Chart Types (a): This heatmap shows the change in performance (in percentages) for various chart types when different elements (e.g., data labels, X/Y labels, size, color) are modified. Performance varies widely, with removing data labels and X labels often result in a significant drop in performance, while larger data labels and different colors often improve performance.

Task Types (b): This heatmap shows how the performance of different task types changes when similar chart elements change. Tasks such as reasoning and finding extremum are particularly sensitive to change, and performance can be significantly degraded when data labels are removed or reduced.

\begin{figure*}[hbpt!]
	\centering	\includegraphics[width=\textwidth]{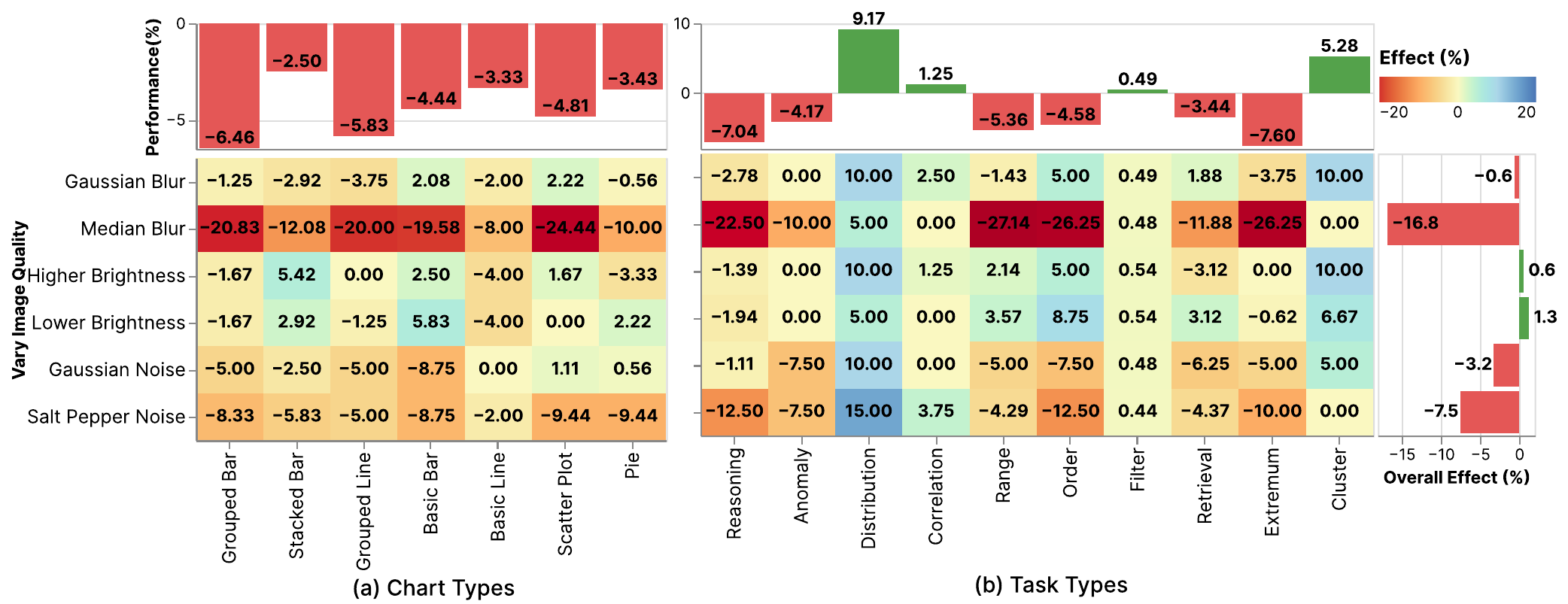}
	\caption{
 The Impact of Image Quality on GPT-4o's Performance.
 }
	\label{fig:vary_quality}
\end{figure*}

\paragraph{More Discussions on Exp-6.}
Heatmaps (Figure~\ref{fig:vary_quality}) show how different types of image quality changes affect GPT-4o's performance on various chart types and tasks. Median blur has the most negative impact, making it hard for GPT-4o to read numbers and lowering its performance significantly. On the other hand, changing brightness (either higher or lower) slightly improves performance on most tasks. Overall, poor image quality generally hurts GPT-4o's recognition capability.

\begin{figure*}[htbp!]
	\centering
\includegraphics[width=\textwidth]{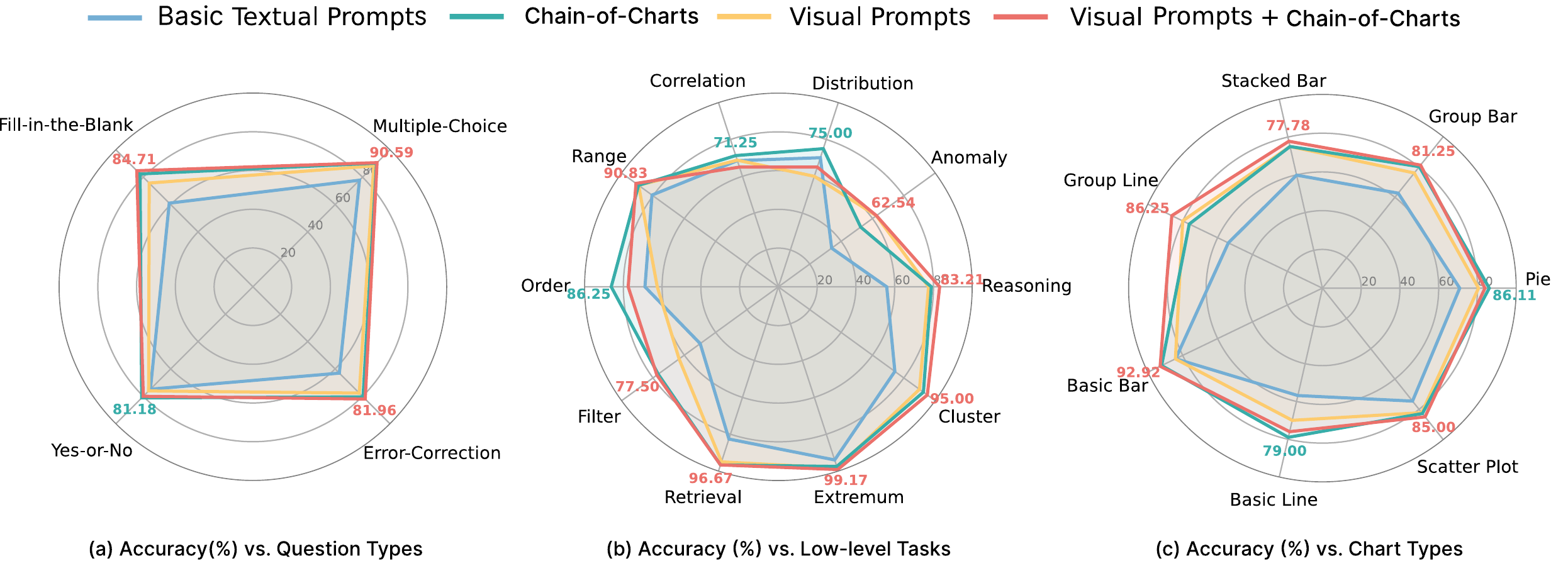}
	\caption{Comparing Different Prompting Methods.}
	\label{fig:gpt_4oradar}
\end{figure*}

\paragraph{Optimizations on Textual and Visual Prompts.}
The radar charts (Figure~\ref{fig:gpt_4oradar}) clearly demonstrate that the effectiveness of different prompt types varies depending on the question, task and chart types. The accuracy of GPT-4o in Exp-1, Exp-3, Exp-4, and Exp-7 is concatenated in series. The combination of visual prompts and chain-of-chart prompts notably enhances performance in various aspects, showcasing their effectiveness in improving GPT-4o's analytical capabilities.


%% file: secs/methods.tex

\section{Lessons Learned}
\label{app:lesson learned}

\paragraph{Effectiveness of Textual Prompts, Visual Prompts and Their Combination.} 
Incorporating various prompt strategies, including textual and visual prompts, significantly impacts \mllms (\eg GPT-4o) accuracy. Textual prompts with structured candidate answers enhance reasoning capabilities, while visual prompts enhance the chart understanding through visual attention, particularly in anomaly detection and filtering tasks.

\paragraph{Importance of Chart Elements and Image Quality.} 
Alterations in chart elements and the quality of chart images influence \mllms's performance. Specifically, certain modifications like larger labels or the absence of data labels can improve the model's efficiency in specific tasks by focusing its attention on visual comparisons. However, image quality degradation, especially median blurring, negatively affects the model's ability to process numerical values accurately.

\paragraph{Strengths and Weaknesses of \mllms in Low-level ChartQA Tasks.} 
\mllms perform well in tasks requiring direct data retrieval and basic comparisons, showing high accuracy in Query and Search task categories. However, \mllms face challenges in more complex reasoning, anomaly detection, and correlation tasks, indicating a need for further optimization of prompting strategies and model training to overcome these limitations.

\paragraph{Potential for Future Application and Development.}
The experiments demonstrate a promising direction for enhancing MLLMs' performance in visual data analysis through the development of specialized prompting strategies and the careful manipulation of visual elements. 

%% file: secs/related.tex
\section{Additional Related Work}
\label{sec:related}

\subsection{Low-Level Analysis Tasks on Charts}
Visualization charts offer numerous insights that aid users in performing data analysis tasks~\cite{DBLP:journals/tvcg/ShenSLYHZTW23,DBLP:journals/tkde/LuoQCTLL22,DBLP:journals/tvcg/LuoTLTCQ22,DBLP:conf/sigmod/TangLOLC22,DBLP:conf/sigmod/Luo00CLQ21,DBLP:journals/debu/Luo00LZY20, DBLP:journals/pvldb/LuoLZYZ0020, DBLP:journals/pvldb/LuoCQ0020, DBLP:conf/icde/LuoQ0018, DBLP:conf/sigmod/LuoQ00W18, DBLP:conf/icde/LuoCQ0020,DBLP:journals/vldb/QinLTL20,vistalk,galvis}. Low-level data analysis tasks typically involve activities requiring direct interpretation and processing of specific visual elements within a chart, such as data retrieval, outlier identification, and correlation determination~\cite{visualizationanalysis, DBLP:conf/infovis/AmarES05, chartqaforblind}. Amar et al.~\cite{DBLP:conf/infovis/AmarES05} identified ten low-level tasks, highlighting the real-world activities users undertake with visualization tools to understand their data. Subsequently, Saket et al.~\cite{lowleveltasks} evaluated the effectiveness of five basic charts across ten low-level analysis tasks using two datasets through a crowdsourced experiment.
In this paper, we aim to evaluate how effectively GPT-4o can interpret charts by using these ten low-level data analysis tasks as a framework.

\subsection{Multimodal Large Language Models}
The field of Multimodal Large Language Models (MLLMs) is experiencing rapid advancements, with efforts concentrated on developing artificial intelligence systems capable of processing and producing multi-modal content, including text, images, videos, and more~\cite{DBLP:journals/corr/abs-2406-07815}. Early research such as CLIP~\cite{clip} demonstrated the effective combination of visual and linguistic information through contrastive learning, while subsequent work like DALL-E~\cite{dalle2} further showcased the potential of Transformer~\cite{transformer} architecture in generating images that match text descriptions.
Building on these foundational successes, the research community has ventured into refining these models for diverse multi-modal applications, employing strategies like fine-tuning and prompt-based learning. For example, VisualGPT~\cite{visualgpt} and BLIP~\cite{blip} have been adapted for Visual Question Answering (VQA) tasks, significantly enhancing their multi-modal task performance. Concurrently, the development of various benchmarks~\cite{aesbench, seedbench, factsbench, mvediobench, videobench2}, including MME~\cite{mmeBenchmark}, has been crucial. These benchmarks provide a wide array of tasks and datasets, facilitating a comprehensive evaluation of MLLMs' abilities across different contexts.
In this paper, we try to harness the  off-the-shelf \mllms for low-level data analysis tasks on charts.

\begin{figure*}[t!]
	\centering
	\begin{tcolorbox}[colback=black!5, colframe=black!75, title=An Example of Basic Textual Prompt]
		\textbf{Initial Question}: What is the minimum value in this chart?
		
		\textbf{Fill-in-the-Blank:} What is the minimum value in this chart? Begin your answer with `My answer is [].'
		
		\textbf{Multiple-Choice:} What is the minimum value in this chart? NOTE: Choose your answer from the following options [A, B ,C]. Begin your answer with `My option is [Your option].'
		
		\textbf{Yes-or-No:} Is the minimum value in this chart equal to ......? NOTE: You only need to answer `Yes' or `No'.
		
		\textbf{Error-Correction:} The minimum value in this chart is equal to ...... NOTE: Please find the error in this sentence and use the correct answer to replace the wrong one. Do not change the structure or grammar of this sentence.    
	\end{tcolorbox}
	\vspace{-.5em}
	\caption{An Example of Basic Textual Prompt}
	\label{fig:basic_textp}
\end{figure*}

 \subsection{MLLMs for Chart Question Answering}
With the advancements in MLLMs, such as GPT-4o, it becomes increasingly promising to automatically comprehend charts and extract insights according to user queries~\cite{DBLP:journals/tvcg/ShenSLYHZTW23, chartqa, DBLP:conf/chi/ZengB23, huang2024pixels}. 
This process is known as chart question answering, \textit{i.e.} ChartQA for short.
Recent research efforts have focused on understanding the capabilities of MLLMs in performing ChartQA tasks. These studies can be categorized into two groups: evaluation studies and the construction of datasets for ChartQA.

\paragraph{Evaluating \mllms on ChartQA Tasks.}
Several recent studies~\cite{chartreader, enhanced_chart_understanding, LVLMS_understanding_charts, chartbench, chartllama, chartx, unichart} have attempted to leverage the capabilities of \mllms to perform  high-level ChartQA tasks such as chart captioning and chart-to-text.
For example, Huang et al. evaluated the capabilities of representative \mllms, such as \gptfv and Bard (\textit{i.e.} Gemini)~\cite{gemini}, on chart captioning tasks. Their findings indicated that GPT-4o faces challenges in generating captions that accurately reflect the factual information presented in charts. Moreover, these studies have highlighted various promising directions for future research in this field.
Diverging from the emphasis on high-level tasks in previous works, our research uniquely targets \textit{low-level} ChartQA tasks~\cite{DBLP:conf/infovis/AmarES05, lowleveltasks}.

\paragraph{ChartQA Datasets.} 
In the last decade, several ChartQA datasets have been presented~\cite{plotqa, figureqa, chartbench,chartllama,chartqa,dvqa,mmc,chartx,chartreader}, as shown in Section~\ref{sec:overview} Table~\ref{Datasets_comparison}.
For example, ChartBench~\cite{chartbench} includes 2.1K charts for four types of ChartQA tasks.
However, a gap remains evident in the landscape of existing ChartQA datasets: none are tailored to comprehensively evaluate the 10 \textit{low-level} tasks identified as critical to the ChartQA task.
Moreover, to conduct more customized evaluations, such as modifying the visual elements or adding a visual prompt, we need access to the metadata (\textit{e.g.} the underlying data) of the charts, not just the chart images. 
Therefore, we curate a large-scale dataset \tdataset, which consists of a total of 89,388 quartets, each including a chart, a specified task, a corresponding query, and its answer. 

%% file: appendix/prompt.tex
\section{Prompts}
\label{app:prompt}
In this section, we provide detailed descriptions and examples of the prompt strategies used in our evaluations.

\subsection{Basic Textual Prompts}
\label{subsec: textual_prompt}
Each ChartQA task can be framed in four different question types: Fill-in-the-Blank, Error Correction, Multiple-Choice, and Yes-or-No questions. While the core meaning remains the same across these types, Fill-in-the-Blank and Error Correction questions are more open-ended, Multiple-Choice questions require selecting the correct option from several choices, and Yes-or-No questions involve determining the truthfulness of a statement. 
Figure~\ref{fig:basic_textp} shows examples of basic textual prompts for the four question types mentioned above.

\begin{figure*}[t]
	\centering
	\begin{tcolorbox}[colback=black!5, colframe=black!75, title=An Example of RolePlay Prompt]
		\textbf{Initial Question}: What is the minimum value in this chart?
		
		\textbf{Fill-in-the-Blank With RolePlay}: \textcolor{brown}{You are an expert on chart understanding with specialized skills in numerical analysis. Your keen eye for detail allows you to accurately identify and extract numerical values from various chart elements, such as the x-axis/y-axis categories and the legend keys. Your role is to analyze charts, promptly 
			determine the sum or average of specified elements, and communicate your findings in an accessible manner.} What is the minimum value in this chart? Begin your answer with `My answer is [].'
		
		\textbf{Multiple-Choice With RolePlay}: \textcolor{brown}{You are an expert on chart understanding with specialized skills in numerical analysis. Your keen eye for detail allows you to accurately identify and extract numerical values from various chart elements, such as the x-axis/y-axis categories and the legend keys. Your role is to analyze charts, promptly determine the sum or average of specified elements, and communicate your findings in an accessible manner.} What is the minimum value in this chart? NOTE: Choose your answer from the following options [A, B ,C]. Begin your answer with `My option is [Your option].'
		
		\textbf{Yes-or-No With RolePlay:} \textcolor{brown}{You are an expert on chart understanding with specialized skills in numerical analysis. Your keen eye for detail allows you to accurately identify and extract numerical values from various chart elements, such as the x-axis/y-axis categories and the legend keys. Your role is to analyze charts, promptly determine the sum or average of specified elements, and communicate your findings in an accessible manner.} Is the minimum value in this chart equal to ......? NOTE: You only need to answer `Yes' or `No'.
		
		\textbf{Error-Correction With RolePlay:} \textcolor{brown}{You are an expert on chart understanding with specialized skills in numerical analysis. Your keen eye for detail allows you to accurately identify and extract numerical values from various chart elements, such as the x-axis/y-axis categories and the legend keys. Your role is to analyze charts, promptly determine the sum or average of specified elements, and communicate your findings in an accessible manner.} The minimum value in this chart is equal to ...... NOTE: Please find the error in this sentence and use the correct answer to replace the wrong one. Do not change the structure or grammar of this sentence.
	\end{tcolorbox}
	\vspace{-.5em}
	\caption{An Example of RolePlay Prompt}
	\label{fig:roleplayp}
\end{figure*}

\begin{figure*}[t!]
	\centering
	\begin{tcolorbox}[colback=black!5, colframe=black!75, title=An Example of Tutorial Prompt]
		\textbf{Initial Question:} What is the minimum value in this chart?
		
		\textbf{Fill-in-the-Blank With Tutorial:}What is the minimum value in this chart? NOTE: Begin your answer with `My answer is [your answer]'. \textcolor{brown}{Firstly, identify the chart's basic structure and type to understand the visual elements used in the chart and how these elements represent data. Subsequently, observing the chart title, legend, and axes, which provide essential information about the data's theme and measurement units. Next, identify key data points, such as significant highs, lows, or trends. Further steps include comparing relationships between different data series and interpreting the proportions of the data. Finally, summarize the information gathered.}
		
		\textbf{Multiple-Choice With Tutorial:} What is the minimum value in this chart? NOTE: Choose your answer from the following options [A, B ,C]. Begin your answer with `My option is [Your option].' \textcolor{brown}{Firstly, identify the chart's basic structure and type to understand the visual elements used in the chart and how these elements represent data. Subsequently, observing the chart title, legend, and axes, which provide essential information about the data's theme and measurement units. Next, identify key data points, such as significant highs, lows, or trends. Further steps include comparing relationships between different data series and interpreting the proportions of the data. Finally, summarize the information gathered.}
		
		\textbf{Yes-or-No With Tutorial:} Is the minimum value in this chart equal to ......? NOTE: You only need to answer `Yes' or `No'. \textcolor{brown}{Firstly, identify the chart's basic structure and type to understand the visual elements used in the chart and how these elements represent data. Subsequently, observing the chart title, legend, and axes, which provide essential information about the data's theme and measurement units. Next, identify key data points, such as significant highs, lows, or trends. Further steps include comparing relationships between different data series and interpreting the proportions of the data. Finally, summarize the information gathered.}
		
		\textbf{Error-Correction With Tutorial:} The minimum value in this chart is equal to ...... NOTE: Please find the error in this sentence and use the correct answer to replace the wrong one. Do not change the structure or grammar of this sentence. \textcolor{brown}{Firstly, identify the chart's basic structure and type to understand the visual elements used in the chart and how these elements represent data. Subsequently, observing the chart title, legend, and axes, which provide essential information about the data's theme and measurement units. Next, identify key data points, such as significant highs, lows, or trends. Further steps include comparing relationships between different data series and interpreting the proportions of the data. Finally, summarize the information gathered.}
	\end{tcolorbox}
	\vspace{-.5em}
	\caption{An Example of Tutorial Prompt}
	\label{fig:tutorialp}
\end{figure*}

\subsection{Visual Prompts}
\label{app:visual_prompts}

In this paper, we design three types of visual prompts: handwriting, regular shape, and special design.
Figure~\ref{fig:visualprompts} shows examples of these visual prompts in our ChartInsights and evaluations.

\bi
\item \textit{Handwriting Visual Prompts}: These prompts involve manually annotating the relevant visual elements directly on the chart, simulating handwritten notes. This style is particularly useful for tasks like Find Extreme and Data Retrieval, where specific elements need to be located. The handwritten annotations guide the MLLMs to focus on the pertinent parts of the chart.

\item \textit{Regular Shape Visual Prompts}: These prompts use simple geometric shapes, such as circles, rectangles, and arrows, to highlight key areas of the chart. This method provides clear and precise indications of important elements and regions, aiding the MLLMs in understanding the chart structure and data distribution.

\item \textit{Specially Designed Visual Prompts}: These prompts are tailored to specific low-level chart analysis tasks and incorporate customized visual elements that align with the unique requirements of each task. For instance, color-coded overlays or patterned highlights might be used to draw attention to particular data trends or anomalies.
\ei 

We manually annotate these visual prompts to assist MLLMs in understanding the specific requirements of low-level chart analysis tasks. By providing clear and targeted visual cues, we aim to enhance the models' ability to accurately interpret and analyze chart data.


\begin{figure*}[t!]
	\centering
	\begin{tcolorbox}[colback=black!5, colframe=black!75, title=An Example of ChartCoT Prompt]
		\textbf{Initial Question:} What is the minimum value in this chart?
		\textbf{Fill-in-the-Blank With ChartCoT:} \textcolor{brown}{Let's answer following questions one by one: 1. What type is this chart? 2. What are the labels of x-axis? 3. What are the data labels of each element? }4. What is the minimum value in this chart? NOTE: Begin your answer with `My answer is [your answer]'.
		
		\textbf{Multiplt-Choice With ChartCoT:} \textcolor{brown}{Let's answer following questions one by one: 1. What type is this chart? 2. What are the labels of x-axis? 3. What are the data labels of each element? }4. What is the minimum value in this chart? NOTE: Choose your answer from the following options [178747, 95096, 59369]. Begin your answer with `My option is [Your option].'
		
		\textbf{Yes-or-No With ChartCoT:} \textcolor{brown}{Let's answer following questions one by one: 1. What type is this chart? 2. What are the labels of x-axis? 3. What are the data labels of each element?} 4. Is the minimum value in this chart equal to ......? NOTE: You only need to answer `Yes' or `No'.
		
		\textbf{Error-Correction With ChartCoT:} \textcolor{brown}{Let's answer following questions one by one: 1. What type is this chart? 2. What are the labels of x-axis? 3. What are the data labels of each element?} The minimum value in this chart is equal to ...... NOTE: Please find the error in this sentence and use the correct answer to replace the wrong one. Do not change the structure or grammar of this sentence. 
	\end{tcolorbox}
	\vspace{-.5em}
	\caption{An Example of ChartCoT Prompt}
	\label{fig:chartcot}
\end{figure*}

\subsection{RolePlay Prompts}
\label{app:roleplayprompt}

RolePlay prompts guide (multimodal) large language models to adopt specific roles, allowing them to perform tasks in accordance with the behaviors and expertise of those roles~\cite{roleplay}. In this paper, we assign the role of a data visualization expert to the MLLMs. By simulating the thought processes and actions of an expert, the model can better interpret and analyze chart data. This approach helps the model generate more accurate and contextually relevant responses.

Figure~\ref{fig:roleplayp} shows examples of RolePlay prompts for the four question types mentioned above, demonstrating how the model, acting as a data visualization expert, addresses the low-level ChartQA tasks.

\subsection{Tutorial Prompts}
\label{app:tutorialprompt}
Through our experiments, we have found that the more details provided in the input, the more comprehensive and often more accurate the output from MLLMs becomes. Based on this discovery, we propose the Tutorial Textual Prompt. This approach involves breaking down the steps for reading and understanding visualization charts to guide MLLMs through the analysis process.
The Tutorial Prompt provides a detailed, step-by-step explanation of how to interpret various elements of a chart. By explicitly outlining these steps, we aim to enhance the model's ability to process and analyze chart data accurately. This method helps the model to follow a structured approach, ensuring that it considers all relevant aspects of the chart in its analysis.

Figure~\ref{fig:tutorialp} shows examples of Tutorial prompts for the four question types mentioned above.
Specifically, the tutorial might start by instructing the model to identify the type of chart and its key components, such as axes, labels, and legends. It then guides the model through interpreting the data presented in the chart, noting trends, outliers, and significant data points. By providing this structured guidance, the model can generate more precise and contextually relevant responses.

\subsection{ChartCoT Prompts}
\label{app:chartcot}

The design of ChartCoT is based on the concept of Chain of Thought~\cite{chainofthought}, which involves crafting a series of guiding questions to enable MLLMs to produce high-quality responses. This method encourages the model to think step-by-step, enhancing its reasoning capabilities and ensuring a thorough understanding of the chart data.

In our evaluations, we set up three progressively detailed questions to guide the model's thought process:
\bi
\item \textit{Chart Type Identification}: The first question typically pertains to identifying the type of chart. This step ensures that the model correctly understands the basic structure and purpose of the chart, whether it is a bar chart, line chart, pie chart, etc.
\item \textit{Coordinate Information}: The second question relates to the coordinate information of the chart. Here, the model is prompted to recognize and interpret the axes, scales, and any legends or labels that provide context for the data points. This step is crucial for understanding how the data is organized and presented.
\item \textit{Numerical Information}: The third question concerns the numerical information of the elements within the chart. This includes extracting specific data values, identifying trends, and making comparisons between different data points. This step ensures that the model can accurately read and analyze the quantitative aspects of the chart.
\ei

By structuring the prompts in this way, ChartCoT guides the model through a logical progression of understanding, from basic chart recognition to detailed data analysis. This approach helps in generating more accurate and contextually relevant responses.

Figure~\ref{fig:chartcot} shows examples of ChartCoT prompts for the four question types mentioned above. These examples illustrate how the model, guided by a chain of thought, addresses Fill-in-the-Blank, Error Correction, Multiple-Choice, and Yes-or-No questions effectively.

\begin{figure*}[t!]
	\centering
	\begin{tcolorbox}[colback=black!5, colframe=black!75, title=An Example of Chain-of-Charts Prompt]
		\textbf{Initial Question:} What is the minimum value in this chart?
		
		\textbf{Fill-in-the-Blank With Chain-of-Charts:} \textcolor{brown}{Learn from the previous three questions and answers first, and then answer the last question.1. Q: What type is this chart? A: ...... 2. Q: What are the labels of x-axis? A: ...... 3. Q: What are the data labels of each element? A: ......} 4. Q: What is the minimum value in this chart? NOTE: Begin your answer with `My answer is [your answer]'. A: ......
		
		\textbf{Multiplt-Choice With Chain-of-Charts:} \textcolor{brown}{Learn from the previous three questions and answers first, and then answer the last question. 1. Q: What type is this chart? A: ...... 2. Q: What are the labels of x-axis? A: ...... 3. Q: What are the data labels of each element? A: ......} 4. Q: What is the minimum value in this chart? NOTE: Choose your answer from the following options [A, B, C ]. Begin your answer with `My option is [Your option].'
		
		\textbf{Yes-or-No With Chain-of-Charts:} \textcolor{brown}{Learn from the previous three questions and answers first, and then answer the last question.1. Q: What type is this chart? A: ...... 2. Q: What are the labels of x-axis? A: ...... 3. Q: What are the data labels of each element? A: ......} 4. Is the minimum value in this chart equal to ......? NOTE: You only need to answer `Yes' or `No'.
		
		\textbf{Error-Correction With Chain-of-Charts:} \textcolor{brown}{Learn from the previous three questions and answers first, and then answer the last question.1. Q: What type is this chart? A: ...... 2. Q: What are the labels of x-axis? A: ...... 3. Q: What are the data labels of each element? A: ......} 4. The minimum value in this chart is equal to ...... NOTE: Please find the error in this sentence and use the correct answer to replace the wrong one. Do not change the structure or grammar of this sentence. 
	\end{tcolorbox}
	\vspace{-.5em}
	\caption{An Example of Chain-of-Chats Prompt}
	\label{fig:cot}
\end{figure*}

\subsection{Chain-of-Charts Prompts}
\label{app:chain_of_charts_prompt}
Chain-of-Charts is a new textual prompt we have developed based on the ChartCoT method in Appendix~\ref{app:chartcot}. 
In our experiments, we observed that due to hallucinations in MLLMs, merely setting up guiding questions does not always ensure that the model correctly grasps the chart information. 

To address this issue, Chain-of-Charts also includes the answers to each guiding question, providing the model with immediate feedback and reinforcement. By inputting both the guiding questions and their answers, as shown in Figure~\ref{fig:cot}, Chain-of-Charts aims to mitigate the risk of hallucinations and improve the model's comprehension and accuracy. This approach helps the model build a more reliable understanding of the chart data, as it can cross-reference its responses with the provided answers.

The structured processes of Chain-of-Charts are:

\bi
\item \textit{Guiding Questions}: Similar to ChartCoT, we begin with a series of progressively detailed guiding questions, covering chart type identification, coordinate information, and numerical information.

\item \textit{Provided Answers}: For each guiding question, we input the corresponding answer. This step ensures that the model receives immediate clarification and can adjust its understanding based on accurate information.

\item \textit{Enhanced Responses}: By continuously referencing the answers to the guiding questions, the model can generate more accurate and contextually relevant responses for the final task.
\ei



%% file: main.bbl
\begin{thebibliography}{55}
\providecommand{\natexlab}[1]{#1}

\bibitem[{qwe()}]{qwen_plus_max}

\newblock Alibaba qwen-vl.
\newblock \url{https://cn.aliyun.com/}.

\bibitem[{cla()}]{claude3}

\newblock Anthropic claude3.
\newblock \url{https://www.anthropic.com/}.

\bibitem[{gem()}]{gemini}

\newblock Google gemini.
\newblock \url{https://gemini.google.com/app}.
\newblock Accessed: 2024-03-23.

\bibitem[{Omn()}]{Omnilmm}

\newblock Openbmb omnilmm-12b.
\newblock
  \url{https://github.com/OpenBMB/MiniCPM-V?tab=readme-ov-file#omnilmm-12b}.

\bibitem[{cha()}]{chatglm}

\newblock zhipu chatglm-4v.
\newblock \url{https://open.bigmodel.cn/}.

\bibitem[{Amar et~al.(2005)Amar, Eagan, and
  Stasko}]{DBLP:conf/infovis/AmarES05}
Robert Amar, James Eagan, and John Stasko. 2005.
\newblock Low-level components of analytic activity in information
  visualization.
\newblock In \emph{IEEE Symposium on Information Visualization, 2005. INFOVIS
  2005.}, pages 111--117. IEEE.

\bibitem[{Bai et~al.(2023)Bai, Bai, Yang, Wang, Tan, Wang, Lin, Zhou, and
  Zhou}]{qwenvl}
Jinze Bai, Shuai Bai, Shusheng Yang, Shijie Wang, Sinan Tan, Peng Wang, Junyang
  Lin, Chang Zhou, and Jingren Zhou. 2023.
\newblock \href {https://arxiv.org/abs/2308.12966} {Qwen-vl: A versatile
  vision-language model for understanding, localization, text reading, and
  beyond}.
\newblock \emph{Preprint}, arXiv:2308.12966.

\bibitem[{Cai et~al.(2023)Cai, Liu, Mustikovela, Meyer, Chai, Park, and
  Lee}]{vipllava}
Mu~Cai, Haotian Liu, Siva~Karthik Mustikovela, Gregory~P Meyer, Yuning Chai,
  Dennis Park, and Yong~Jae Lee. 2023.
\newblock Making large multimodal models understand arbitrary visual prompts.
\newblock \emph{arXiv preprint arXiv:2312.00784}.

\bibitem[{Chen et~al.(2022)Chen, Guo, Yi, Li, and Elhoseiny}]{visualgpt}
Jun Chen, Han Guo, Kai Yi, Boyang Li, and Mohamed Elhoseiny. 2022.
\newblock \href {https://arxiv.org/abs/2102.10407} {Visualgpt: Data-efficient
  adaptation of pretrained language models for image captioning}.
\newblock \emph{Preprint}, arXiv:2102.10407.

\bibitem[{Cheng et~al.(2023)Cheng, Dai, and Hauptmann}]{chartreader}
Zhi-Qi Cheng, Qi~Dai, and Alexander~G. Hauptmann. 2023.
\newblock Chartreader: A unified framework for chart derendering and
  comprehension without heuristic rules.
\newblock In \emph{Proceedings of the IEEE/CVF International Conference on
  Computer Vision (ICCV)}, pages 22202--22213.

\bibitem[{Fu et~al.(2024)Fu, Chen, Shen, Qin, Zhang, Lin, Yang, Zheng, Li, Sun,
  Wu, and Ji}]{mmeBenchmark}
Chaoyou Fu, Peixian Chen, Yunhang Shen, Yulei Qin, Mengdan Zhang, Xu~Lin,
  Jinrui Yang, Xiawu Zheng, Ke~Li, Xing Sun, Yunsheng Wu, and Rongrong Ji.
  2024.
\newblock \href {https://arxiv.org/abs/2306.13394} {Mme: A comprehensive
  evaluation benchmark for multimodal large language models}.
\newblock \emph{Preprint}, arXiv:2306.13394.

\bibitem[{Han et~al.(2023)Han, Zhang, Chen, Yang, Wang, Yu, Fu, and
  Zhang}]{chartllama}
Yucheng Han, Chi Zhang, Xin Chen, Xu~Yang, Zhibin Wang, Gang Yu, Bin Fu, and
  Hanwang Zhang. 2023.
\newblock \href {https://arxiv.org/abs/2311.16483} {Chartllama: A multimodal
  llm for chart understanding and generation}.
\newblock \emph{Preprint}, arXiv:2311.16483.

\bibitem[{Hu et~al.(2023{\natexlab{a}})Hu, Yao, Wang, Wang, Pan, Chen, Yu, Wu,
  Zhao, Zhang, Han, Lin, Xue, Li, Liu, and Sun}]{viscpm}
Jinyi Hu, Yuan Yao, Chongyi Wang, Shan Wang, Yinxu Pan, Qianyu Chen, Tianyu Yu,
  Hanghao Wu, Yue Zhao, Haoye Zhang, Xu~Han, Yankai Lin, Jiao Xue, Dahai Li,
  Zhiyuan Liu, and Maosong Sun. 2023{\natexlab{a}}.
\newblock Large multilingual models pivot zero-shot multimodal learning across
  languages.
\newblock \emph{arXiv preprint arXiv:2308.12038}.

\bibitem[{Hu et~al.(2024)Hu, Tu, Han, He, Cui, Long, Zheng, Fang, Huang, Zhao,
  Zhang, Thai, Zhang, Wang, Yao, Zhao, Zhou, Cai, Zhai, Ding, Jia, Zeng, Li,
  Liu, and Sun}]{minicpm}
Shengding Hu, Yuge Tu, Xu~Han, Chaoqun He, Ganqu Cui, Xiang Long, Zhi Zheng,
  Yewei Fang, Yuxiang Huang, Weilin Zhao, Xinrong Zhang, Zheng~Leng Thai,
  Kaihuo Zhang, Chongyi Wang, Yuan Yao, Chenyang Zhao, Jie Zhou, Jie Cai,
  Zhongwu Zhai, Ning Ding, Chao Jia, Guoyang Zeng, Dahai Li, Zhiyuan Liu, and
  Maosong Sun. 2024.
\newblock \href {https://arxiv.org/abs/2404.06395} {Minicpm: Unveiling the
  potential of small language models with scalable training strategies}.
\newblock \emph{Preprint}, arXiv:2404.06395.

\bibitem[{Hu et~al.(2023{\natexlab{b}})Hu, Chen, Li, Guo, Wen, Yu, and
  Guo}]{factsbench}
Xuming Hu, Junzhe Chen, Xiaochuan Li, Yufei Guo, Lijie Wen, Philip~S. Yu, and
  Zhijiang Guo. 2023{\natexlab{b}}.
\newblock \href {https://arxiv.org/abs/2310.05177} {Do large language models
  know about facts?}
\newblock \emph{Preprint}, arXiv:2310.05177.

\bibitem[{Huang et~al.(2024{\natexlab{a}})Huang, Chan, Fung, Qiu, Zhou, Joty,
  Chang, and Ji}]{huang2024pixels}
Kung-Hsiang Huang, Hou~Pong Chan, Yi~R. Fung, Haoyi Qiu, Mingyang Zhou, Shafiq
  Joty, Shih-Fu Chang, and Heng Ji. 2024{\natexlab{a}}.
\newblock \href {https://arxiv.org/abs/2403.12027} {From pixels to insights: A
  survey on automatic chart understanding in the era of large foundation
  models}.
\newblock \emph{Preprint}, arXiv:2403.12027.

\bibitem[{Huang et~al.(2023)Huang, Zhou, Chan, Fung, Wang, Zhang, Chang, and
  Ji}]{LVLMS_understanding_charts}
Kung-Hsiang Huang, Mingyang Zhou, Hou~Pong Chan, Yi~R. Fung, Zhenhailong Wang,
  Lingyu Zhang, Shih-Fu Chang, and Heng Ji. 2023.
\newblock \href {https://arxiv.org/abs/2312.10160} {Do lvlms understand charts?
  analyzing and correcting factual errors in chart captioning}.
\newblock \emph{Preprint}, arXiv:2312.10160.

\bibitem[{Huang et~al.(2024{\natexlab{b}})Huang, Yuan, Sheng, Yang, Wu, Chen,
  Yang, Li, and Lin}]{aesbench}
Yipo Huang, Quan Yuan, Xiangfei Sheng, Zhichao Yang, Haoning Wu, Pengfei Chen,
  Yuzhe Yang, Leida Li, and Weisi Lin. 2024{\natexlab{b}}.
\newblock \href {https://arxiv.org/abs/2401.08276} {Aesbench: An expert
  benchmark for multimodal large language models on image aesthetics
  perception}.
\newblock \emph{Preprint}, arXiv:2401.08276.

\bibitem[{Jia et~al.(2022)Jia, Tang, Chen, Cardie, Belongie, Hariharan, and
  Lim}]{jia2022visual}
Menglin Jia, Luming Tang, Bor-Chun Chen, Claire Cardie, Serge Belongie, Bharath
  Hariharan, and Ser-Nam Lim. 2022.
\newblock Visual prompt tuning.
\newblock In \emph{European Conference on Computer Vision}, pages 709--727.
  Springer.

\bibitem[{Kafle et~al.(2018)Kafle, Price, Cohen, and Kanan}]{dvqa}
Kushal Kafle, Brian Price, Scott Cohen, and Christopher Kanan. 2018.
\newblock Dvqa: Understanding data visualizations via question answering.
\newblock In \emph{Proceedings of the IEEE Conference on Computer Vision and
  Pattern Recognition (CVPR)}.

\bibitem[{Kahou et~al.(2018)Kahou, Michalski, Atkinson, Kadar, Trischler, and
  Bengio}]{figureqa}
Samira~Ebrahimi Kahou, Vincent Michalski, Adam Atkinson, Akos Kadar, Adam
  Trischler, and Yoshua Bengio. 2018.
\newblock \href {https://arxiv.org/abs/1710.07300} {Figureqa: An annotated
  figure dataset for visual reasoning}.
\newblock \emph{Preprint}, arXiv:1710.07300.

\bibitem[{Kim et~al.(2023)Kim, Srinivasan, Kim, and Kim}]{chartqaforblind}
Jiho Kim, Arjun Srinivasan, Nam~Wook Kim, and Yea-Seul Kim. 2023.
\newblock Exploring chart question answering for blind and low vision users.
\newblock In \emph{Proceedings of the 2023 CHI Conference on Human Factors in
  Computing Systems}, pages 1--15.

\bibitem[{Kong and Agrawala(2012)}]{overlays}
Nicholas Kong and Maneesh Agrawala. 2012.
\newblock Graphical overlays: Using layered elements to aid chart reading.
\newblock \emph{{IEEE} Trans. Vis. Comput. Graph.}, 18(12):2631--2638.

\bibitem[{Li et~al.(2023{\natexlab{a}})Li, Wang, Wang, Ge, Ge, and
  Shan}]{seedbench}
Bohao Li, Rui Wang, Guangzhi Wang, Yuying Ge, Yixiao Ge, and Ying Shan.
  2023{\natexlab{a}}.
\newblock \href {https://arxiv.org/abs/2307.16125} {Seed-bench: Benchmarking
  multimodal llms with generative comprehension}.
\newblock \emph{Preprint}, arXiv:2307.16125.

\bibitem[{Li et~al.(2023{\natexlab{b}})Li, Li, Savarese, and Hoi}]{blip2}
Junnan Li, Dongxu Li, Silvio Savarese, and Steven Hoi. 2023{\natexlab{b}}.
\newblock Blip-2: Bootstrapping language-image pre-training with frozen image
  encoders and large language models.
\newblock In \emph{International conference on machine learning}, pages
  19730--19742. PMLR.

\bibitem[{Li et~al.(2022)Li, Li, Xiong, and Hoi}]{blip}
Junnan Li, Dongxu Li, Caiming Xiong, and Steven Hoi. 2022.
\newblock \href {https://arxiv.org/abs/2201.12086} {Blip: Bootstrapping
  language-image pre-training for unified vision-language understanding and
  generation}.
\newblock \emph{Preprint}, arXiv:2201.12086.

\bibitem[{Li et~al.(2024)Li, Wang, He, Li, Wang, Liu, Wang, Xu, Chen, Luo,
  Wang, and Qiao}]{mvediobench}
Kunchang Li, Yali Wang, Yinan He, Yizhuo Li, Yi~Wang, Yi~Liu, Zun Wang, Jilan
  Xu, Guo Chen, Ping Luo, Limin Wang, and Yu~Qiao. 2024.
\newblock \href {https://arxiv.org/abs/2311.17005} {Mvbench: A comprehensive
  multi-modal video understanding benchmark}.
\newblock \emph{Preprint}, arXiv:2311.17005.

\bibitem[{Lin et~al.(2023)Lin, Liu, Zhang, Gao, Qiu, Xiao, Qiu, Lin, Shao, Chen
  et~al.}]{sphinx}
Ziyi Lin, Chris Liu, Renrui Zhang, Peng Gao, Longtian Qiu, Han Xiao, Han Qiu,
  Chen Lin, Wenqi Shao, Keqin Chen, et~al. 2023.
\newblock Sphinx: The joint mixing of weights, tasks, and visual embeddings for
  multi-modal large language models.
\newblock \emph{arXiv preprint arXiv:2311.07575}.

\bibitem[{Liu et~al.(2023)Liu, Wang, Yao, Chen, Song, Cho, Yacoob, and
  Yu}]{mmc}
Fuxiao Liu, Xiaoyang Wang, Wenlin Yao, Jianshu Chen, Kaiqiang Song, Sangwoo
  Cho, Yaser Yacoob, and Dong Yu. 2023.
\newblock \href {https://arxiv.org/abs/2311.10774} {Mmc: Advancing multimodal
  chart understanding with large-scale instruction tuning}.
\newblock \emph{Preprint}, arXiv:2311.10774.

\bibitem[{Liu et~al.(2024{\natexlab{a}})Liu, Li, Li, Li, Zhang, Shen, and
  Lee}]{llavanext}
Haotian Liu, Chunyuan Li, Yuheng Li, Bo~Li, Yuanhan Zhang, Sheng Shen, and
  Yong~Jae Lee. 2024{\natexlab{a}}.
\newblock \href {https://llava-vl.github.io/blog/2024-01-30-llava-next/}
  {Llava-next: Improved reasoning, ocr, and world knowledge}.

\bibitem[{Liu et~al.(2024{\natexlab{b}})Liu, Li, Wu, and Lee}]{llava1.5}
Haotian Liu, Chunyuan Li, Qingyang Wu, and Yong~Jae Lee. 2024{\natexlab{b}}.
\newblock Visual instruction tuning.
\newblock \emph{Advances in neural information processing systems}, 36.

\bibitem[{Luo et~al.(2021)Luo, Tang, and Li}]{nvbench}
Yuyu Luo, Jiawei Tang, and Guoliang Li. 2021.
\newblock nvbench: A large-scale synthesized dataset for cross-domain natural
  language to visualization task.
\newblock \emph{arXiv preprint arXiv:2112.12926}.

\bibitem[{Masry et~al.(2023)Masry, Kavehzadeh, Do, Hoque, and Joty}]{unichart}
Ahmed Masry, Parsa Kavehzadeh, Xuan~Long Do, Enamul Hoque, and Shafiq Joty.
  2023.
\newblock \href {https://arxiv.org/abs/2305.14761} {Unichart: A universal
  vision-language pretrained model for chart comprehension and reasoning}.
\newblock \emph{Preprint}, arXiv:2305.14761.

\bibitem[{Masry et~al.(2022)Masry, Long, Tan, Joty, and Hoque}]{chartqa}
Ahmed Masry, Do~Xuan Long, Jia~Qing Tan, Shafiq~R. Joty, and Enamul Hoque.
  2022.
\newblock Chartqa: {A} benchmark for question answering about charts with
  visual and logical reasoning.
\newblock In \emph{ACL (Findings)}, pages 2263--2279. Association for
  Computational Linguistics.

\bibitem[{Meng et~al.(2024)Meng, Shao, Lu, Gao, Zhang, Qiao, and
  Luo}]{chartassisstant}
Fanqing Meng, Wenqi Shao, Quanfeng Lu, Peng Gao, Kaipeng Zhang, Yu~Qiao, and
  Ping Luo. 2024.
\newblock Chartassisstant: A universal chart multimodal language model via
  chart-to-table pre-training and multitask instruction tuning.
\newblock \emph{arXiv preprint arXiv:2401.02384}.

\bibitem[{Methani et~al.(2020)Methani, Ganguly, Khapra, and Kumar}]{plotqa}
Nitesh Methani, Pritha Ganguly, Mitesh~M. Khapra, and Pratyush Kumar. 2020.
\newblock Plotqa: Reasoning over scientific plots.
\newblock In \emph{Proceedings of the IEEE/CVF Winter Conference on
  Applications of Computer Vision (WACV)}.

\bibitem[{Munzner(2014)}]{visualizationanalysis}
Tamara Munzner. 2014.
\newblock \emph{Visualization analysis and design}.
\newblock CRC press.

\bibitem[{Ning et~al.(2023)Ning, Zhu, Xie, Lin, Cui, Yuan, Chen, and
  Yuan}]{videobench2}
Munan Ning, Bin Zhu, Yujia Xie, Bin Lin, Jiaxi Cui, Lu~Yuan, Dongdong Chen, and
  Li~Yuan. 2023.
\newblock \href {https://arxiv.org/abs/2311.16103} {Video-bench: A
  comprehensive benchmark and toolkit for evaluating video-based large language
  models}.
\newblock \emph{Preprint}, arXiv:2311.16103.

\bibitem[{OpenAI(2023)}]{gpt4v}
OpenAI. 2023.
\newblock \href {https://openai.com/research/gpt-4v-system-card}
  {{GPT-4V(ision)} system card}.

\bibitem[{Pan et~al.(2023)Pan, Saxon, Xu, Nathani, Wang, and
  Wang}]{pan2023automatically}
Liangming Pan, Michael Saxon, Wenda Xu, Deepak Nathani, Xinyi Wang, and
  William~Yang Wang. 2023.
\newblock Automatically correcting large language models: Surveying the
  landscape of diverse self-correction strategies.
\newblock \emph{arXiv preprint arXiv:2308.03188}.

\bibitem[{Radford et~al.(2021)Radford, Kim, Hallacy, Ramesh, Goh, Agarwal,
  Sastry, Askell, Mishkin, Clark, Krueger, and Sutskever}]{clip}
Alec Radford, Jong~Wook Kim, Chris Hallacy, Aditya Ramesh, Gabriel Goh,
  Sandhini Agarwal, Girish Sastry, Amanda Askell, Pamela Mishkin, Jack Clark,
  Gretchen Krueger, and Ilya Sutskever. 2021.
\newblock \href {https://arxiv.org/abs/2103.00020} {Learning transferable
  visual models from natural language supervision}.
\newblock \emph{Preprint}, arXiv:2103.00020.

\bibitem[{Ramesh et~al.(2022)Ramesh, Dhariwal, Nichol, Chu, and Chen}]{dalle2}
Aditya Ramesh, Prafulla Dhariwal, Alex Nichol, Casey Chu, and Mark Chen. 2022.
\newblock \href {https://arxiv.org/abs/2204.06125} {Hierarchical
  text-conditional image generation with clip latents}.
\newblock \emph{Preprint}, arXiv:2204.06125.

\bibitem[{Saket et~al.(2019)Saket, Endert, and Demiralp}]{lowleveltasks}
Bahador Saket, Alex Endert, and {\c{C}}agatay Demiralp. 2019.
\newblock Task-based effectiveness of basic visualizations.
\newblock \emph{{IEEE} Trans. Vis. Comput. Graph.}, 25(7):2505--2512.

\bibitem[{Shanahan et~al.(2023)Shanahan, McDonell, and
  Reynolds}]{DBLP:journals/nature/ShanahanMR23}
Murray Shanahan, Kyle McDonell, and Laria Reynolds. 2023.
\newblock Role play with large language models.
\newblock \emph{Nat.}, 623(7987):493--498.

\bibitem[{Shen et~al.(2023)Shen, Shen, Luo, Yang, Hu, Zhang, Tai, and
  Wang}]{DBLP:journals/tvcg/ShenSLYHZTW23}
Leixian Shen, Enya Shen, Yuyu Luo, Xiaocong Yang, Xuming Hu, Xiongshuai Zhang,
  Zhiwei Tai, and Jianmin Wang. 2023.
\newblock Towards natural language interfaces for data visualization: {A}
  survey.
\newblock \emph{IEEE Trans. Vis. Comput. Graph.}, 29(6):3121--3144.

\bibitem[{Vaswani et~al.(2023)Vaswani, Shazeer, Parmar, Uszkoreit, Jones,
  Gomez, Kaiser, and Polosukhin}]{transformer}
Ashish Vaswani, Noam Shazeer, Niki Parmar, Jakob Uszkoreit, Llion Jones,
  Aidan~N. Gomez, Lukasz Kaiser, and Illia Polosukhin. 2023.
\newblock \href {https://arxiv.org/abs/1706.03762} {Attention is all you need}.
\newblock \emph{Preprint}, arXiv:1706.03762.

\bibitem[{Wang et~al.(2024)Wang, Lv, Yu, Hong, Qi, Wang, Ji, Yang, Zhao, Song,
  Xu, Xu, Li, Dong, Ding, and Tang}]{cogvlm}
Weihan Wang, Qingsong Lv, Wenmeng Yu, Wenyi Hong, Ji~Qi, Yan Wang, Junhui Ji,
  Zhuoyi Yang, Lei Zhao, Xixuan Song, Jiazheng Xu, Bin Xu, Juanzi Li, Yuxiao
  Dong, Ming Ding, and Jie Tang. 2024.
\newblock \href {https://arxiv.org/abs/2311.03079} {Cogvlm: Visual expert for
  pretrained language models}.
\newblock \emph{Preprint}, arXiv:2311.03079.

\bibitem[{Wei et~al.(2022)Wei, Wang, Schuurmans, Bosma, Ichter, Xia, Chi, Le,
  and Zhou}]{chainofthought}
Jason Wei, Xuezhi Wang, Dale Schuurmans, Maarten Bosma, Brian Ichter, Fei Xia,
  Ed~H. Chi, Quoc~V. Le, and Denny Zhou. 2022.
\newblock Chain-of-thought prompting elicits reasoning in large language
  models.
\newblock In \emph{NeurIPS}.

\bibitem[{Wu et~al.(2023)Wu, Bansal, Zhang, Wu, Zhang, Zhu, Li, Jiang, Zhang,
  and Wang}]{wu2023autogen}
Qingyun Wu, Gagan Bansal, Jieyu Zhang, Yiran Wu, Shaokun Zhang, Erkang Zhu,
  Beibin Li, Li~Jiang, Xiaoyun Zhang, and Chi Wang. 2023.
\newblock Autogen: Enabling next-gen llm applications via multi-agent
  conversation framework.
\newblock \emph{arXiv preprint arXiv:2308.08155}.

\bibitem[{Xia et~al.(2024)Xia, Zhang, Ye, Yan, Liu, Zhou, Chen, Dou, Shi, Yan
  et~al.}]{chartx}
Renqiu Xia, Bo~Zhang, Hancheng Ye, Xiangchao Yan, Qi~Liu, Hongbin Zhou, Zijun
  Chen, Min Dou, Botian Shi, Junchi Yan, et~al. 2024.
\newblock Chartx \& chartvlm: A versatile benchmark and foundation model for
  complicated chart reasoning.
\newblock \emph{arXiv preprint arXiv:2402.12185}.

\bibitem[{Xu et~al.(2023)Xu, Du, Qi, Xu, Yuan, and Guo}]{chartbench}
Zhengzhuo Xu, Sinan Du, Yiyan Qi, Chengjin Xu, Chun Yuan, and Jian Guo. 2023.
\newblock Chartbench: A benchmark for complex visual reasoning in charts.
\newblock \emph{arXiv preprint arXiv:2312.15915}.

\bibitem[{Ye et~al.(2023)Ye, Xu, Ye, Yan, Hu, Liu, Qian, Zhang, Huang, and
  Zhou}]{mplugowl2}
Qinghao Ye, Haiyang Xu, Jiabo Ye, Ming Yan, Anwen Hu, Haowei Liu, Qi~Qian,
  Ji~Zhang, Fei Huang, and Jingren Zhou. 2023.
\newblock \href {https://arxiv.org/abs/2311.04257} {mplug-owl2: Revolutionizing
  multi-modal large language model with modality collaboration}.
\newblock \emph{Preprint}, arXiv:2311.04257.

\bibitem[{Ye et~al.(2024)Ye, Hao, Hou, Wang, Xiao, Luo, and
  Zeng}]{ye2024generative}
Yilin Ye, Jianing Hao, Yihan Hou, Zhan Wang, Shishi Xiao, Yuyu Luo, and Wei
  Zeng. 2024.
\newblock \href {https://arxiv.org/abs/2404.18144} {Generative ai for
  visualization: State of the art and future directions}.
\newblock \emph{Preprint}, arXiv:2404.18144.

\bibitem[{Zeng and Battle(2023)}]{DBLP:conf/chi/ZengB23}
Zehua Zeng and Leilani Battle. 2023.
\newblock A review and collation of graphical perception knowledge for
  visualization recommendation.
\newblock In \emph{{CHI}}, pages 820:1--820:16. {ACM}.

\bibitem[{Zhou et~al.(2023)Zhou, Fung, Chen, Thomas, Ji, and
  Chang}]{enhanced_chart_understanding}
Mingyang Zhou, Yi~R. Fung, Long Chen, Christopher Thomas, Heng Ji, and Shih-Fu
  Chang. 2023.
\newblock \href {https://arxiv.org/abs/2305.18641} {Enhanced chart
  understanding in vision and language task via cross-modal pre-training on
  plot table pairs}.
\newblock \emph{Preprint}, arXiv:2305.18641.

\end{thebibliography}


\begin{thebibliography}{70}
\providecommand{\natexlab}[1]{#1}

\bibitem[{Amar et~al.(2005)Amar, Eagan, and
  Stasko}]{DBLP:conf/infovis/AmarES05}
Robert Amar, James Eagan, and John Stasko. 2005.
\newblock Low-level components of analytic activity in information
  visualization.
\newblock In \emph{IEEE Symposium on Information Visualization, 2005. INFOVIS
  2005.}, pages 111--117. IEEE.

\bibitem[{Anthropic(2024)}]{claude3}
Anthropic. 2024.
\newblock \href {https://www.anthropic.com/} {Claude 3}.

\bibitem[{Bai et~al.(2023)Bai, Bai, Yang, Wang, Tan, Wang, Lin, Zhou, and
  Zhou}]{qwenvl}
Jinze Bai, Shuai Bai, Shusheng Yang, Shijie Wang, Sinan Tan, Peng Wang, Junyang
  Lin, Chang Zhou, and Jingren Zhou. 2023.
\newblock \href {https://arxiv.org/abs/2308.12966} {Qwen-vl: A versatile
  vision-language model for understanding, localization, text reading, and
  beyond}.
\newblock \emph{Preprint}, arXiv:2308.12966.

\bibitem[{Cai et~al.(2023)Cai, Liu, Mustikovela, Meyer, Chai, Park, and
  Lee}]{vipllava}
Mu~Cai, Haotian Liu, Siva~Karthik Mustikovela, Gregory~P Meyer, Yuning Chai,
  Dennis Park, and Yong~Jae Lee. 2023.
\newblock Making large multimodal models understand arbitrary visual prompts.
\newblock \emph{arXiv preprint arXiv:2312.00784}.

\bibitem[{Chen et~al.(2022)Chen, Guo, Yi, Li, and Elhoseiny}]{visualgpt}
Jun Chen, Han Guo, Kai Yi, Boyang Li, and Mohamed Elhoseiny. 2022.
\newblock \href {https://arxiv.org/abs/2102.10407} {Visualgpt: Data-efficient
  adaptation of pretrained language models for image captioning}.
\newblock \emph{Preprint}, arXiv:2102.10407.

\bibitem[{Cheng et~al.(2023)Cheng, Dai, and Hauptmann}]{chartreader}
Zhi-Qi Cheng, Qi~Dai, and Alexander~G. Hauptmann. 2023.
\newblock Chartreader: A unified framework for chart derendering and
  comprehension without heuristic rules.
\newblock In \emph{Proceedings of the IEEE/CVF International Conference on
  Computer Vision (ICCV)}, pages 22202--22213.

\bibitem[{Fu et~al.(2024)Fu, Chen, Shen, Qin, Zhang, Lin, Yang, Zheng, Li, Sun,
  Wu, and Ji}]{mmeBenchmark}
Chaoyou Fu, Peixian Chen, Yunhang Shen, Yulei Qin, Mengdan Zhang, Xu~Lin,
  Jinrui Yang, Xiawu Zheng, Ke~Li, Xing Sun, Yunsheng Wu, and Rongrong Ji.
  2024.
\newblock \href {https://arxiv.org/abs/2306.13394} {Mme: A comprehensive
  evaluation benchmark for multimodal large language models}.
\newblock \emph{Preprint}, arXiv:2306.13394.

\bibitem[{Han et~al.(2023)Han, Zhang, Chen, Yang, Wang, Yu, Fu, and
  Zhang}]{chartllama}
Yucheng Han, Chi Zhang, Xin Chen, Xu~Yang, Zhibin Wang, Gang Yu, Bin Fu, and
  Hanwang Zhang. 2023.
\newblock \href {https://arxiv.org/abs/2311.16483} {Chartllama: A multimodal
  llm for chart understanding and generation}.
\newblock \emph{Preprint}, arXiv:2311.16483.

\bibitem[{Hu et~al.(2023{\natexlab{a}})Hu, Yao, Wang, Wang, Pan, Chen, Yu, Wu,
  Zhao, Zhang, Han, Lin, Xue, Li, Liu, and Sun}]{viscpm}
Jinyi Hu, Yuan Yao, Chongyi Wang, Shan Wang, Yinxu Pan, Qianyu Chen, Tianyu Yu,
  Hanghao Wu, Yue Zhao, Haoye Zhang, Xu~Han, Yankai Lin, Jiao Xue, Dahai Li,
  Zhiyuan Liu, and Maosong Sun. 2023{\natexlab{a}}.
\newblock Large multilingual models pivot zero-shot multimodal learning across
  languages.
\newblock \emph{arXiv preprint arXiv:2308.12038}.

\bibitem[{Hu et~al.(2024)Hu, Tu, Han, He, Cui, Long, Zheng, Fang, Huang, Zhao,
  Zhang, Thai, Zhang, Wang, Yao, Zhao, Zhou, Cai, Zhai, Ding, Jia, Zeng, Li,
  Liu, and Sun}]{minicpm}
Shengding Hu, Yuge Tu, Xu~Han, Chaoqun He, Ganqu Cui, Xiang Long, Zhi Zheng,
  Yewei Fang, Yuxiang Huang, Weilin Zhao, Xinrong Zhang, Zheng~Leng Thai,
  Kaihuo Zhang, Chongyi Wang, Yuan Yao, Chenyang Zhao, Jie Zhou, Jie Cai,
  Zhongwu Zhai, Ning Ding, Chao Jia, Guoyang Zeng, Dahai Li, Zhiyuan Liu, and
  Maosong Sun. 2024.
\newblock \href {https://arxiv.org/abs/2404.06395} {Minicpm: Unveiling the
  potential of small language models with scalable training strategies}.
\newblock \emph{Preprint}, arXiv:2404.06395.

\bibitem[{Hu et~al.(2023{\natexlab{b}})Hu, Chen, Li, Guo, Wen, Yu, and
  Guo}]{factsbench}
Xuming Hu, Junzhe Chen, Xiaochuan Li, Yufei Guo, Lijie Wen, Philip~S. Yu, and
  Zhijiang Guo. 2023{\natexlab{b}}.
\newblock \href {https://arxiv.org/abs/2310.05177} {Do large language models
  know about facts?}
\newblock \emph{Preprint}, arXiv:2310.05177.

\bibitem[{Huang et~al.(2024{\natexlab{a}})Huang, Chan, Fung, Qiu, Zhou, Joty,
  Chang, and Ji}]{huang2024pixels}
Kung-Hsiang Huang, Hou~Pong Chan, Yi~R. Fung, Haoyi Qiu, Mingyang Zhou, Shafiq
  Joty, Shih-Fu Chang, and Heng Ji. 2024{\natexlab{a}}.
\newblock \href {https://arxiv.org/abs/2403.12027} {From pixels to insights: A
  survey on automatic chart understanding in the era of large foundation
  models}.
\newblock \emph{Preprint}, arXiv:2403.12027.

\bibitem[{Huang et~al.(2023)Huang, Zhou, Chan, Fung, Wang, Zhang, Chang, and
  Ji}]{LVLMS_understanding_charts}
Kung-Hsiang Huang, Mingyang Zhou, Hou~Pong Chan, Yi~R. Fung, Zhenhailong Wang,
  Lingyu Zhang, Shih-Fu Chang, and Heng Ji. 2023.
\newblock \href {https://arxiv.org/abs/2312.10160} {Do lvlms understand charts?
  analyzing and correcting factual errors in chart captioning}.
\newblock \emph{Preprint}, arXiv:2312.10160.

\bibitem[{Huang et~al.(2024{\natexlab{b}})Huang, Yuan, Sheng, Yang, Wu, Chen,
  Yang, Li, and Lin}]{aesbench}
Yipo Huang, Quan Yuan, Xiangfei Sheng, Zhichao Yang, Haoning Wu, Pengfei Chen,
  Yuzhe Yang, Leida Li, and Weisi Lin. 2024{\natexlab{b}}.
\newblock \href {https://arxiv.org/abs/2401.08276} {Aesbench: An expert
  benchmark for multimodal large language models on image aesthetics
  perception}.
\newblock \emph{Preprint}, arXiv:2401.08276.

\bibitem[{Kafle et~al.(2018)Kafle, Price, Cohen, and Kanan}]{dvqa}
Kushal Kafle, Brian Price, Scott Cohen, and Christopher Kanan. 2018.
\newblock Dvqa: Understanding data visualizations via question answering.
\newblock In \emph{Proceedings of the IEEE Conference on Computer Vision and
  Pattern Recognition (CVPR)}.

\bibitem[{Kahou et~al.(2018)Kahou, Michalski, Atkinson, Kadar, Trischler, and
  Bengio}]{figureqa}
Samira~Ebrahimi Kahou, Vincent Michalski, Adam Atkinson, Akos Kadar, Adam
  Trischler, and Yoshua Bengio. 2018.
\newblock \href {https://arxiv.org/abs/1710.07300} {Figureqa: An annotated
  figure dataset for visual reasoning}.
\newblock \emph{Preprint}, arXiv:1710.07300.

\bibitem[{Kim et~al.(2023)Kim, Srinivasan, Kim, and Kim}]{chartqaforblind}
Jiho Kim, Arjun Srinivasan, Nam~Wook Kim, and Yea-Seul Kim. 2023.
\newblock Exploring chart question answering for blind and low vision users.
\newblock In \emph{Proceedings of the 2023 CHI Conference on Human Factors in
  Computing Systems (ACM CHI)}, pages 1--15.

\bibitem[{Kong and Agrawala(2012)}]{overlays}
Nicholas Kong and Maneesh Agrawala. 2012.
\newblock Graphical overlays: Using layered elements to aid chart reading.
\newblock \emph{{IEEE} Trans. Vis. Comput. Graph.}, 18(12):2631--2638.

\bibitem[{Li et~al.(2023{\natexlab{a}})Li, Wang, Wang, Ge, Ge, and
  Shan}]{seedbench}
Bohao Li, Rui Wang, Guangzhi Wang, Yuying Ge, Yixiao Ge, and Ying Shan.
  2023{\natexlab{a}}.
\newblock \href {https://arxiv.org/abs/2307.16125} {Seed-bench: Benchmarking
  multimodal llms with generative comprehension}.
\newblock \emph{Preprint}, arXiv:2307.16125.

\bibitem[{Li et~al.(2024{\natexlab{a}})Li, Luo, Chai, Li, and
  Tang}]{DBLP:journals/pvldb/LiLCLT24}
Boyan Li, Yuyu Luo, Chengliang Chai, Guoliang Li, and Nan Tang.
  2024{\natexlab{a}}.
\newblock The dawn of natural language to {SQL:} are we fully ready?
\newblock \emph{Proc. {VLDB} Endow.}, 17(11):3318--3331.

\bibitem[{Li et~al.(2024{\natexlab{b}})Li, Li, Feng, Zhang, Luo, and
  Liu}]{DBLP:journals/tvcg/LiLFZLL24}
Guozheng Li, Runfei Li, Yunshan Feng, Yu~Zhang, Yuyu Luo, and Chi~Harold Liu.
  2024{\natexlab{b}}.
\newblock Coinsight: Visual storytelling for hierarchical tables with connected
  insights.
\newblock \emph{{IEEE} Trans. Vis. Comput. Graph.}, 30(6):3049--3061.

\bibitem[{Li et~al.(2023{\natexlab{b}})Li, Li, Savarese, and Hoi}]{blip2}
Junnan Li, Dongxu Li, Silvio Savarese, and Steven Hoi. 2023{\natexlab{b}}.
\newblock Blip-2: Bootstrapping language-image pre-training with frozen image
  encoders and large language models.
\newblock In \emph{International conference on machine learning (ICML)}, pages
  19730--19742. PMLR.

\bibitem[{Li et~al.(2022)Li, Li, Xiong, and Hoi}]{blip}
Junnan Li, Dongxu Li, Caiming Xiong, and Steven Hoi. 2022.
\newblock \href {https://arxiv.org/abs/2201.12086} {Blip: Bootstrapping
  language-image pre-training for unified vision-language understanding and
  generation}.
\newblock \emph{Preprint}, arXiv:2201.12086.

\bibitem[{Li et~al.(2024{\natexlab{c}})Li, Wang, He, Li, Wang, Liu, Wang, Xu,
  Chen, Luo, Wang, and Qiao}]{mvediobench}
Kunchang Li, Yali Wang, Yinan He, Yizhuo Li, Yi~Wang, Yi~Liu, Zun Wang, Jilan
  Xu, Guo Chen, Ping Luo, Limin Wang, and Yu~Qiao. 2024{\natexlab{c}}.
\newblock \href {https://arxiv.org/abs/2311.17005} {Mvbench: A comprehensive
  multi-modal video understanding benchmark}.
\newblock \emph{Preprint}, arXiv:2311.17005.

\bibitem[{Lin et~al.(2023)Lin, Liu, Zhang, Gao, Qiu, Xiao, Qiu, Lin, Shao, Chen
  et~al.}]{sphinx}
Ziyi Lin, Chris Liu, Renrui Zhang, Peng Gao, Longtian Qiu, Han Xiao, Han Qiu,
  Chen Lin, Wenqi Shao, Keqin Chen, et~al. 2023.
\newblock Sphinx: The joint mixing of weights, tasks, and visual embeddings for
  multi-modal large language models.
\newblock \emph{arXiv preprint arXiv:2311.07575}.

\bibitem[{Liu et~al.(2023)Liu, Wang, Yao, Chen, Song, Cho, Yacoob, and
  Yu}]{mmc}
Fuxiao Liu, Xiaoyang Wang, Wenlin Yao, Jianshu Chen, Kaiqiang Song, Sangwoo
  Cho, Yaser Yacoob, and Dong Yu. 2023.
\newblock \href {https://arxiv.org/abs/2311.10774} {Mmc: Advancing multimodal
  chart understanding with large-scale instruction tuning}.
\newblock \emph{Preprint}, arXiv:2311.10774.

\bibitem[{Liu et~al.(2024{\natexlab{a}})Liu, Li, Li, Li, Zhang, Shen, and
  Lee}]{llavanext}
Haotian Liu, Chunyuan Li, Yuheng Li, Bo~Li, Yuanhan Zhang, Sheng Shen, and
  Yong~Jae Lee. 2024{\natexlab{a}}.
\newblock \href {https://llava-vl.github.io/blog/2024-01-30-llava-next/}
  {Llava-next: Improved reasoning, ocr, and world knowledge}.

\bibitem[{Liu et~al.(2024{\natexlab{b}})Liu, Li, Wu, and Lee}]{llava1.5}
Haotian Liu, Chunyuan Li, Qingyang Wu, and Yong~Jae Lee. 2024{\natexlab{b}}.
\newblock Visual instruction tuning.
\newblock \emph{Advances in neural information processing systems (NeurIPS)},
  36.

\bibitem[{Liu et~al.(2024{\natexlab{c}})Liu, Shen, Li, Ma, Jiang, Luo, Zhang,
  Fan, Li, and Tang}]{DBLP:journals/corr/abs-2408-05109}
Xinyu Liu, Shuyu Shen, Boyan Li, Peixian Ma, Runzhi Jiang, Yuyu Luo, Yuxin
  Zhang, Ju~Fan, Guoliang Li, and Nan Tang. 2024{\natexlab{c}}.
\newblock A survey of {NL2SQL} with large language models: Where are we, and
  where are we going?
\newblock \emph{CoRR}, abs/2408.05109.

\bibitem[{Lu et~al.(2024)Lu, Bansal, Xia, Liu, Li, Hajishirzi, Cheng, Chang,
  Galley, and Gao}]{mathvista}
Pan Lu, Hritik Bansal, Tony Xia, Jiacheng Liu, Chunyuan Li, Hannaneh
  Hajishirzi, Hao Cheng, Kai-Wei Chang, Michel Galley, and Jianfeng Gao. 2024.
\newblock \href {https://arxiv.org/abs/2310.02255} {Mathvista: Evaluating
  mathematical reasoning of foundation models in visual contexts}.
\newblock \emph{Preprint}, arXiv:2310.02255.

\bibitem[{Luo et~al.(2020{\natexlab{a}})Luo, Chai, Qin, Tang, and
  Li}]{DBLP:conf/icde/LuoCQ0020}
Yuyu Luo, Chengliang Chai, Xuedi Qin, Nan Tang, and Guoliang Li.
  2020{\natexlab{a}}.
\newblock Interactive cleaning for progressive visualization through composite
  questions.
\newblock In \emph{{ICDE}}, pages 733--744. {IEEE}.

\bibitem[{Luo et~al.(2020{\natexlab{b}})Luo, Chai, Qin, Tang, and
  Li}]{DBLP:journals/pvldb/LuoCQ0020}
Yuyu Luo, Chengliang Chai, Xuedi Qin, Nan Tang, and Guoliang Li.
  2020{\natexlab{b}}.
\newblock Visclean: Interactive cleaning for progressive visualization.
\newblock \emph{Proc. {VLDB} Endow.}, 13(12):2821--2824.

\bibitem[{Luo et~al.(2020{\natexlab{c}})Luo, Li, Zhao, Yu, Zhang, Li, and
  Tang}]{DBLP:journals/pvldb/LuoLZYZ0020}
Yuyu Luo, Wenbo Li, Tianyu Zhao, Xiang Yu, Lixi Zhang, Guoliang Li, and Nan
  Tang. 2020{\natexlab{c}}.
\newblock Deeptrack: Monitoring and exploring spatio-temporal data - {A} case
  of tracking {COVID-19} -.
\newblock \emph{Proc. {VLDB} Endow.}, 13(12):2841--2844.

\bibitem[{Luo et~al.(2022{\natexlab{a}})Luo, Qin, Chai, Tang, Li, and
  Li}]{DBLP:journals/tkde/LuoQCTLL22}
Yuyu Luo, Xuedi Qin, Chengliang Chai, Nan Tang, Guoliang Li, and Wenbo Li.
  2022{\natexlab{a}}.
\newblock Steerable self-driving data visualization.
\newblock \emph{{IEEE} Trans. Knowl. Data Eng.}, 34(1):475--490.

\bibitem[{Luo et~al.(2018{\natexlab{a}})Luo, Qin, Tang, and
  Li}]{DBLP:conf/icde/LuoQ0018}
Yuyu Luo, Xuedi Qin, Nan Tang, and Guoliang Li. 2018{\natexlab{a}}.
\newblock Deepeye: Towards automatic data visualization.
\newblock In \emph{{ICDE}}, pages 101--112. {IEEE} Computer Society.

\bibitem[{Luo et~al.(2018{\natexlab{b}})Luo, Qin, Tang, Li, and
  Wang}]{DBLP:conf/sigmod/LuoQ00W18}
Yuyu Luo, Xuedi Qin, Nan Tang, Guoliang Li, and Xinran Wang.
  2018{\natexlab{b}}.
\newblock Deepeye: Creating good data visualizations by keyword search.
\newblock In \emph{{SIGMOD} Conference}, pages 1733--1736. {ACM}.

\bibitem[{Luo et~al.(2021)Luo, Tang, Li, Chai, Li, and
  Qin}]{DBLP:conf/sigmod/Luo00CLQ21}
Yuyu Luo, Nan Tang, Guoliang Li, Chengliang Chai, Wenbo Li, and Xuedi Qin.
  2021.
\newblock Synthesizing natural language to visualization {(NL2VIS)} benchmarks
  from {NL2SQL} benchmarks.
\newblock In \emph{{SIGMOD} Conference}, pages 1235--1247. {ACM}.

\bibitem[{Luo et~al.(2020{\natexlab{d}})Luo, Tang, Li, Li, Zhao, and
  Yu}]{DBLP:journals/debu/Luo00LZY20}
Yuyu Luo, Nan Tang, Guoliang Li, Wenbo Li, Tianyu Zhao, and Xiang Yu.
  2020{\natexlab{d}}.
\newblock Deepeye: {A} data science system for monitoring and exploring
  {COVID-19} data.
\newblock \emph{{IEEE} Data Eng. Bull.}, 43(2):121--132.

\bibitem[{Luo et~al.(2022{\natexlab{b}})Luo, Tang, Li, Tang, Chai, and
  Qin}]{DBLP:journals/tvcg/LuoTLTCQ22}
Yuyu Luo, Nan Tang, Guoliang Li, Jiawei Tang, Chengliang Chai, and Xuedi Qin.
  2022{\natexlab{b}}.
\newblock Natural language to visualization by neural machine translation.
\newblock \emph{{IEEE} Trans. Vis. Comput. Graph.}, 28(1):217--226.

\bibitem[{Luo et~al.(2023)Luo, Zhou, Tang, Li, Chai, and
  Shen}]{DBLP:journals/pacmmod/LuoZ00CS23}
Yuyu Luo, Yihui Zhou, Nan Tang, Guoliang Li, Chengliang Chai, and Leixian Shen.
  2023.
\newblock Learned data-aware image representations of line charts for
  similarity search.
\newblock \emph{Proc. {ACM} Manag. Data}, 1(1):88:1--88:29.

\bibitem[{Masry et~al.(2023)Masry, Kavehzadeh, Do, Hoque, and Joty}]{unichart}
Ahmed Masry, Parsa Kavehzadeh, Xuan~Long Do, Enamul Hoque, and Shafiq Joty.
  2023.
\newblock \href {https://arxiv.org/abs/2305.14761} {Unichart: A universal
  vision-language pretrained model for chart comprehension and reasoning}.
\newblock \emph{Preprint}, arXiv:2305.14761.

\bibitem[{Masry et~al.(2022)Masry, Long, Tan, Joty, and Hoque}]{chartqa}
Ahmed Masry, Do~Xuan Long, Jia~Qing Tan, Shafiq~R. Joty, and Enamul Hoque.
  2022.
\newblock Chartqa: {A} benchmark for question answering about charts with
  visual and logical reasoning.
\newblock In \emph{ACL (Findings)}, pages 2263--2279. Association for
  Computational Linguistics.

\bibitem[{Meng et~al.(2024)Meng, Shao, Lu, Gao, Zhang, Qiao, and
  Luo}]{chartassisstant}
Fanqing Meng, Wenqi Shao, Quanfeng Lu, Peng Gao, Kaipeng Zhang, Yu~Qiao, and
  Ping Luo. 2024.
\newblock Chartassisstant: A universal chart multimodal language model via
  chart-to-table pre-training and multitask instruction tuning.
\newblock \emph{arXiv preprint arXiv:2401.02384}.

\bibitem[{Methani et~al.(2020)Methani, Ganguly, Khapra, and Kumar}]{plotqa}
Nitesh Methani, Pritha Ganguly, Mitesh~M. Khapra, and Pratyush Kumar. 2020.
\newblock Plotqa: Reasoning over scientific plots.
\newblock In \emph{Proceedings of the IEEE/CVF Winter Conference on
  Applications of Computer Vision (WACV)}.

\bibitem[{Modelbest(2024)}]{Omnilmm}
Modelbest. 2024.
\newblock Openbmb omnilmm-12b.
\newblock \url{https://github.com/OpenBMB/OmniLMM}.

\bibitem[{Munzner(2014)}]{visualizationanalysis}
Tamara Munzner. 2014.
\newblock \emph{Visualization analysis and design}.
\newblock CRC press.

\bibitem[{Ning et~al.(2023)Ning, Zhu, Xie, Lin, Cui, Yuan, Chen, and
  Yuan}]{videobench2}
Munan Ning, Bin Zhu, Yujia Xie, Bin Lin, Jiaxi Cui, Lu~Yuan, Dongdong Chen, and
  Li~Yuan. 2023.
\newblock \href {https://arxiv.org/abs/2311.16103} {Video-bench: A
  comprehensive benchmark and toolkit for evaluating video-based large language
  models}.
\newblock \emph{Preprint}, arXiv:2311.16103.

\bibitem[{OpenAI(2023)}]{gpt4v}
OpenAI. 2023.
\newblock \href {https://openai.com/research/gpt-4v-system-card}
  {{GPT-4V(ision)} system card}.

\bibitem[{OpenAI et~al.(2024)OpenAI, Achiam, Adler, Agarwal, and
  et~al.}]{openai2024gpt4}
OpenAI, Josh Achiam, Steven Adler, Sandhini Agarwal, and et~al. 2024.
\newblock \href {https://arxiv.org/abs/2303.08774} {Gpt-4 technical report}.
\newblock \emph{Preprint}, arXiv:2303.08774.

\bibitem[{Qin et~al.(2020)Qin, Luo, Tang, and Li}]{DBLP:journals/vldb/QinLTL20}
Xuedi Qin, Yuyu Luo, Nan Tang, and Guoliang Li. 2020.
\newblock Making data visualization more efficient and effective: a survey.
\newblock \emph{{VLDB} J.}, 29(1):93--117.

\bibitem[{Radford et~al.(2021)Radford, Kim, Hallacy, Ramesh, Goh, Agarwal,
  Sastry, Askell, Mishkin, Clark, Krueger, and Sutskever}]{clip}
Alec Radford, Jong~Wook Kim, Chris Hallacy, Aditya Ramesh, Gabriel Goh,
  Sandhini Agarwal, Girish Sastry, Amanda Askell, Pamela Mishkin, Jack Clark,
  Gretchen Krueger, and Ilya Sutskever. 2021.
\newblock \href {https://arxiv.org/abs/2103.00020} {Learning transferable
  visual models from natural language supervision}.
\newblock \emph{Preprint}, arXiv:2103.00020.

\bibitem[{Ramesh et~al.(2022)Ramesh, Dhariwal, Nichol, Chu, and Chen}]{dalle2}
Aditya Ramesh, Prafulla Dhariwal, Alex Nichol, Casey Chu, and Mark Chen. 2022.
\newblock \href {https://arxiv.org/abs/2204.06125} {Hierarchical
  text-conditional image generation with clip latents}.
\newblock \emph{Preprint}, arXiv:2204.06125.

\bibitem[{Saket et~al.(2019)Saket, Endert, and Demiralp}]{lowleveltasks}
Bahador Saket, Alex Endert, and {\c{C}}agatay Demiralp. 2019.
\newblock Task-based effectiveness of basic visualizations.
\newblock \emph{{IEEE} Trans. Vis. Comput. Graph.}, 25(7):2505--2512.

\bibitem[{Shanahan et~al.(2023)Shanahan, McDonell, and Reynolds}]{roleplay}
Murray Shanahan, Kyle McDonell, and Laria Reynolds. 2023.
\newblock Role play with large language models.
\newblock \emph{Nat.}, 623(7987):493--498.

\bibitem[{Shen et~al.(2023)Shen, Shen, Luo, Yang, Hu, Zhang, Tai, and
  Wang}]{DBLP:journals/tvcg/ShenSLYHZTW23}
Leixian Shen, Enya Shen, Yuyu Luo, Xiaocong Yang, Xuming Hu, Xiongshuai Zhang,
  Zhiwei Tai, and Jianmin Wang. 2023.
\newblock Towards natural language interfaces for data visualization: {A}
  survey.
\newblock \emph{{IEEE} Trans. Vis. Comput. Graph.}, 29(6):3121--3144.

\bibitem[{Tang et~al.(2022)Tang, Luo, Ouzzani, Li, and
  Chen}]{DBLP:conf/sigmod/TangLOLC22}
Jiawei Tang, Yuyu Luo, Mourad Ouzzani, Guoliang Li, and Hongyang Chen. 2022.
\newblock Sevi: Speech-to-visualization through neural machine translation.
\newblock In \emph{{SIGMOD} Conference}, pages 2353--2356. {ACM}.

\bibitem[{Tang et~al.(2024)Tang, Yang, Fan, Cao, Luo, and
  Halevy}]{DBLP:conf/cidr/0001YF0LH24}
Nan Tang, Chenyu Yang, Ju~Fan, Lei Cao, Yuyu Luo, and Alon~Y. Halevy. 2024.
\newblock Verifai: Verified generative {AI}.
\newblock In \emph{{CIDR}}. www.cidrdb.org.

\bibitem[{Team et~al.(2024)Team, Anil, Borgeaud, Wu, Alayrac, Yu, Soricut,
  Schalkwyk, Dai, Hauth et~al.}]{gemini}
Gemini Team, Rohan Anil, Sebastian Borgeaud, Yonghui Wu, Jean-Baptiste Alayrac,
  Jiahui Yu, Radu Soricut, Johan Schalkwyk, Andrew~M Dai, Anja Hauth, et~al.
  2024.
\newblock Gemini: a family of highly capable multimodal models.

\bibitem[{Vaswani et~al.(2023)Vaswani, Shazeer, Parmar, Uszkoreit, Jones,
  Gomez, Kaiser, and Polosukhin}]{transformer}
Ashish Vaswani, Noam Shazeer, Niki Parmar, Jakob Uszkoreit, Llion Jones,
  Aidan~N. Gomez, Lukasz Kaiser, and Illia Polosukhin. 2023.
\newblock \href {https://arxiv.org/abs/1706.03762} {Attention is all you need}.
\newblock \emph{Preprint}, arXiv:1706.03762.

\bibitem[{Wang et~al.(2024)Wang, Lv, Yu, Hong, Qi, Wang, Ji, Yang, Zhao, Song,
  Xu, Xu, Li, Dong, Ding, and Tang}]{cogvlm}
Weihan Wang, Qingsong Lv, Wenmeng Yu, Wenyi Hong, Ji~Qi, Yan Wang, Junhui Ji,
  Zhuoyi Yang, Lei Zhao, Xixuan Song, Jiazheng Xu, Bin Xu, Juanzi Li, Yuxiao
  Dong, Ming Ding, and Jie Tang. 2024.
\newblock \href {https://arxiv.org/abs/2311.03079} {Cogvlm: Visual expert for
  pretrained language models}.
\newblock \emph{Preprint}, arXiv:2311.03079.

\bibitem[{Wei et~al.(2022)Wei, Wang, Schuurmans, Bosma, Ichter, Xia, Chi, Le,
  and Zhou}]{chainofthought}
Jason Wei, Xuezhi Wang, Dale Schuurmans, Maarten Bosma, Brian Ichter, Fei Xia,
  Ed~H. Chi, Quoc~V. Le, and Denny Zhou. 2022.
\newblock Chain-of-thought prompting elicits reasoning in large language
  models.
\newblock In \emph{NeurIPS}.

\bibitem[{Xia et~al.(2024)Xia, Zhang, Ye, Yan, Liu, Zhou, Chen, Dou, Shi, Yan
  et~al.}]{chartx}
Renqiu Xia, Bo~Zhang, Hancheng Ye, Xiangchao Yan, Qi~Liu, Hongbin Zhou, Zijun
  Chen, Min Dou, Botian Shi, Junchi Yan, et~al. 2024.
\newblock Chartx \& chartvlm: A versatile benchmark and foundation model for
  complicated chart reasoning.
\newblock \emph{arXiv preprint arXiv:2402.12185}.

\bibitem[{Xie et~al.(2024)Xie, Luo, Li, and
  Tang}]{DBLP:journals/pvldb/XieLLT24}
Yupeng Xie, Yuyu Luo, Guoliang Li, and Nan Tang. 2024.
\newblock Haichart: Human and {AI} paired visualization system.
\newblock \emph{Proc. {VLDB} Endow.}, 17(11):3178--3191.

\bibitem[{Xu et~al.(2023)Xu, Du, Qi, Xu, Yuan, and Guo}]{chartbench}
Zhengzhuo Xu, Sinan Du, Yiyan Qi, Chengjin Xu, Chun Yuan, and Jian Guo. 2023.
\newblock Chartbench: A benchmark for complex visual reasoning in charts.
\newblock \emph{arXiv preprint arXiv:2312.15915}.

\bibitem[{Ye et~al.(2023)Ye, Xu, Ye, Yan, Hu, Liu, Qian, Zhang, Huang, and
  Zhou}]{mplugowl2}
Qinghao Ye, Haiyang Xu, Jiabo Ye, Ming Yan, Anwen Hu, Haowei Liu, Qi~Qian,
  Ji~Zhang, Fei Huang, and Jingren Zhou. 2023.
\newblock \href {https://arxiv.org/abs/2311.04257} {mplug-owl2: Revolutionizing
  multi-modal large language model with modality collaboration}.
\newblock \emph{Preprint}, arXiv:2311.04257.

\bibitem[{Ye et~al.(2024)Ye, Hao, Hou, Wang, Xiao, Luo, and
  Zeng}]{ye2024generative}
Yilin Ye, Jianing Hao, Yihan Hou, Zhan Wang, Shishi Xiao, Yuyu Luo, and Wei
  Zeng. 2024.
\newblock \href {https://arxiv.org/abs/2404.18144} {Generative ai for
  visualization: State of the art and future directions}.
\newblock \emph{Preprint}, arXiv:2404.18144.

\bibitem[{Zeng et~al.(2023)Zeng, Liu, Du, Wang, Lai, Ding, Yang, Xu, Zheng,
  Xia, Tam, Ma, Xue, Zhai, Chen, Zhang, Dong, and Tang}]{GLM}
Aohan Zeng, Xiao Liu, Zhengxiao Du, Zihan Wang, Hanyu Lai, Ming Ding, Zhuoyi
  Yang, Yifan Xu, Wendi Zheng, Xiao Xia, Weng~Lam Tam, Zixuan Ma, Yufei Xue,
  Jidong Zhai, Wenguang Chen, Peng Zhang, Yuxiao Dong, and Jie Tang. 2023.
\newblock \href {https://arxiv.org/abs/2210.02414} {Glm-130b: An open bilingual
  pre-trained model}.
\newblock \emph{Preprint}, arXiv:2210.02414.

\bibitem[{Zeng and Battle(2023)}]{DBLP:conf/chi/ZengB23}
Zehua Zeng and Leilani Battle. 2023.
\newblock A review and collation of graphical perception knowledge for
  visualization recommendation.
\newblock In \emph{{CHI}}, pages 820:1--820:16. {ACM}.

\bibitem[{Zhou et~al.(2023)Zhou, Fung, Chen, Thomas, Ji, and
  Chang}]{enhanced_chart_understanding}
Mingyang Zhou, Yi~R. Fung, Long Chen, Christopher Thomas, Heng Ji, and Shih-Fu
  Chang. 2023.
\newblock \href {https://arxiv.org/abs/2305.18641} {Enhanced chart
  understanding in vision and language task via cross-modal pre-training on
  plot table pairs}.
\newblock \emph{Preprint}, arXiv:2305.18641.

\bibitem[{Zhu et~al.(2024)Zhu, Du, Li, Luo, and
  Tang}]{DBLP:journals/corr/abs-2406-07815}
Yizhang Zhu, Shiyin Du, Boyan Li, Yuyu Luo, and Nan Tang. 2024.
\newblock Are large language models good statisticians?
\newblock \emph{CoRR}, abs/2406.07815.

\end{thebibliography}
